\documentclass{article}

\usepackage{microtype}
\usepackage{graphicx}
\usepackage{subfigure}
\usepackage{booktabs} % for professional tables

\usepackage{hyperref}

\usepackage[accepted]{icml2023}

\usepackage{amsmath}
\usepackage{amssymb}
\usepackage{mathtools}
\usepackage{amsthm}

\usepackage{amsmath}
\usepackage{amssymb}
\usepackage{mathtools}
\usepackage{enumitem}
\usepackage{amsthm}
\usepackage{physics}
\usepackage{algorithm}
\usepackage{algorithmic}
\usepackage{tikz}
\usepackage{placeins}
\usetikzlibrary{calc}

\let\P\relax
\DeclareMathOperator{\P}{P}

\newcommand{\bcket}[3]{\left#1 #3 \right#2}
\newcommand{\mbcket}[5]{\left#1 #4 \middle#2 #5 \right#3}
\renewcommand{\b}{\bcket{(}{)}}
\newcommand{\sqb}{\bcket{[}{]}}

\renewcommand{\P}[1][]{\operatorname{P}_{#1}\b}
\newcommand{\Pdgp}{\P[\text{DGP}]}

\newcommand{\Qdgp}{\Qs[\text{DGP}]}

\newcommand{\Q}[1][]{\operatorname{Q}_{#1}\b}
\newcommand{\Qs}[1][]{\operatorname{Q}_{#1}^*\b}
\newcommand{\N}{\mathcal{N}\b}

\newcommand{\K}{\mathbf{K}}
\newcommand{\Kbnn}{\mathbf{K}_\text{BNN}}

\newcommand{\Gbnn}{\mathbf{G}_\text{BNN}}
\newcommand{\Gdgp}{\mathbf{G}_\text{DGP}}

\renewcommand{\L}{\mathcal{L}}

\newcommand{\R}{\mathbf{R}}
\newcommand{\T}{\mathbf{T}}

\newcommand{\Kg}[2][]{\K^{#1}(\G_{#2})}

\newcommand{\Gt}[2][]{\G^{#1}_\theta(\theta_{#2})}
\newcommand{\U}{\mathbf{U}}
\newcommand{\A}{\mathbf{A}}

\newcommand{\D}{\mathbf{D}}
\newcommand{\V}{\mathbf{V}}
\newcommand{\W}{\mathbf{W}}
\newcommand{\I}{\mathbf{I}}
\newcommand{\Y}{\mathbf{Y}}
\newcommand{\X}{\mathbf{X}}
\newcommand{\y}{\mathbf{y}}
\newcommand{\F}{\mathbf{F}}

\newcommand{\Yr}{\smash{\mathbf{\tilde{Y}}}}
\newcommand{\Fr}{\smash{\mathbf{\tilde{F}}}}
\newcommand{\yr}{\smash{\mathbf{\tilde{y}}}}
\newcommand{\fr}{\smash{\mathbf{\tilde{f}}}}
\newcommand{\G}{\mathbf{G}}

\newcommand{\La}{\mathbf{\Lambda}}

\newcommand{\f}{\mathbf{f}}

\newcommand{\z}{\mathbf{z}}

\newcommand{\w}{\mathbf{w}}
\newcommand{\m}{\boldsymbol{\mu}}
\renewcommand{\S}{\mathbf{\Sigma}}

\newcommand{\const}{\operatorname{const}}
\newcommand{\0}{{\bf {0}}}

\renewcommand{\dd}[2][]{\frac{\partial #1}{\partial #2}}

\newcommand{\slL}[1]{\{ #1 \}_{\ell=1}^L}

\newcommand{\tprod}{{\textstyle \prod}}
\newcommand{\prodln}{{\textstyle \prod}_{\lambda=1}^{N_\ell}}
\newcommand{\prodlnlp}{{\textstyle \prod}_{\lambda=1}^{\nu_{L+1}}}

\DeclareMathOperator{\E}{\mathbb{E}}
\DeclareMathOperator*{\Var}{\mathbb{V}}

\newcommand{\KL}{\operatorname{D}_\text{KL}\mbcket{(}{\Vert}{)}}
\newcommand{\Nx}{\nu_0}
\newcommand{\Ny}{\nu_{L+1}}
\newcommand{\tsum}{{\textstyle \sum}}
\newcommand{\Pc}[1][]{\operatorname{P}_{#1}\bc}
\newcommand{\bc}{\mbcket{(}{\vert}{)}}

\newcommand{\LM}{\mathcal{L}}

\newcommand{\limit}{Bayesian representation learning limit}
\newcommand{\dkmearly}{DKM objective}

\DeclareMathSymbol{\mdot}{\mathord}{symbols}{"01}

\icmltitlerunning{A theory of representation learning gives a deep generalisation of kernel methods}

\begin{document}

\twocolumn[
\icmltitle{A theory of representation learning\\gives a deep generalisation of kernel methods}



\icmlsetsymbol{equal}{*}

\begin{icmlauthorlist}
\icmlauthor{Adam X. Yang}{bris}
\icmlauthor{Maxime Robeyns}{bris}
\icmlauthor{Edward Milsom}{bris}
\icmlauthor{Ben Anson}{bris}
\icmlauthor{Nandi Schoots}{kings}
\icmlauthor{Laurence Aitchison}{bris}
\end{icmlauthorlist}

\icmlaffiliation{bris}{University of Bristol}
\icmlaffiliation{kings}{Kings College London}

\icmlcorrespondingauthor{Laurence Aitchison}{laurence.aitchison@gmail.com}

\icmlkeywords{Machine Learning, ICML}

\vskip 0.3in
]



\printAffiliationsAndNotice{}  % leave blank if no need to mention equal contribution

\begin{abstract}
The successes of modern deep machine learning methods are founded on their ability to transform inputs across multiple layers to build good high-level representations. 
It is therefore critical to understand this process of representation learning.
However, standard theoretical approaches (formally NNGPs) involving infinite width limits eliminate representation learning.
We therefore develop a new infinite width limit, the Bayesian representation learning limit, that exhibits representation learning mirroring that in finite-width models, yet at the same time, retains some of the simplicity of standard infinite-width limits.
In particular, we show that Deep Gaussian processes (DGPs) in the Bayesian representation learning limit have exactly multivariate Gaussian posteriors, and the posterior covariances can be obtained by optimizing an interpretable objective combining a log-likelihood to improve performance with a series of KL-divergences which keep the posteriors close to the prior.
We confirm these results experimentally in wide but finite DGPs.
Next, we introduce the possibility of using this limit and objective as a flexible, deep generalisation of kernel methods, that we call deep kernel machines (DKMs).
Like most naive kernel methods, DKMs scale cubically in the number of datapoints.
We therefore use methods from the Gaussian process inducing point literature to develop a sparse DKM that scales linearly in the number of datapoints.
Finally, we extend these approaches to NNs (which have non-Gaussian posteriors) in the Appendices.

%We develop a new infinite width limit, the representation learning limit, that exhibits representation learning mirroring that in finite-width networks, yet retains the theoretical tractability of standard infinite width limits.
%The learned representations are governed given by optimizing performance, while regularising the representations to be similar to those in the standard infinite width limit (formally, the NNGP).
%We apply this limit to both deep neural networks (DNNs) and deep Gaussian processes (DGPs). 
%For DGPs, the objective and posteriors are particularly simple (the posteriors are exactly multivariate Gaussian).
%Finally, we use this limit and objective to develop a flexible, deep generalisation of kernel methods, that we call deep kernel machines (DKMs).
%We show that DKMs can be scaled to large datasets using methods inspired by inducing point methods from the Gaussian process literature, and we show that DKMs exhibit superior performance to other kernel-based approaches.
\end{abstract}

\newcommand{\sty}{icml2023}
\section{Introduction}
\label{sec:intro}

The successes of modern machine learning methods from neural networks (NNs) to deep Gaussian processes \citep[DGPs][]{damianou2013deep,salimbeni2017doubly} is based on their ability to use depth to transform the input into high-level representations that are good for solving difficult tasks \citep{bengio2013representation,lecun2015deep}.
However, theoretical approaches using infinite limits to understand deep models struggle to capture representation learning.
In particular, there are two broad families of infinite limit, and while they both use kernel-matrix-like objects they are ultimately very different.
First is the neural network Gaussian process \citep[NNGP][]{neal1996priors,lee2017deep,matthews2018gaussian} which applies to Bayesian models like Bayesian neural networks (BNNs) and DGPs and describes the representations at each layer (formally, the NNGP kernel is raw second moment of the activities).
Second is the neural tangent kernel \citep[NTK][]{jacot2018neural}, which is a very different quantity that involves gradients, and describes how predictions at all datapoints change if we do a gradient update on a single datapoint.
As such, the NNGP and NTK are suited to asking very different theoretical questions.
For instance, the NNGP is better suited to understanding the transformation of representations across layers, while the NTK is better suited to understanding how predictions change through NN training.

While challenges surrounding representation learning have recently been addressed in the NTK setting \cite{yang2020feature}, we are the first to address this challenge in the NNGP setting.

At the same time, kernel methods \citep{smola1998learning,shawe2004kernel,hofmann2008kernel} were a leading machine learning approach prior to the deep learning revolution \cite{krizhevsky2012imagenet}.
However, kernel methods were eclipsed by \textit{deep} NNs because depth gives NNs the flexibility to learn a good top-layer representation \citep{aitchison2020bigger}.
In contrast, in a standard kernel method, the kernel (or equivalently the representation) is highly inflexible --- there are usually a few tunable hyperparameters, but nothing that approaches the enormous flexibility of the top-layer representation in a deep model.
There is therefore a need to develop flexible, deep generalisations of kernel method.
Remarkably, our advances in understanding representation learning in DGPs give such a flexible, deep kernel method.

Follow-up work has generalised the DKM to the convolutional setting, and obtains results of e.g.\ 92\% on CIFAR-10 and 71\% on CIFAR-100 \citep{milsom2023convolutional}.

\section{Contributions}
\begin{itemize}[noitemsep]
  \item We present a new infinite width limit, the \limit{}, that retains representation learning in deep Bayesian models including DGPs.  The key insight is that as the width goes to infinity, the prior becomes stronger, and eventually overwhelms the likelihood.  We can fix this by rescaling the likelihood to match the prior.  This rescaling can be understood in a Bayesian context as copying the labels (Sec.~\ref{sec:dkm}).
  \item We show that in the \limit{}, DGP posteriors are exactly zero-mean multivariate Gaussian, $\P{\f_\lambda^\ell| \X, \y} = \N{\f_\lambda^\ell; \0, \G_\ell}$ where $\f_\lambda^\ell$, is the activation of the $\lambda$th feature in layer $\ell$ for all inputs (Sec.~\ref{sec:exact_dgp_post} and Appendix~\ref{sec:true_post:bnn}).
  \item We show that the posterior covariances can be obtained by optimizing the ``deep kernel machine objective'',
  \begin{multline}
    %\label{eq:post_dkm_obj}
    %\nonumber
    \L(\G_1,\dotsc,\G_L) = \log \P{\Y| \G_L} \\- \tsum_{\ell=1}^L \nu_\ell \KL{\N{\0, \G_\ell}}{\N{\0, \Kg{\ell-1}}},
  \end{multline}
  where $\G_\ell$ are the posterior covariances, $\Kg{\ell-1}$ are the kernel matrices, and $\nu_\ell$ accounts for any differences in layer width (Sec.~\ref{sec:dkm}).
  \item We give an interpretation of this objective, with $\log \P{\Y| \G_L}$ encouraging improved performance, while the KL-divergence terms act as a regulariser, keeping posteriors, $\N{\0, \G_\ell}$, close to the prior, $\N{\0, \Kg{\ell-1}}$ (Sec.~\ref{sec:intuitive}).
  \item We introduce a sparse DKM, which takes inspiration from GP inducing point literature to obtain a practical, scalable method that is linear in the number of datapoints. In contrast, naively computing/optimizing the DKM objective is cubic in the number of datapoints (as with most other naive kernel methods; Sec.~\ref{sec:performance}).
  \item We extend these results to BNNs (which have non-Gaussian posteriors) in Appendix~\ref{sec:bnn}.
\end{itemize}

\section{Related work}
Our work is focused on DGPs and gives new results such as the extremely simple multivariate Gaussian form for DGP true posteriors.
As such, our work is very different from previous work on NNs, where such results are not available.
There are at least four families of such work.

First, there is recent work on representation learning in the very different NTK setting \citep{jacot2018neural,yang2019scaling,yang2020feature} (see Sec.~\ref{sec:intro}).
In contrast, here we focus on NNGPs \citep{neal1996priors,williams1996computing,lee2017deep,matthews2018gaussian,novak2018bayesian,garriga2018deep,jacot2018neural}, where the challenge of representation learning has yet to be addressed.
As work on the NTK does not consider the Bayesian posterior, and does not consider the DGP setting, they do not find results such as the extremely simple multivariate Gaussian form for the true posterior.

Second, there is a body of work using methods from physics to understand representation learning in neural networks \citep{antognini2019finite,dyer2019asymptotics,hanin2019finite,aitchison2020bigger,li2020statistical,yaida2020non,naveh2020predicting,zavatone2021asymptotics,zavatone2021exact,roberts2021principles,naveh2021self,halverson2021neural}.
Again, this work focuses on the finite NN setting, so does give results such as the extremely simple multivariate Gaussian form for the true posterior.

Third, there is a body of theoretical work including \citep{mei2018mean,nguyen2019connected,sirignano2020mean,sirignano2020meanb,nguyen2020rigorous}
which establishes properties such as convergence to the global optimum.
This work is focused on two-layer (or one-hidden layer network) networks, and like the NTK, considers learning under SGD rather than Bayesian posteriors.

Fourth, deep kernel~\textit{processes} (rather than our \textit{machines}).
These are fully Bayesian deep nonlinear function approximators that optimize a variational approximate posterior over Gram matrices \citep{aitchison2020deep,ober2021variational,ober2023improveddkp}.
Similarly to DKMs, deep kernel processes (DKPs) propagate kernels between layers, though DKPs require sampling kernels at each layer, whereas each kernel in a DKM is deterministic.

Another related line of work uses kernels to give a closed-form expression for the weights of a neural network, based on a greedy-layerwise objective \citep{wu2022deep}.
This work differs in that it uses the HSIC objective, and therefore does not have a link to DGPs or Bayesian neural networks, and in that it uses a greedy-layerwise objective, rather than end-to-end gradient descent.

%
%
%Finally, our work has implications for understanding convergence and loss-landscapes in NNs and DGPs.
%In particular, we have conjectured that the deep kernel machine objective is unimodal, which would suggest that multimodality in DGPs arises solely from the unitary symmetries Eq.~\eqref{eq:gram-rotation}, and that multimodality in NNs arises from permutation symmetries, where the exact same input-output function can be achieved by swapping the identities of the intermediate layer weights \citep{mackay1992practical,chen1993geometry}.
%This hypothesis was recently conjectured for finite neural networks \citep{entezari2021role} as a potential explanation of mode connectivity \citep{garipov2018loss}.

\section{Results}
\label{sec:results}
%\subsection{A general model containing DNNs and DGPs}
%\label{sec:back:dgp}
We start by defining a DGP; see Appendix~\ref{sec:bnn} for Bayesian NN (BNNs).
%In particular, our model contains BNNs with an IID Gaussian prior over the weights (, which are equivalent to non-Bayesian BNNs where we add Gaussian noise during training \citealp{welling2011bayesian}).
This model maps from inputs, $\X \in \mathbb{R}^{P\times \Nx}$, to outputs, $\Y\in\mathbb{R}^{P\times \Ny}$, where $P$ is the number of input points, $\Nx$ is the number of input features, and $\Ny$ is the number of output features. 
The model has $L$ intermediate layers, indexed $\ell\in\{1,\dotsc,L\}$, and at each intermediate layer there are $N_\ell$ features, $\F_\ell \in \mathbb{R}^{P\times N_\ell}$.
Both $\F_\ell$ and $\Y$ can be written as a stack of vectors,
\newcommand{\flpos}[2]{#2}
\begin{subequations}
\begin{align}
\F_\ell &= (\flpos{1}{\f_1^\ell} \quad \flpos{2}{\f_2^\ell} \quad \dotsm  \quad \f_{N_\ell}^\ell) \\
\Y &= (\flpos{1}{\y_1} \quad \flpos{2}{\y_2} \quad \dotsm  \quad \y_{\nu_{L+1}}),
\end{align}
\end{subequations}
where $\f_\lambda^\ell \in \mathbb{R}^P$ gives the value of one feature and $\y_\lambda \in \mathbb{R}^P$ gives the value of one output for all $P$ input points.
The features, $\F_1,\dotsc,\F_L$, and (for regression) the outputs, $\Y$, are sampled from a Gaussian process (GP) with a covariance which depends on the previous layer features (Fig.~\ref{fig:graphical_model} top),
\begin{subequations}
\label{eq:deepgp}
\begin{align}
  \label{eq:deepgp:top:F}
  \P{\F_\ell| \F_{\ell-1}} &= \prodln \N{\f_\lambda^\ell; \0, \K(\G(\F_{\ell-1}))}\\
  \label{eq:deepgp:like}
  \P{\Y| \F_{L}} &= \prodlnlp \N{\y_\lambda; \0, \K(\G(\F_L)) + \sigma^2 \I}.
\end{align}
\end{subequations}
Note we only use the regression likelihood~\eqref{eq:deepgp:like} to give a concrete example; we could equally use an alternative likelihood e.g.\ for classification (Appendix~\ref{app:like}).
The distinction between DGPs and BNNs arises through the choice of $\K(\cdot)$ and $\G(\cdot)$.
For BNNs, see Appendix~\ref{sec:bnn}.
For DGPs, $\G(\cdot)$, which takes the features and computes the corresponding $P \times P$ Gram matrix, is
\begin{align}
  \nonumber
  \G(\F_{\ell-1}) &= \tfrac{1}{N_{\ell-1}} \tsum_{\lambda=1}^{N_{\ell-1}} \f_\lambda^{\ell-1} (\f_\lambda^{\ell-1})^T \\
  \label{eq:Gdgp}
  &= \tfrac{1}{N_{\ell-1}} \F_{\ell-1} \F_{\ell-1}^T.
\end{align}
Now, we introduce random variables representing the Gram matrices, $\G_{\ell-1} = \G(\F_{\ell-1})$, where $\G_{\ell-1}$ is a random variable representing the Gram matrix at layer $\ell-1$, whereas $\G(\cdot)$ is a deterministic function that takes features and computes the corresponding Gram matrix using Eq.~\eqref{eq:Gdgp}.
Finally, $\K(\cdot)$, transforms the Gram matrices, $\G_{\ell-1}$ to the final kernel.
%For BNNs, this function is the identity, while for DGPs, we can choose many options depending on the kernel.  As an example, here we give the form for an isotropic kernel which depends only on the (normalized) squared distance, $R_{ij}(\G_{\ell-1})$ %(remember that non-isotropic kernels such as the arccos also work \citep{cho2009kernel}),
Many kernels of interest are isotropic, meaning they depend only on the normalized squared distance between datapoints, $R_{ij}$,
%\begin{subequations}
\begin{align}
  %\label{eq:Kbnn}
  %\Kbnn(\G_{\ell-1}) &= \G_{\ell-1},\\
  \label{eq:Kdgp}
  K_{\text{isotropic}; ij}(\G_{\ell-1}) &= k_\text{isotropic}\b{R_{ij}(\G_{\ell-1})}.
\end{align}
%\end{subequations}
%For BNNs, $\Kbnn(\cdot)$ is the identity, so that combining Eq.~\eqref{eq:Gbnn} with Eq.~\eqref{eq:Kbnn} gives Eq.~\eqref{eq:bnn:kernel}. 
%meaning they depend only on the (normalized) squared distance, and we can compute this squared distance from $\G_{\ell-1}$, without needing $\F_{\ell-1}$,
Importantly, we can compute this squared distance from $\G_{\ell-1}$, without needing $\F_{\ell-1}$,
\begin{align}
  %\nonumber
  %R_{\text{f}; ij}(\F_{\ell-1}) &= \tfrac{1}{N_{\ell-1}} \tsum_{\lambda=1}^{N_{\ell-1}} \big(F^{\ell-1}_{i \lambda} - F^{\ell-1}_{j \lambda} \big)^2
  %= \tfrac{1}{N_{\ell-1}} \tsum_{\lambda=1}^{N_{\ell-1}} \big(\big(F^{\ell-1}_{i \lambda}\big)^2 - 2F^{\ell-1}_{i \lambda}F^{\ell-1}_{j \lambda} + \big(F^{\ell-1}_{j \lambda}\big)^2\big) \\
  %\label{eq:R}
  %&= G^{\ell-1}_{ii} - 2 G^{\ell-1}_{ij} + G^{\ell-1}_{jj} = R_{ij}(\G).
  \nonumber
  R_{ij}(\G) &= \tfrac{1}{N} \tsum_{\lambda=1}^{N} \big(F_{i \lambda} - F_{j \lambda} \big)^2\\
  \nonumber
  &= \tfrac{1}{N} \tsum_{\lambda=1}^{N} \big(\big(F_{i \lambda}\big)^2 - 2F_{i \lambda}F_{j \lambda} + \big(F_{j \lambda}\big)^2\big) \\
  \label{eq:R}
  &= G_{ii} - 2 G_{ij} + G_{jj},
\end{align}
where $\lambda$ indexes features, $i$ and $j$ index datapoints and we have omitted the layer index for simplicity.
Importantly, we are not restricted to isotropic kernels: other kernels that depend only on the Gram matrix, such as the arccos kernels from the infinite NN literature \citep{cho2009kernel} can also be used (for further details, see \citealp{aitchison2020deep}).
%However, it turns out that almost all standard kernels used in the GP literature depend on the features only through the Gram matrix (for further details, see \citealp{aitchison2020deep}).
%This includes all isotropic kernels that depend on distance, but also non-isotropic kernels such as the arccos kernels from the infinite NN literature \citep{cho2009kernel}.
%Isotropic kernels work because the (normalized) squared distance, $R_{ij}(\G)$, is a function of the Gram matrix,
%
%Except where specifically indicated, all results in the paper will apply to BNNs and DGPs, and indeed combinations of these models obtained by combining BNN nonlinearities in Eq.~\eqref{eq:Gbnn} and DGP kernels in Eq.~\eqref{eq:Kdgp}.

\subsection{BNN and DGP priors can be written purely in terms of Gram matrices}
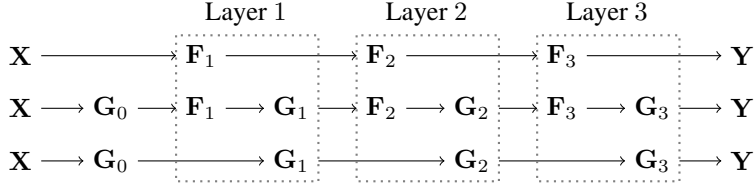
\begin{figure*}
  \centering
  \begin{tikzpicture}
    \def\d{1.2cm}
    \def\dy{0.7cm}
    \node[] (tX)  at ({0*\d}, 0)         {$\X$};
    \node[] (tF1) at ({2*\d}, 0)         {$\F_1$};
    %\node[] (tH1) at ({3*\d}, 0)         {$\H_1$};
    \node[] (tF2) at ({4*\d}, 0)         {$\F_2$};
    %\node[] (tH2) at ({6*\d}, 0)         {$\H_2$};
    \node[] (tF3) at ({6*\d}, 0)         {$\F_3$};
    %\node[] (tH3) at ({9*\d}, 0)         {$\H_3$};
    \node[] (tY)  at ({8*\d}, 0)         {$\Y$};
    \draw[->] (tX)  -- (tF1);
    %\draw[->] (tF1) -- (tH1);
    %\draw[->] (tH1) -- (tF2);
    \draw[->] (tF1) -- (tF2);
    %\draw[->] (tF2) -- (tH2);
    %\draw[->] (tH2) -- (tF3);
    \draw[->] (tF2) -- (tF3);
    %\draw[->] (tF3) -- (tH3);
    %\draw[->] (tH3) -- (tY);
    \draw[->] (tF3) -- (tY);
    
    \node[] (mX)  at ({0*\d}, -\dy) {$\X$};
    \node[] (mK0) at ({1*\d}, -\dy) {$\G_0$};
    \node[] (mF1) at ({2*\d}, -\dy) {$\F_1$};
    \node[] (mK1) at ({3*\d}, -\dy) {$\G_1$};
    \node[] (mF2) at ({4*\d}, -\dy) {$\F_2$};
    \node[] (mK2) at ({5*\d}, -\dy) {$\G_2$};
    \node[] (mF3) at ({6*\d}, -\dy) {$\F_3$};
    \node[] (mK3) at ({7*\d}, -\dy) {$\G_3$};
    \node[] (mY)  at ({8*\d}, -\dy) {$\Y$};
    \draw[->] (mX)  -- (mK0);
    \draw[->] (mK0) -- (mF1);
    \draw[->] (mF1) -- (mK1);
    \draw[->] (mK1) -- (mF2);
    \draw[->] (mF2) -- (mK2);
    \draw[->] (mK2) -- (mF3);
    \draw[->] (mF3) -- (mK3);
    \draw[->] (mK3) -- (mY);
    
    \node[] (bX)  at ({0*\d}, {-2*\dy}) {$\X$};
    \node[] (bK0) at ({1*\d}, {-2*\dy}) {$\G_0$};
    \node[] (bK1) at ({3*\d}, {-2*\dy}) {$\G_1$};
    \node[] (bK2) at ({5*\d}, {-2*\dy}) {$\G_2$};
    \node[] (bK3) at ({7*\d}, {-2*\dy}) {$\G_3$};
    \node[] (bY)  at ({8*\d}, {-2*\dy}) {$\Y$};
    \draw[->] (bX)  -- (bK0);
    \draw[->] (bK0) -- (bK1);
    \draw[->] (bK1) -- (bK2);
    \draw[->] (bK2) -- (bK3);
    \draw[->] (bK3) -- (bY);
    
    \draw[gray,thick,dotted] (tF1.north west) rectangle (bK1.south east);
    \draw[gray,thick,dotted] (tF2.north west) rectangle (bK2.south east);
    \draw[gray,thick,dotted] (tF3.north west) rectangle (bK3.south east);
    
    \node[anchor=south] at ($0.5*(tF1.north) + 0.5*(tF1.north -| mK1.north)$) {Layer $1$};
    \node[anchor=south] at ($0.5*(tF2.north) + 0.5*(tF2.north -| mK2.north)$) {Layer $2$};
    \node[anchor=south] at ($0.5*(tF3.north) + 0.5*(tF3.north -| mK3.north)$) {Layer $3$};
  \end{tikzpicture}
  \caption{
    The graphical model structure for each of our generative models for $L=3$.
    \textbf{Top}. The standard model (Eq.~\ref{eq:deepgp}), written purely in terms of features, $\F_\ell$.
    \textbf{Middle}. The standard model, including Gram matrices as random variables (Eq.~\ref{eq:prior:bnn})
    \textbf{Bottom}. Integrating out the activations, $\F_\ell$, 
    \label{fig:graphical_model}
  }
\end{figure*}
Notice that $\F_\ell$ depends on $\F_{\ell-1}$ only through $\G_{\ell-1}=\G(\F_{\ell-1})$, and $\Y$ depends on $\F_L$ only through $\G_L=\G(\F_L)$ (Eq.~\ref{eq:deepgp}).
We can therefore write the graphical model in terms of those Gram matrices (Fig.~\ref{fig:graphical_model} middle).
%, $\G_\ell = \G(\F_\ell)$, where $\G_\ell$ is a random variable representing the Gram matrix at layer $\ell$, whereas $\G(\cdot)$ is a deterministic function that takes features and computes the corresponding Gram matrix using Eq.~\eqref{eq:bnn:Gs}.
%The generative model can then be rewritten in terms of the Gram matrices (Fig.~\ref{fig:graphical_model} middle),
\begin{subequations}
\label{eq:prior:bnn}
\begin{align}
  \label{eq:F:like}
  \P{\F_\ell| \G_{\ell-1}} &= \tprod_{\lambda=1}^{N_\ell} \N{\f_\lambda^\ell; \0, \Kg{\ell-1}}\\
  \label{eq:K:like}
  \P{\G_\ell| \F_\ell} &= \delta\b{\G_\ell - \G(\F_\ell)}\\
  \label{eq:like}
  \P{\Y| \G_L} &= \prodlnlp \N{\y_\lambda; \0, \Kg{L} + \sigma^2 \I}.
\end{align}
\end{subequations}
where $\delta$ is the Dirac-delta, and $\G_0$ depends on $\X$ (e.g.\ $\G_0 = \tfrac{1}{\nu_0} \X \X^T$).
Again, for concreteness we have used a regression likelihood, but other likelihoods could also be used.

Now, we can integrate $\F_\ell$ out of the model, in which case, we get an equivalent generative model written solely in terms of Gram matrices (Fig.~\ref{fig:graphical_model} bottom), with
%\begin{subequations}
\begin{align}
  \label{eq:dkp:prior}
  \P{\G_\ell| \G_{\ell-1}} &= \int d\F_\ell \; \P{\G_\ell| \F_\ell} \P{\F_\ell| \G_{\ell-1}},%\\
  %\label{eq:dkp:like}
  %\P{\Y| \G_L} &= \prodlnlp \N{\y_\lambda; \0, \Kg{L} + \sigma^2 \I}.
\end{align}
%\end{subequations}
and with the usual likelihood (e.g.\ Eq.~\ref{eq:like}).
This looks intractable (and indeed, in general it is intractable).
However for DGPs, an analytic form is available.
In particular, note the Gram matrix (Eq.~\ref{eq:Gdgp}) is the outer product of IID Gaussian distributed vectors (Eq.~\ref{eq:deepgp:top:F}). 
This matches the definition of the Wishart distribution~\citep{gupta2018matrix}, so we have,
\begin{align}
  \label{eq:P[DGP](G|G)}
  \P{\G_\ell| \G_{\ell-1}} &= \text{Wishart}\b{\G_\ell; \tfrac{1}{N_\ell} \Kg{\ell-1}, N_\ell}
\end{align}
\vspace{-24pt}
\begin{multline}
  \log \P{\G_\ell| \G_{\ell-1}} =\\ \tfrac{N_\ell {-} P {-} 1}{2} \log \abs{\G_\ell} {-} \tfrac{N_\ell}{2} \log \abs{\Kg{\ell-1}} \\{-} \tfrac{N_\ell}{2} \Tr\b{\Kg[-1]{\ell-1} \G_\ell} {+} \alpha_\ell,
\end{multline}
where
\begin{align}
  \label{eq:wishart_const}
  \alpha_\ell\ =  -\tfrac{N_{\ell} P}{2}\log 2 +\tfrac{N_{\ell}P}{2}\log N_\ell - \log \Gamma_P\left(\tfrac{N_l}{2}\right)
\end{align}
is constant wrt all $\G_\ell$ and $\Gamma_P$ is the multivariate Gamma function.
This distribution over Gram matrices is valid for DGPs of any width (though we need to be careful in the low-rank setting where $N_\ell < P$).
%However, for BNNs no such analytic form is available; instead, to make progress we need to consider the infinite width limit.
%we get considerably more theoretical insight, and tractable BNN posteriors, in the limit of an infinitely wide network.
%However, for BNNs the integral in Eq.~\eqref{eq:dkp:prior} is indeed intractable, so we can only make useful progress in 
We are going to leverage these Wishart distributions to understand the behaviour of the Gram matrices in the infinite width limit.

\subsection{Standard infinite width limits of DGPs lack representation learning}
\label{sec:back:inf}
We are now in a position to take a new viewpoint on the DGP analogue of standard NNGP results \citep{lee2017deep,matthews2018gaussian,hron2020exact,pleiss2021limitations}.
We can then evaluate the log-posterior for a model written only in terms of Gram matrices,
\begin{multline}
  \label{eq:dgp:joint}
  \log \P{\G_1,\dotsc,\G_L| \X, \Y} = \log \P{\Y| \G_L} \\+ \tsum_{\ell=1}^L \log \P{\G_\ell| \G_{\ell-1}} + \const,
\end{multline}
where $\const$ is constant and independent of $\G_1\dotsc,\G_L$.
Then we take the limit of infinite width,
\begin{align}
  %\label{eq:limit}
  N_\ell &= N \; \nu_\ell &&\text{for} & \ell &\in \{1,\dotsc,L\} && \text{with} & N &\rightarrow \infty.
\end{align}
%
%\begin{gather}
%  %\label{eq:limit}
%  N_\ell = N \nu_\ell \quad \text{for} \quad \ell \in \{1,\dotsc,L\} \quad \text{with} \quad N \rightarrow \infty.
%%\end{align}
%%This limit is highly analytically tractable because as $N_\ell \rightarrow \infty$, the prior and posterior distribution over $\G_\ell$ become point distributions at $\K(\G_{\ell-1})$.
%%This result was originally derived using the law of large numbers.
%%In particular,
%%\begin{align}
%%  \
%%\end{align}
%%It is straightforward to reproduce this result using our form for $\log \P{\G_\ell| \G_{\ell-1}}$.
%%The log-posterior is,
%%begin{align}
%  %\label{eq:dgp:joint}
%  %\log \P{\G_1,\dotsc,\G_L| \X, \Y} = \log \P{\Y| \G_L} + \tsum_{\ell=1}^L \log \P{\G_\ell| \G_{\ell-1}} + \const.
%\end{gather}
This limit modifies $\log \P{\G_\ell| \G_{\ell-1}}$ (Eq.~\ref{eq:P[DGP](G|G)}), but does not modify $\G_1,\dotsc,\G_L$ in Eq.~\eqref{eq:dgp:joint} as we get to choose the values of $\G_1,\dotsc,\G_L$ at which to evaluate the log-posterior.
Specifically, the log-prior, $\log \P{\G_\ell| \G_{\ell-1}}$ (Eq.~\ref{eq:P[DGP](G|G)}), scales with $N_\ell$ and hence with $N$.
To get a finite limit, we therefore need to divide by $N$,
\begin{align}
  \label{eq:log_wishart_lim}
  &\lim_{N\rightarrow \infty}  \tfrac{1}{N} \log \P{\G_\ell| \G_{\ell-1}}\\
  \nonumber
  &=\tfrac{\nu_\ell}{2} \b{\log \abs{\Kg[-1]{\ell-1} \G_\ell} - \Tr\b{\Kg[-1]{\ell-1} \G_\ell}}\nonumber \\
  &\quad+\lim_{N\rightarrow\infty}\tfrac{\alpha_\ell}{N}\nonumber\\
  &= -\nu_\ell \KL{\N{\0, \G_\ell}}{\N{\0, \Kg{\ell-1}}} + \const,
\end{align}
We justify that $\lim_{N\rightarrow\infty} \alpha_\ell / N$ exists and is constant in Appendix~\ref{app:asymp_wishart}.
Remarkably limit~\eqref{eq:log_wishart_lim} can be written as the KL-divergence between two multivariate Gaussians.
In contrast, the log likelihood, $\log \P{\Y| \G_L}$, is constant wrt $N$ (Eq.~\ref{eq:like}), so $\lim_{N \rightarrow \infty} \tfrac{1}{N} \log \P{\Y| \G_L} = 0$.  
The limiting log-posterior is thus,
%We therefore expect the prior to dominate the likelihood, and this can be seen explicitly by considering the limit of the log-joint divided by $N$ (we divide by $N$ to ensure we get a finite quantity in the infinite limit),
\begin{multline}
  \label{eq:nngp_lim}
  \lim_{N\rightarrow \infty}  \tfrac{1}{N} \log \P{\G_1,\dotsc,\G_L| \X, \Y} 
  = \\- \tsum_{\ell=1}^L \nu_\ell \KL{\N{\0, \G_\ell}}{\N{\0, \Kg{\ell-1}}} + \const.
\end{multline}
This form highlights that the log-posterior scales with $N$, so in the limit as $N\rightarrow \infty$, the posterior converges to a point distribution at the global maximum, denoted $\G_1^*,\dotsc,\G_L^*$, (see Appendix~\ref{app:dgp:weak} for a formal discussion of weak convergence),
\begin{align}
  \label{eq:point}
  \lim_{N \rightarrow \infty} \P{\G_1,\dotsc,\G_L| \X, \Y} &= \tprod_{\ell=1}^L \delta\b{\G_\ell - \G_\ell^*}.
\end{align}
Moreover, it is evident from the KL-divergence form for the log-posterior (Eq.~\ref{eq:nngp_lim}) that the unique global maximum can be computed recursively as $\G^*_\ell = \K(\G^*_{\ell-1})$, with e.g.\ $\G_0^* = \tfrac{1}{\nu_0} \X \X^T$.
Thus, the limiting posterior over Gram matrices does not depend on the training targets, so there is no possibility of representation learning \citep{aitchison2020bigger}.
This is concerning as the successes of modern deep learning arise from flexibly learning good top-layer representations.

%Indeed, the posterior over Gram matrices takes on a very simple, easily computable form.
%In particular, the log-posterior scales with $N$, so in the limit as $N\rightarrow \infty$, the posterior, $\P[\text{DGP}]{\G_1,\dotsc,\G_1| \X, \Y}$, becomes a point distribution concentrated around the global maximum of the log-posterior, denoted $\G_1^*,\dotsc,\G_L^*$,
%\begin{align}
%  \label{eq:point}
%  \lim_{N \rightarrow \infty} \P{\G_1,\dotsc,\G_L| \X, \Y} &= \tprod_{\ell=1}^L \delta\b{\G_\ell - \G_\ell^*}.
%\end{align}
%In the DGP case, the global maximum of Eq.~\eqref{eq:nngp_lim} is $\G^*_\ell = \K(\G^*_{\ell-1})$, with $\G_0^* = \tfrac{1}{\nu_0} \X \X^T$.
%We give a formal argument for weak convergence in Appendix~\ref{app:dgp:weak}.

\begin{figure*}[t!]
    \centering
    \includegraphics[width=\textwidth]{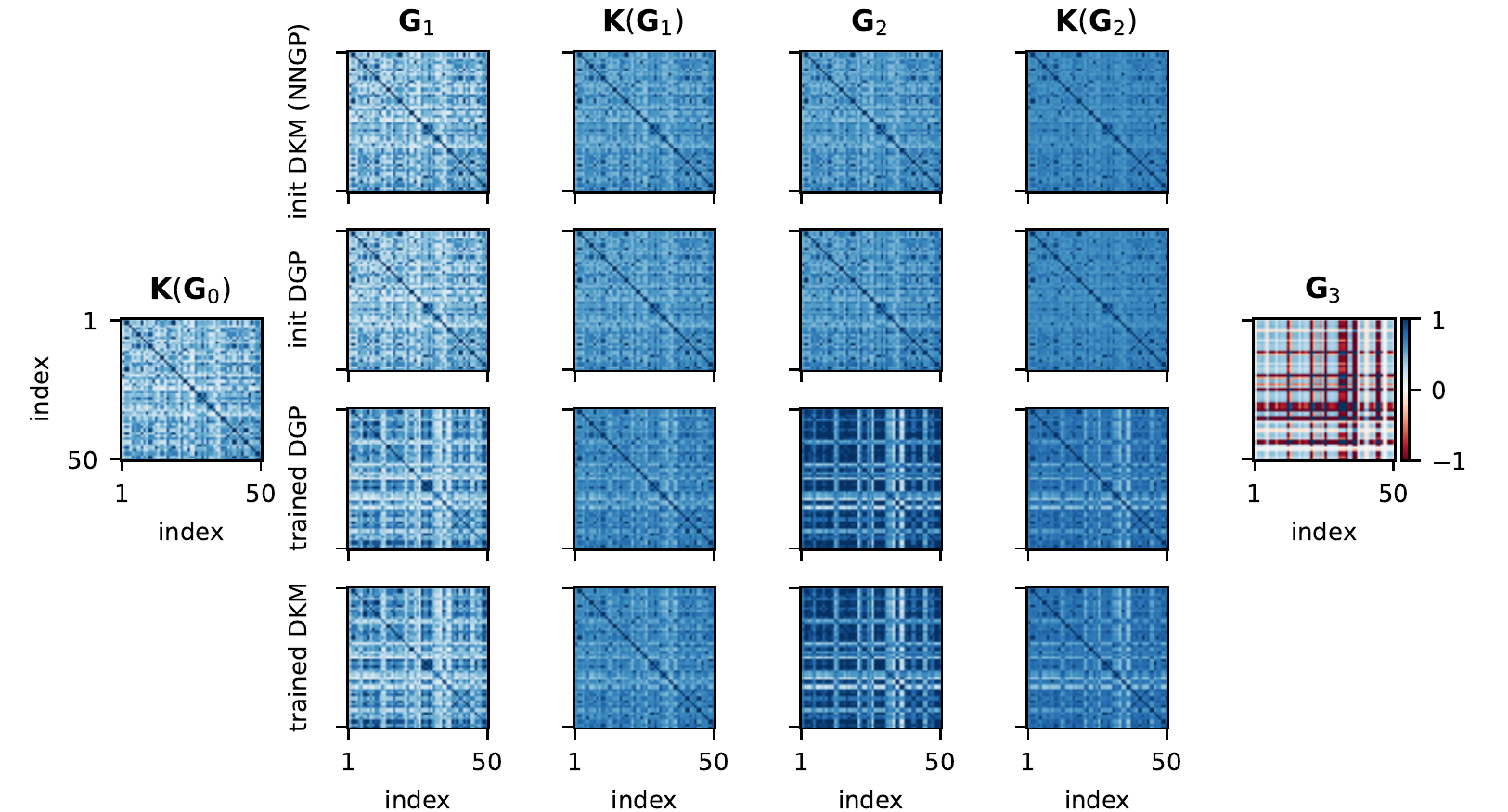}
    \caption{A two hidden layer DGP with 1024 units per hidden layer and DKM with squared exponential kernels match closely. 
    The data was the first 50 datapoints of the yacht dataset.
    The first column, $\K_0$ is a fixed squared exponential kernel applied to the inputs, and the last column, $\G_3 = \y \y^T$ is the fixed output Gram matrix.
    The first row is the DKM initialization at the prior Gram matrices and kernels which is equivalent to an NNGP.  The second row is the DGP, which is initialized by sampling from the prior.  
    As expected, the finite width DGP prior closely matches the infinite-width DKM initialization, which corresponds to the standard infinite width limit.
    The third row is the Gram matrices and kernels for the trained DGP, which has changed dramatically relative to its initialization (second row) in order to better fit the data.
    The fourth row is the Gram matrices and kernels for the optimized DKM, which closely matches those for the trained DGP.
    \label{fig:twolayer_yacht}
  }
\end{figure*}
\begin{figure*}[t!]
  \centering
  \includegraphics[width=0.8\textwidth]{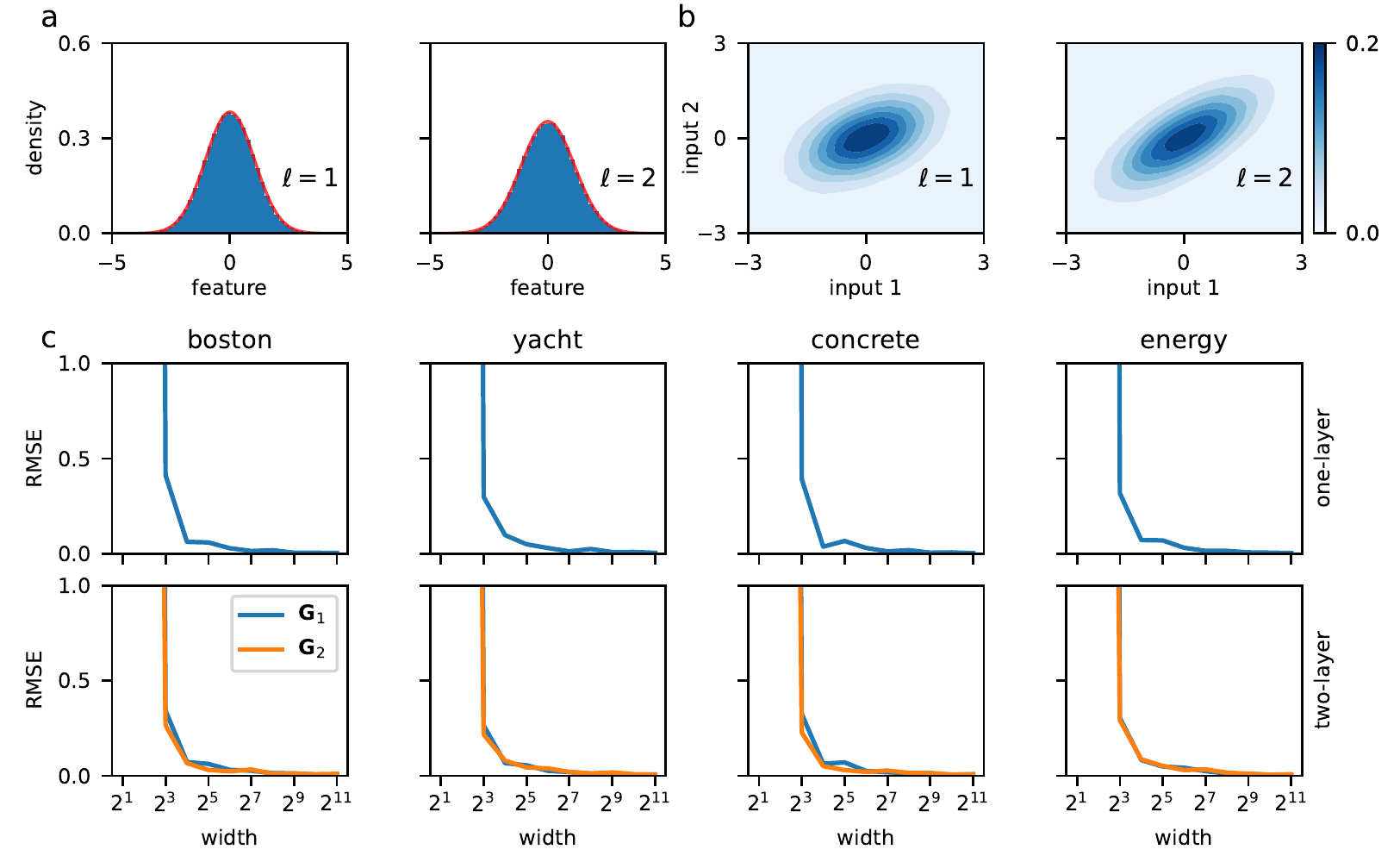}
  \caption{
    Wide DGP posteriors converge to the DKM.
    Here, we trained DGPs with Langevin sampling (see Appendix~\ref{sec:methods}), and compared to a trained DKM.
    \textbf{a} Marginal distribution over features for one input datapoint for a two-layer DGP trained on a subset of yacht.
    We used a width of $N_{1\dotsc L}=1024$ and $\nu_{1\dotsc L}=5$ in all plots to ensure that the data had a strong effect on the learned representations.
    The marginals (blue histogram) are very close to Gaussian (the red line shows the closest fitted Gaussian).
    Remember that the true posterior over features is IID (Eq.~\ref{eq:post_feat}), so each column aggregates the distribution over features (and over 10 parallel chains with 100 samples from each chain) for a single input datapoint.
    \textbf{b} The 2D marginal distributions for the same DGP for two input points (horizontal and vertical axes).
    \textbf{c} Element-wise RMSE (normalized Frobenius distance) between Gram matrices from a trained DKM compared to trained DGPs of increasing width. The DGP Gram matrices converge to the DKM solution as width becomes larger.
    \label{fig:SGLD_marg_1d}
  }
\end{figure*}

\subsection{The \limit{}}
\label{sec:dkm}
In the previous section, we saw that standard infinite width limits eliminate representation learning because as $N\rightarrow \infty$ the log-prior terms, $\log \P{\G_\ell| \G_{\ell-1}}$, in Eq.~\eqref{eq:dgp:joint} dominated the log-likelihood, $\P{\Y| \G_L}$, and the likelihood is the only term that depends on the labels.
We therefore introduce the ``\limit{}'' which retains representation learning.
The \limit{} sends the number of output features, $N_{L+1}$, to infinity as the layer-widths go to infinity,
\begin{align}
  \label{eq:limit}
  N_\ell &= N \; \nu_\ell &&\text{for} & \ell &\in \{1,\dotsc,L+1\} && \text{with} & N &\rightarrow \infty.
\end{align}
Importantly, the \limit{} gives a valid probabilistic model with a well-defined posterior, arising from the prior, (Eq.~\ref{eq:dkp:prior}) and a likelihood which assumes each output channel is IID,
\begin{align}
  \label{eq:deepdkm:like}
  \P{\Yr| \G_L} &= \tprod_{\lambda=1}^{N_{L+1}} \N{\yr_\lambda; \0, \Kg{L} + \sigma^2 \I}.
\end{align}
where $\Yr \in \mathbb{R}^{P \times N_{L+1}}$ is infinite width (Eq.~\ref{eq:limit}) whereas the usual DGP data, $\Y \in \mathbb{R}^{P \times \nu_{L+1}}$, is finite width.
Of course, infinite-width data is unusual if not unheard-of. 
%, but the likelihood can be computed, at least in principle, because we can compute Eq.~\eqref{eq:deepdkm:like} using the output Gram matrix,
%%\begin{align}
%  $\G_{L+1} = \tfrac{1}{N_{L+1}} \Yr \Yr^T \in \mathbb{R}^{P\times P}$,
%%\end{align}
%by analogy with Eq.~\eqref{eq:PFF}.
In practice, real data, $\Y \in \mathbb{R}^{P \times \nu_{L+1}}$, almost always has a finite number of features, $\nu_{L+1}$.
How do we apply the DKM to such data?
The answer is to define $\Yr$ as $N$ copies of the underlying data, $\Y$, i.e.\
%\begin{align}
  $\Yr = \begin{pmatrix} \Y & \dotsm & \Y \end{pmatrix}$.
%\end{align}
As each channel is assumed to be IID (Eq.~\ref{eq:like} and \ref{eq:deepdkm:like}) the likelihood is $N$ times larger,
\begin{align}
  \label{eq:dkm_like}
  \log \P{\Yr| \G_L} &= N \log \P{\Y| \G_L},
\end{align}
%allowing us to retain $\Y$ dependence and representation learning even in an infinitely wide network.
%
The log-posterior in the \limit{} is very similar to the log-posterior in the standard limit (Eq.~\ref{eq:nngp_lim}).
The only difference is that the likelihood, $\log \P{\Yr| \G_L}$ now scales with $N$, so it does not disappear as we take the limit, allowing us to retain representation learning,
\begin{align}
  \label{eq:post_dkm_obj}
  \L&(\G_1,\dotsc,\G_L) \\
  \nonumber
  &= \lim_{N \rightarrow \infty} \tfrac{1}{N} \log \P{\G_1,\dotsc,\G_L| \X, \Yr} + \const,\\
  \nonumber
  &= \log \P{\Y| \G_L} \\
  \nonumber
  &\quad - \tsum_{\ell=1}^L \nu_\ell \KL{\N{\0, \G_\ell}}{\N{\0, \Kg{\ell-1}}}.
\end{align}
Here, we denote the limiting log-posterior as $\L(\G_1,\dotsc,\G_L)$, and this forms the DKM objective.
%We call this quantity the DKM objective, as the \limit{} results in a practical method, called the deep kernel machine, or DKM, which uses this objective and which we discuss later (Sec.~\ref{sec:performance}).
Again, as long as the global maximum of the DKM objective is unique, the posterior is again a point distribution around that maximum (Eq.~\ref{eq:point}).
Of course, the inclusion of the likelihood term means that the global optimum $\G_1^*,\dotsc,\G_L^*$ cannot be computed recursively, but instead we need to optimize, e.g.\ using gradient descent (see Sec.~\ref{sec:performance}).
Unlike in the standard limit (Eq.~\ref{eq:nngp_lim}), it is no longer possible to guarantee uniqueness of the global maximum. 
We can nonetheless say that the posterior converges to a point distribution as long as the global maximum of $\L(\G_1,\dotsc,\G_L)$ is unique, (i.e.\ we can have any number of local maxima, as long as they all lie below the unique global maximum).
We do expect the global maximum to be unique in most practical settings: we know the maximum is unique when the prior dominates (Eq.~\ref{eq:nngp_lim}), in Appendix~\ref{app:unimodality}, we prove uniqueness for linear models, and in Appendix~\ref{sec:unimodality}, we give a number of experiments in nonlinear models in which optimizing from very different initializations found the same global maximum, indicating uniqueness in practical settings.

\subsection{The exact DGP posterior over features is multivariate Gaussian}
\label{sec:exact_dgp_post}

Above, we noted that the DGP posterior over Gram matrices in the \limit{} is a point distribution, as long as the DKM objective has a unique global maximum.
Remarkably, in this setting, the corresponding posterior over features is multivariate Gaussian (see Appendix~\ref{sec:true_post:bnn} for the full derivation), 
\begin{align}
  \label{eq:post}
  \P{\f_\lambda^\ell| \X, \y} &= \N{\f_\lambda^\ell; \0, \G_\ell^*} 
\end{align}
While such a simple result might initially seem remarkable, it should not surprise us too much.
In particular, the prior is Gaussian (Eq.~\ref{eq:deepgp}). 
In addition, in Fig.~\ref{fig:graphical_model} (middle), we saw that the next layer features depend on the current layer features only through the Gram matrices, which are just the raw second moment of the features, Eq.~\eqref{eq:Gdgp}.
Thus, in effect the likelihood only constrains the raw second moments of the features.
Critically, that constraints on the raw second moment are tightly connected to Gaussian distributions: under the MaxEnt framework, a Gaussian distribution arises by maximizing the entropy under constraints on the raw second moment of the features \citep{jaynes2003probability}.
Thus it is entirely plausible that a Gaussian prior combined with a likelihood that ``constrains'' the raw second moment would give rise to Gaussian posteriors (though of course this is not a proof; see Appendix~\ref{sec:true_post:bnn} for the full derivation).

Finally, note that we appear to use $\G_\ell$ or $\G_\ell^*$ in two separate senses: as $\tfrac{1}{N_\ell} \F_\ell \F_\ell^T$ in Eq.~\eqref{eq:Gdgp} and as the posterior covariance in the \limit{} (Eq.~\ref{eq:post}).
In the infinite limit, these two uses are consistent.
In particular, consider the value of $\G_\ell$ defined by Eq.~\eqref{eq:Gdgp} under the posterior,
\begin{align}
  \nonumber
  \G_\ell &= \lim_{N \rightarrow \infty} \tfrac{1}{N_{\ell}} \tsum_{\lambda=1}^{N_{\ell}} \f_\lambda^{\ell} (\f_\lambda^{\ell})^T \\
  &= \E_{\P{\f_\lambda^\ell| \X, \y}}\sqb{\f_\lambda^{\ell} (\f_\lambda^{\ell})^T} = \G_\ell^*.
\end{align}
The second equality arises by noticing that we are computing the average of infinitely many terms, $\f_\lambda^{\ell} (\f_\lambda^{\ell})^T$, which are IID under the true posterior (Eq.~\ref{eq:post}), so we can apply the law of large numbers, and the final expectation arises by computing moments under Eq.~\eqref{eq:post}.

\setlength{\tabcolsep}{4pt}
\begin{table*}
    \centering
    \caption{RMSE for inducing point methods. (Equal) best methods are displayed in bold.  Error bars give two stderrs for a paired test, which uses differences in performance between that method and best method, (so there are no meaningful error bars on the best performing method itself). The MAP objective was numerically unstable on the Boston dataset, and thus did not run to completion.}
    %\resizebox{\columnwidth}{!}{%
    \begin{tabular}{crccc}
        \toprule
        dataset & P & NNGP & MAP & $\LM$\\
        \midrule
        boston & 506 & $\mathbf{4.41 \pm 0.31}$ & --- & $\mathbf{4.35 \pm 0.51}$\\
        concrete & 1,030 & $5.38 \pm 0.098$ & $5.60 \pm 0.15$ & $\mathbf{5.10}$\\
        energy & 768 & $0.83 \pm 0.076$ & $0.73 \pm 0.049$ & $\mathbf{0.47}$\\
        kin8nm & 8,192 & $(7.3 \pm 0.06)\mdot 10^{\text{-}2}$ & $(7.4 \pm 0.05)\mdot 10^{\text{-}2}$ & $\mathbf{6.6\mdot 10^{\text{-}2}}$\\
        naval & 11,934 & $(6.4 \pm 0.6)\mdot 10^{\text{-}4}$ & $(5.4 \pm 0.5)\mdot 10^{\text{-}4}$ & $\mathbf{4.6\mdot 10^{\text{-}4}}$\\
        power & 9,568 & $3.81 \pm 0.091$ & $3.73 \pm 0.14$ & $\mathbf{3.58}$\\
        protein & 45,730 & $4.21 \pm 0.029$ & $4.30 \pm 0.033$ & $\mathbf{4.10}$\\
        wine & 1,599 & $0.68 \pm 0.0084$ & $0.66 \pm 0.0067$ & $\mathbf{0.64}$\\
        yacht & 308 & $0.94 \pm 0.058$ & $1.14 \pm 0.077$ & $\mathbf{0.58}$\\
        \bottomrule
    \end{tabular}%
    %}
    \label{tab:perf}
\end{table*}

\subsection{The \dkmearly{} gives intuition for representation learning}
\label{sec:intuitive}

The form for the \dkmearly{} in Eq.~\eqref{eq:post_dkm_obj} gives a strong intuition for how representation learning occurs in deep networks.
In particular, the likelihood, $\log \P{\Y| \G_L}$, encourages the model to find a representation giving good performance on the training data. 
At the same time, the KL-divergence terms keep the posterior over features, $\N{\0, \G_\ell}$, (Eq.~\ref{eq:post}) close to the prior $\N{\0, \Kg{\ell-1}}$ (Eq.~\ref{eq:deepgp:top:F}).
This encourages the optimized representations, $\G_\ell$, to lie close to their value under the standard infinite-width limit, $\K(\G_{\ell-1})$.
%
%Critically, the KL-divergence terms in the objective act as a regulariser, encouraging $\G_\ell$ to lie close to $\K(\G_{\ell-1})$, their value under the infinite width prior (Sec.~\ref{sec:back:inf}).
%At the same time, the likelihood, $\log \P{\Y| \G_L}$, encourages the model to find a representation giving good performance on the training data.
We could use any form for the likelihood including classification and regression, but to understand how the likelihood interacts with the other KL-divergence terms, it is easiest to consider regression (Eq.~\ref{eq:like}), as this log-likelihood can also be written as a KL-divergence,
\begin{multline}
   \log \P{\Y| \G_{L}} =\\ 
   - \nu_{L+1} \KL{\N{\0, \G_{L+1}}}{\N{\0, \Kg{L} + \sigma^2 \I}} \\+ \const
\end{multline}
Thus, the likelihood encourages $\Kg{L} + \sigma^2 \I$ to be close to the covariance of the data, $\G_{L+1} = \tfrac{1}{\Ny} \Y \Y^T$, while the DGP prior terms encourage all $\G_\ell$ to lie close to $\K(\G_{\ell-1})$.
In combination, we would expect the optimal Gram matrices to ``interpolate'' between the input kernel, $\G_0 = \tfrac{1}{\nu_0} \X \X^T$ and the output kernel, $\G_{L+1}$. 

To make the notion of interpolation explicit, we consider $\sigma^2=0$ with a linear kernel, $\K(\G_{\ell-1}) = \G_{\ell-1}$, so named because it corresponds to a linear neural network layer.
With this kernel and with all $\nu_\ell = \nu$, there is an analytic solution for the (unique) optimum of the DKM objective (Appendix~\ref{sec:unimodality_linear}),
\begin{align}
  \G^*_\ell &= \G_0 \b{\G_0^{-1} \G_{L+1}}^{\ell/(L+1)},
\end{align}
which explicitly geometrically interpolates between $\G_0$ and $\G_{L+1}$.
Of course, this discussion was primarily for DGPs, but the exact same intuitions hold for BNNs, in that maximizing the \dkmearly{} finds a sequence of Gram matrices, $\G^*_1,\dotsc,\G^*_L$ that interpolate between the input kernel, $\G_0$ and the output kernel, $\G_{L+1}$.
The only difference is in details of $\P{\G_\ell| \G_{\ell-1}}$, and specifically as slight differences in the KL-divergence terms (see below).

\subsection{The DKM objective mirrors representation learning in finite networks}
\label{sec:finite}
%Here, we confirm that these insights apply to representation learning in finite networks.
%We begin by noting a theoretical result that we give in full in the Appendix: that the representations learned by MAP inference over DGP features (rather than Gram matrices as we consider in the main text) is exactly the same solution for any $N$ (finite or infinite, as long as $N\geq P$; Appendix~\ref{sec:MAP}).
%Second, in Appendix~\ref{sec:langevin}, we find that Langevin sampling for the DKM posterior are deterministic, equivalent to preconditioned gradient descent on the DKM objective, and equivalent to the expected Langevin updates to the Gram matrices for a finite-width DGP.

Here, we confirm that the optimizing DKM objective for an infinite network matches doing inference in wide but finite-width networks using Langevin sampling (see Appendix~\ref{sec:methods} for details).

We began by looking at DGPs, and confirming that the posterior marginals are Gaussian (Eq.~\ref{eq:post}; Fig.~\ref{fig:SGLD_marg_1d}ab). 
Then, we confirmed that the representations match closely for infinite-width DKMs (Fig.~\ref{fig:twolayer_yacht} top and bottom rows) and finite-width DGPs (Fig.~\ref{fig:twolayer_yacht} middle two rows), both at initialization (Fig.~\ref{fig:twolayer_yacht} top two rows) and after training to convergence (Fig.~\ref{fig:twolayer_yacht} bottom two rows).
Note that the first column, $\K_0$ is a squared exponential kernel applied to the input data, and $\G_3 = \y \y^T$ is the output Gram matrix (in this case, there is only one output feature).
%\begin{figure*}[t]
%    \centering
%    \includegraphics[width=\textwidth]{figures/SGLD_dist_50_0.1.pdf}
%    \caption{The DGP and DKM match more closely as the DGP becomes wider. We trained one layer (top row) and two layer (bottom row) DGPs and DKMs with ReLU kernels on the first 50 datapoints of each dataset, and show the RMSE between elements of the Gram matrices, corresponding to normalized Frobenius norms between those matrices.
%    The datasets are from \citep{gal2016dropout}.
%    \label{fig:SGLD_dist}
%    }
%\end{figure*}

To confirm that the match improves as the DGP gets wider, we considered the RMSE between elements of the Gram matrices for networks of different widths (x-axis) for different UCI datasets (columns) and different numbers of layers (top row is one-layer, bottom row is two-layers; Fig.~\ref{fig:SGLD_marg_1d}c).
In most cases, we found a good match as long as the width was at least 128, which is around the width of typical fully connected neural network, but is a little larger than typical DGP widths \citep[e.g.\ ][]{damianou2013deep,salimbeni2017doubly}.

% \begin{figure*}[t]
%     \centering
%     \includegraphics[width=\textwidth]{figures/layers_2_DWP_mean_data_yacht_-1_var_True_0.01_kernel_exp_lr_1e-03_init_scale_a_0.5_b_3.0.pdf}
%     \caption{A two-layer DKM with squared exponential kernel trained on yacht converges to the same solution, despite very different initializations.  The first column is the spectral norm of the Gram matrices, the middle three columns are three random projections (with unit norm) and the last column is the average Frobenius norm between all pairs of initializations.
%     The first row is for $\G_1$ and the second row is for $\G_2$.
%     \label{fig:two_exp_yacht}
%     }
% \end{figure*}

% \begin{figure*}[t]
%     \centering
%     \includegraphics[width=\textwidth]{figures/layers_DWP_1_DGP_full_data_yacht_-1_var_True_0.01_kernel_exp_lr_1e-03_init_scale_a_0.5_b_3.0.pdf}
%     \caption{A One-layer deep kernel machine with squared-exponential kernel trained on yacht using the MAP objective, where the same initialization scheme finds two solutions.}
%     \label{fig:one_DGP_full}
% \end{figure*}

\subsection{The sparse deep kernel machine as a deep generalisation of kernel methods}
\label{sec:performance}
DGPs in the Bayesian representation learning limit constitute a deep generalisation of kernel methods, with a very flexible learned kernel, which we call the deep kernel machine (DKM; which was introduced earlier just in the context of the objective).
Here, we design a sparse DKM, inspired by sparse methods for DGPs \citep{damianou2013deep,salimbeni2017doubly} (Appendix~\ref{sec:app:pred}).
The sparse DKM scales linearly in the number of datapoints, $P$, as opposed to cubic scaling of the plain DKM (similar to the cubic scaling in most naive kernel methods).

We compared DKMs (Eq.~\ref{eq:post_dkm_obj}), MAP over features (Sec.~\ref{sec:MAP}) for DGPs, and an NNGP (specifically, an infinite width DGP).
The NNGP mirrored the structure of the deep kernel machine but where the only flexibility comes from the hyperparameters. %, $w_1^\ell$, $w_2^\ell$ (see below) and $\sigma^2$.
Formally, this model can be obtained by setting, $\G_\ell = \K(\G_{\ell-1})$ and is denoted ``Kernel Hyper'' in Table~\ref{tab:perf}.
We applied these methods to UCI datasets \citep{gal2016dropout} using a two hidden layer architecture, with a kernel inspired by DGP skip-connections, $\K(\G_\ell) = w^\ell_1 \G_\ell + w^\ell_2 \K_\text{sqexp}(\G_\ell)$.
Here, $w^\ell_1$, $w^\ell_2$ and $\sigma$ are hyperparameters, and $\K_\text{sqexp}(\G_\ell)$ is a squared-exponential kernel.
%Second, we considered an objective given by MAP inference over features, $\F_\ell$ (Eq.~\ref{eq:PFF}), denoted ``MAP''.
%Finally, we considered the deep kernel machine objective, denoted $\LM$.
%We used the fast solver introduced in the companion paper \citep{dkm2}.

We used 300 inducing points fixed to a random subset of the training data and not optimised during training.
We used the Adam optimizer with a learning rate of 0.001, full-batch gradients and 5000 iterations for smaller datasets and 1000 iterations for larger datasets (kin8nm, naval and protein).

We found that the deep kernel machine objective gave better performance than MAP, or the hyperparameter optimization baseline (Tab.~\ref{tab:perf}).
Note that these numbers are not directly comparable to those in the deep GP literature \citep{salimbeni2017doubly}, as deep GPs have a full posterior so offer excellent protection against overfitting, while DKMs give only a point estimate.

% The source code for our experiments has been made publicly available\footnote{\url{https://anonymous.4open.science/r/dkm_submission-1F43}}.

%\setlength{\tabcolsep}{4pt}
%\begin{table}
%  \centering
%  \caption{RMSE after 10 solver iterations.  (Equal) best methods are displayed in bold.  Error bars give two stderrs for a paired tests, which uses differences in performance between that method and best method, (so there are no meaningful error bars on the best performing method itself).  
%  \label{tab:perf}
%  }
%  \begin{tabular}{ccccc}
%    \toprule
%    dataset & kernel hyper & MAP & $\LM$ \\
%    \midrule
%    boston & $2.79 \pm 0.10$ & $2.80 \pm 0.08$ & $\mathbf{2.67}$\\
%    concrete & $5.47 \pm 0.10$ & $5.43 \pm 0.11$ & $\mathbf{5.17}$\\
%    energy & $0.54 \pm 0.01$ & $\mathbf{0.48 \pm 0.02}$ & $\mathbf{0.47}$\\
%    yacht & $0.80 \pm 0.07$ & $\mathbf{0.55}$ & $\mathbf{0.56 \pm 0.04}$\\
%    \bottomrule
%  \end{tabular}
%\end{table}

% Inducing point results:

\section{Conclusion}
We introduced the \limit{}, a new infinite-width limit for BNNs and DGPs that retains representation learning.
Representation learning in this limit is described by the intuitive DKM objective, which is composed of a log-likelihood describing performance on the task (e.g.\ classification or regression) and a sum of KL-divergences keeping representations at every layer close to those under the infinite-width prior.
For DGPs, the exact posteriors are IID across features and are multivariate Gaussian, with covariances given by optimizing the DKM objective.
Empirically, we found that the distribution over features and representations matched those in wide by finite DGPs.
We argued that DGPs in the \limit{} form a new class of practical deep kernel method: DKMs.
We introduce sparse DKMs, which scale linearly in the number of datapoints.
Finally, we give the extension for BNNs where the exact posteriors are intractable and so must be approximated (Appendix~\ref{sec:exact_post}).

\bibliographystyle{\sty}
\bibliography{ref}

\newpage
\appendix
\onecolumn

%{\LARGE\bf Supplementary Information for ``A theory of representation learning in deep neural networks gives a deep generalisation of kernel methods''}
%\renewcommand{\theequation}{Appendix~\arabic{equation}}

\section{Bayesian neural network extension}
\label{sec:bnn}
Consider a neural network of the form,
\begin{subequations}
\label{eq:nn}
\begin{align}
  \F_1 &= \X \W_0\\
  \F_{\ell} &= \phi(\F_{\ell-1}) \W_{\ell-1} \quad \text{for } \ell \in \{2,\dotsc,L+1\}\\
  W^\ell_{\lambda \mu} &\sim \N{0, \tfrac{1}{N_\ell}}\quad \quad
  W^0_{\lambda \mu} \sim \N{0, \tfrac{1}{\nu_0}} 
\end{align}
\end{subequations}
where $\W_0\in\mathbb{R}^{\nu_0 \times N_1}$, $\W_\ell \in \mathbb{R}^{N_\ell \times N_{\ell+1}}$ and $\W_{L+1}\in\mathbb{R}^{N_L\times \nu_{L+1}}$ are weight matrices with independent Gaussian priors and $\phi$ is the usual pointwise nonlinearity.

In principle, we could integrate out the distribution over $\W_\ell$ to find $\P{\F_\ell| \F_{\ell-1}}$
\begin{align}
  \P{\F_\ell| \F_{\ell-1}} &= \int d\W_\ell \P{\W_\ell} \delta\b{\F_\ell -\phi(\F_{\ell-1}) \W_{\ell-1}},
\end{align}
where $\delta$ is the Dirac delta.
In practice, it is much easier to note that conditioned on $\F_{\ell-1}$, the random variables interest, $\F_\ell$ are a linear combination of Gaussian distributed random variables, $\W_\ell$. 
Thus, $\F_\ell$ are themselves Gaussian, and this Gaussian is completely characterised by its mean and variance.
We begin by writing the feature vectors, $\f_\lambda^\ell$ in terms of weight vectors, $\w_\lambda^\ell$,
\begin{align}
  \f_\lambda^\ell &= \phi(\F_{\ell-1}) \w_\lambda^\ell.
\end{align}
As the prior over weight vectors is IID, the prior over features (conditioned on $\F_{\ell-1}$) is also IID,
\begin{align}
  \P{\W} &= \prod_{\lambda=1}^{N_\ell} \P{\w_\lambda^\ell} = \prod_{\lambda=1}^{N_\ell} \N{\w_\lambda^\ell; \0, \tfrac{1}{N_{\ell-1}} \I},\\
  \P{\F_\ell| \F_{\ell-1}} &= \prod_{\lambda=1}^{N_\ell} \P{\f_\lambda^\ell| \F_{\ell-1}}.
\end{align}
The mean of $\f_\lambda^\ell$ conditioned on $\F_{\ell-1}$ is $\0$,
\begin{align}
  \E\sqb{\f_\lambda^\ell| \F_{\ell-1}} &= \E\sqb{\phi(\F_{\ell-1}) \w_\lambda^\ell| \F_{\ell-1}} = \phi(\F_{\ell-1}) \E\sqb{\w_\lambda^\ell| \F_{\ell-1}} = \phi(\F_{\ell-1}) \E\sqb{\w_\lambda^\ell} = \0.
\end{align}
The covariance of $\f_\lambda^\ell$ conditioned on $\F_{\ell-1}$ is,
\begin{align}
  \nonumber
  \E\sqb{\f_\lambda^\ell \b{\f_\lambda^\ell}^T| \F_{\ell-1}} &= \E\sqb{\phi(\F_{\ell-1}) \w_\lambda^\ell \b{\phi(\F_{\ell-1}) \w_\lambda^\ell}^T| \F_{\ell-1}}\\
  \nonumber
  &= \phi(\F_{\ell-1}) \E\sqb{\w_\lambda^\ell(\w_\lambda^\ell)^T}\phi^T(\F_{\ell-1}) \\
  &= \tfrac{1}{N_{\ell-1}} \phi(\F_{\ell-1}) \phi^T(\F_{\ell-1})
\end{align}
This mean and variance imply that Eq.~\eqref{eq:deepgp} captures the BNN prior, as long as we choose $\Kbnn(\cdot)$ and $\Gbnn(\cdot)$ such that,
\begin{align}
  \label{eq:bnn:kernel} 
  \Kbnn(\Gbnn(\F_{\ell-1})) &= \tfrac{1}{N_{\ell-1}} \tsum_{\lambda=1}^{N_{\ell-1}} \phi(\f_\lambda^{\ell-1})\phi^T(\f_\lambda^{\ell-1}),%\\
  %\label{eq:dgp:kernel} 
  %K_{\text{isotropic}; ij}(\Gdgp(\F_{\ell-1})) &= k(R_{ij}(\Gdgp(\F_{\ell-1}))).
\end{align}
Specifically, we choose the kernel function, $\Kbnn(\cdot)$ to be the identity function, and $\Gbnn(\cdot)$ to be the same outer product as in the main text for DGPs (Eq.~\ref{eq:Gdgp}), except where we have applied the NN nonlinearity,
\begin{align}
  \label{eq:Kbnn}
  \Kbnn(\G_{\ell-1}) &= \G_{\ell-1},\\
  \label{eq:Gbnn}
  \Gbnn(\F_{\ell-1}) &= \tfrac{1}{N_{\ell-1}} \tsum_{\lambda=1}^{N_{\ell-1}} \phi(\f_\lambda^{\ell-1})\phi^T(\f_\lambda^{\ell-1}).
\end{align}
This form retains the average-outer-product form for $\Gbnn(\cdot)$, which is important for our derivations.

Now, Eq.~\eqref{eq:post_dkm_obj} only gave the DKM objective for DGPs. 
To get a more general form, we need to consider the implied posteriors over features.
This posterior is IID over features (Appendix~\ref{sec:exact_post}), and for DGPs, it is multivariate Gaussian (Appendix~\ref{sec:mvg_dgp_post}),
\begin{align}
  \label{eq:post_feat}
  \P{\F_\ell| \G_{\ell-1}, \G_\ell} &= \tprod_{\lambda=1}^{N^\ell} \P{\f_\lambda^\ell| \G_{\ell-1}, \G_{\ell}} \underset{\text{\tiny for DGPs}}{=} \tprod_{\lambda=1}^{N^\ell} \N{\f_\lambda^\ell; \0, \G_\ell}.
\end{align}
%If $\L(\G_1,\dotsc,\G_L)$ has a single global optimum, so the posterior over Gram matrices converges to a point at $\G_1^*,\dotsc,\G_L^*$, the full posterior over features becomes,
%\begin{align}
%  \label{eq:post}
%  \P{\F_1,\dotsc,\F_L| \X, \Y} &= \tprod_{\ell=1}^L \tprod_{\lambda=1}^{N^\ell} \P{\f_\lambda^\ell| \G^*_{\ell-1}, \G^*_{\ell}} \underset{\text{\tiny for DGPs}}{=} \tprod_{\ell=1}^L \tprod_{\lambda=1}^{N^\ell} \N{\f_\lambda^\ell; \0, \G_\ell^*}.
%\end{align}
Now, we can see that Eq.~\eqref{eq:post_dkm_obj} is a specific example of a general expression.
In particular, note that the distribution on the left of the KL-divergence in Eq.~\eqref{eq:post_dkm_obj} is the DGP posterior over features (Eq.~\ref{eq:post_feat}).
Thus, the DKM objective can alternatively be written,
\begin{align}
  \label{eq:MAP}
  \L(\G_1,\dotsc,\G_L) &= \log \P{\Y| \G_L} - \tsum_{\ell=1}^L \nu_\ell \KL{\P{\f_\lambda^\ell| \G_{\ell-1}, \G_\ell}}{\N{\0, \K(\G_{\ell-1})}},
\end{align}
and this form holds for both BNNs and DGPs (Appendix~\ref{app:P(GG)}).
%Note that $\P{\f_\lambda^\ell| \G_{\ell-1}, \G_\ell}$ makes sense because the posterior for all models in our class is IID over features, $\f_1^\ell,\dotsc,\f_{N_\ell}^\ell$, but only Gaussian for DGPs,
%\begin{align}
%  \label{eq:iid_post}
%  \P{\F_\ell| \G_{\ell-1}, \G_\ell} &= \tprod_{\lambda=1}^{N_\ell} \P{\f_\lambda^\ell| \G_{\ell-1}, \G_\ell}.
%\end{align}
As in DGPs, the log-posterior is $N$ times $\L(\G_1,\dotsc,\G_L)$ (Eq.~\ref{eq:post_dkm_obj}), so as $N$ is taken to infinity, the posterior for all models becomes a point distribution (Eq.~\ref{eq:point}) if $\L(\G_1,\dotsc,\G_L)$ has a unique global maximum.

\begin{figure}[t!]
  \includegraphics[width=\textwidth]{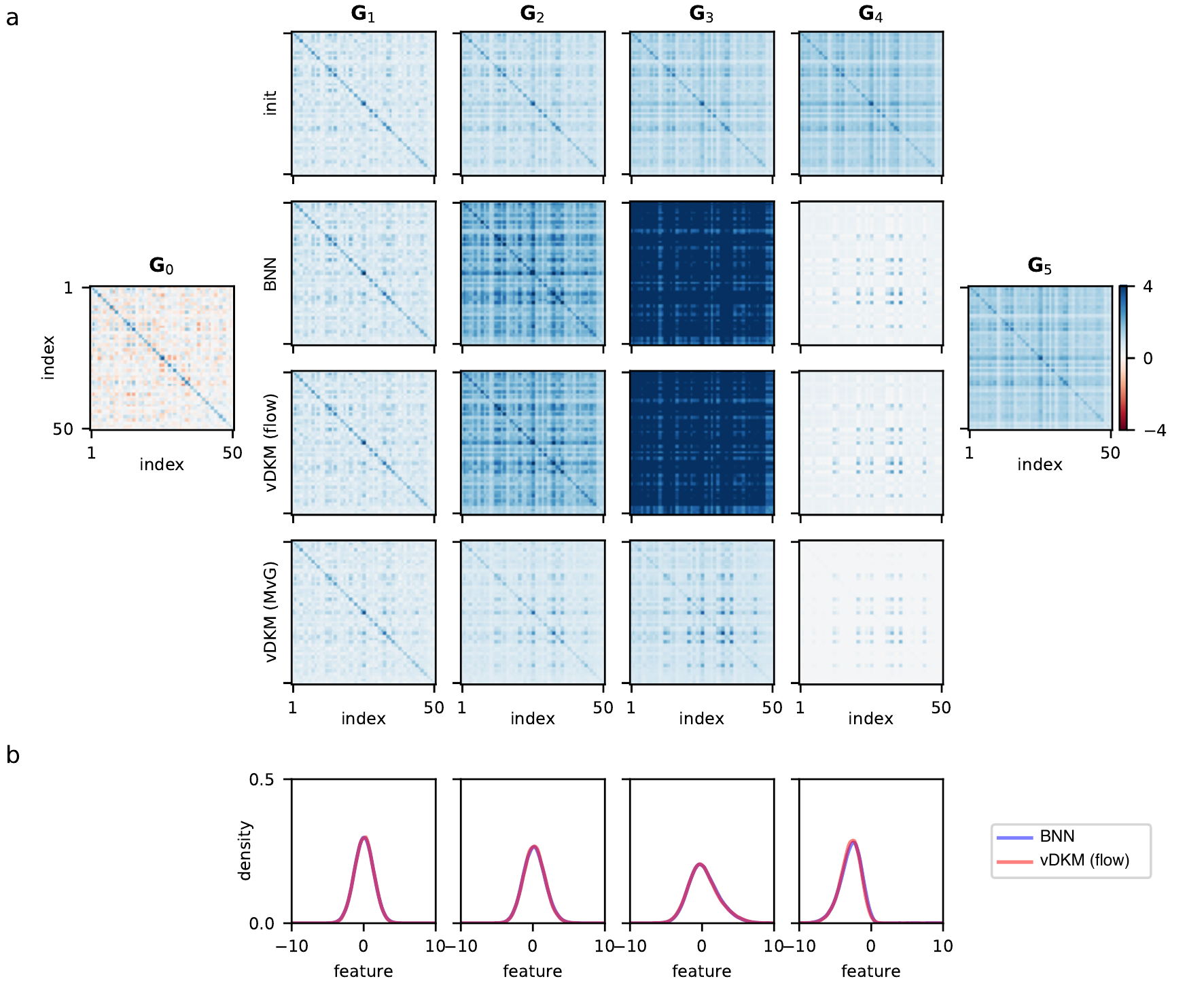}
  \caption{
    The variational DKM closely matches the BNN true posterior obtained with Langevin sampling. \textbf{a} Comparison of Gram matrices. The first two rows show Gram matrices for BNN, with the first row being a random initialization, and the second row being the posterior.  The last two rows show the Gram matrices from variational DKMs with a flow approximate posterior (third row) and a multivariate Gaussian approximate posterior (fourth row). In optimizing the variational DKM, we used Eq.~\eqref{eq:Kt} with $2^{16}$ Monte-Carlo samples. The Gram matrices for the flow posterior (third row) closely match those from the BNN posterior (second row), while those for a multivariate Gaussian approximate posterior (fourth row) do not match. \textbf{b} Marginal distributions over features at each layer for one input datapoint estimated using kernel density estimation. 
    The note that the BNN (blue line) marginals are non-Gaussian, but the variational DKM with a flow posterior (red line) is capable of capturing this non-Gaussianity.
    \label{fig:flow_4}
  }
\end{figure}

In practice, the true posteriors required to evaluate Eq.~\eqref{eq:MAP} are intractable for BNNs, raising the question of how to develop accurate approximations for BNNs.
We develop a variational DKM (vDKM) by taking inspiration from variational inference \citep{jordan1999introduction,blei2017variational} (Appendix~\ref{sec:true_post:bnn:approx}).
Of course, variational inference is usually impossible in infinite width models, because it is impossible to work with infinitely large latent variables.
Our key insight is that as the true posterior factorises across features (Appendix~\ref{sec:exact_post}), we can work with the approximate posterior over only a single feature vector, $\Q[\theta_\ell]{\f_\lambda^\ell}$, where $\theta_\ell$ are the parameters and $\f_\lambda^\ell \in \mathbb{R}^P$ is finite.
This approach allows us to define a vDKM objective, which bounds the true DKM objective,
\begin{align}
%\end{align}
%\begin{align}
  \label{eq:bound}
  \L(\Gt{1},\dotsc,\Gt{L}) &\geq \L_\text{V}(\theta_1,\dotsc,\theta_L),\\
  \nonumber
  \L_\text{V}(\theta_1,\dotsc,\theta_L) &= \log \P{\Y| \Gt{L}} - \tsum_{\ell=1}^L \nu_\ell \KL{\Q[\theta_\ell]{\f_\lambda^\ell}}{\N{\0, \K(\Gt{\ell-1})}}
\end{align}
with equality when the approximate posteriors, $\Q[\theta_\ell]{\f_\lambda^\ell}$, equal the true posteriors, $\P{\f_\lambda^\ell| \G_{\ell-1}, \G_\ell}$.
The only subtlety here is that it is practically difficult to design flexible approximate posteriors $\Q[\theta_\ell]{\f_\lambda^\ell}$ where we explicitly specify and optimize the Gram matrices.
Instead we optimize general approximate posterior parameters, $\theta$, and compute the implied Gram matrices,
\begin{align}
  \label{eq:Kt}
  \Gt{\ell} &= \tfrac{1}{N_\ell} \lim_{N \rightarrow \infty} \tsum_{\lambda=1}^{N_\ell} \phi(\f_\lambda^\ell)\phi^T(\f_\lambda^\ell) =  \E_{\Q[\theta_\ell]{\f_\lambda^\ell}}\sqb{\phi(\f_\lambda^\ell)\phi^T(\f_\lambda^\ell)}.
\end{align}
where $\f_\lambda^\ell$ are sampled from $\Q[\theta_\ell]{\f_\lambda^\ell}$, and the second equality arises from the law of large numbers.
We can compute the Gram matrix analytically in simple cases (such as a multivariate Gaussian), but in general we can always estimate the Gram matrix using a Monte-Carlo estimate of Eq.~\eqref{eq:Kt}.

Finally, we checked that the vDKM objective closely matched the posterior under neural networks.
This is a bit more involved, as the marginal distributions over features are no longer Gaussian (Fig.~\ref{fig:flow_4}b).
To capture these non-Gaussian marginals, we used a simple normalizing flow. 
In particular, we first sampled $\z_\lambda^\ell \sim \N{\m_\ell, \S_\ell}$ from a multivariate Gaussian with a learned mean, $\m_\ell$, and covariance, $\S_\ell$ then we obtained features, $\f_\lambda^\ell = f(\z_\lambda^\ell)$, by passing $\z_\lambda^\ell$ through $f$, a learned pointwise function parameterised as in a neural spline flow \citep{durkan2019neural}.
%\begin{align}
%  \z_\lambda^\ell &\sim \N{\m_\ell, \S_\ell} & \f_\lambda^\ell &= \f(\z_\lambda^\ell)
%\end{align}
The resulting distribution is a high-dimensional Gaussian copula \citep[e.g.][] {cai2018high}. As shown in Fig.~\ref{fig:flow_4}, vDKM with multivariate Gaussian (MvG) approximate posterior cannot match the Gram matrices learned by BNN (Fig.~\ref{fig:flow_4}a), while vDKM with flow is able to capture the non-Gaussian marginals (Fig.~\ref{fig:flow_4}b) and thus match the learned Gram matrices with BNN.

\section{General likelihoods that depend only on Gram matrices}
\label{app:like}
We consider likelihoods which depend only on the top-layer Gram matrix, $\G_L$,
\begin{align}
  \P{\Y| \G_L} &= \int d\F_{L+1} \P{\Y| \F_{L+1}} \P{\F_{L+1}| \G_L}
  \intertext{where,}
  \P{\F_{L+1}| \G_L} &= \prod_{\lambda=1}^{N_{L+1}} \N{\f_\lambda^{L+1}; \0, \K(\G_L)}.
\end{align}
This family of likelihoods captures regression,
\begin{align}
  \P{\y_\lambda| \f_\lambda^{L+1}} &= \N{\y_\lambda^{L+1}; \f_\lambda^{L+1}, \sigma^2 \I},
\end{align}
(which is equivalent to the model used in the main text Eq.~\ref{eq:deepgp:like}) and e.g.\ classification, 
\begin{align}
  \P{\y| \F} &= \text{Categorical}\b{\y; \text{softmax}\b{\F_{L+1}}},
\end{align}
among many others.

\section{Asymptotic behaviour of $\log P(\G_l \mid \G_{l-1})$}
\label{app:asymp_wishart}
\begin{figure*}
    \centering
    \includegraphics[width=.35\textwidth]{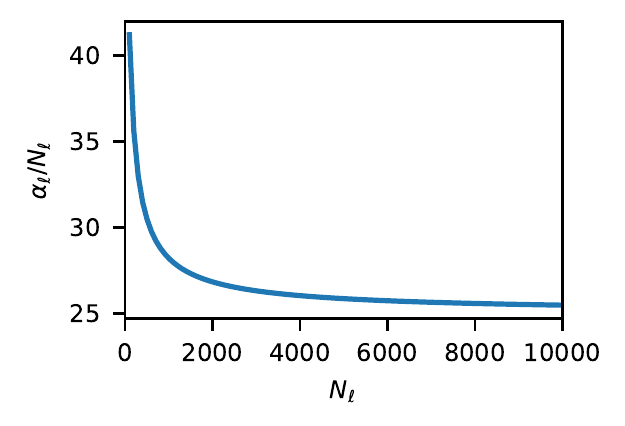}
    \caption{Convergence of $\tfrac{\alpha_\ell}{N_\ell}$, with $P=50$ and $N_\ell\geq 100$.}
    \label{fig:const}
\end{figure*}

To show that $N^{-1} \log P(\G_\ell\mid \G_{\ell - 1})$ has a valid limit, we need to show that $\lim_{N\rightarrow\infty} \alpha_\ell/N$ exists (where $\alpha_\ell$ is defined in Eq.~\ref{eq:wishart_const}).
This can be done using the asymptotic expansion of the Gamma function, $\Gamma_1$, and noting that we can write the multivariate Gamma function, $\Gamma_P$, in terms of $\Gamma_1$,
\begin{align}
\log \Gamma_1(x) &= x\log x - x + \tfrac{1}{2}\log 2\pi x + O(x^{-1})\;\text{as}\;x\rightarrow\infty,\\
\Gamma_P (x) &= \pi^{P(P-1)/4}\prod_{j=1}^P\Gamma_1\left(x + \tfrac{1 - j}{2}\right).
\end{align}
It follows that
\begin{align}
\tfrac{1}{N}\log \Gamma_P\left(\tfrac{N_l}{2}\right) &= O(N^{-1}) + N^{-1}\sum_{j=1}^P \log \Gamma_1\left(\tfrac{N_l+ 1 - j}{2}\right)\\
&= o(1) + N^{-1}\sum_{j=1}^P\left\{\left(\tfrac{N_l+1-j}{2}\right)\log\left(\tfrac{N_l + 1 - j}{2} \right)- \tfrac{N_l + 1 - j}{2} + \tfrac{1}{2}\log \pi N_l + O(N^{-1})\right\}\\
&= o(1) + \const + \sum_{j=1}^P\tfrac{\nu_\ell}{2}\log\left({N_\ell + 1 - j}\right),
\end{align}
which gives
\begin{align}
\tfrac{\alpha_\ell}{N} = \const + 
\tfrac{\nu_\ell}{2}\sum_{j=1}^P \log\left(\tfrac{N_\ell}{N_\ell + 1 - j}\right)
 + o(1) = \const + o(1)
\end{align}
as desired. Figure~\ref{fig:const} confirms the convergence of $\tfrac{\alpha_\ell}{N_\ell}=\tfrac{1}{ \nu_\ell}\tfrac{\alpha_\ell}{N}$ numerically.
\section{Weak convergence}
\label{app:dgp:weak}
Here, we give a formal argument for weak convergence of the DGP posterior over Gram matrices to a point distribution in the limit as $N \rightarrow \infty$,
\begin{align}
  \P[N]{\G_1,\dotsc,\G_L| \X, \Yr} &\overset{d}{\rightarrow} \prod_{\ell=1}^L \delta(\G_\ell - \G_\ell^*)
\end{align}
where we have included $N$ in the subscript of the probability distribution as a reminder that this distribution depends on the width.
By the Portmanteau theorem, weak convergence is established if all expectations of bounded continuous functions, $f$, converge
\begin{align}
  \lim_{N \rightarrow \infty} \E_{\P[N]{\G_1,\dotsc,\G_L| \X, \Yr}}\sqb{f(\G_1,\dotsc,\G_L)} = f(\G_1^*,\dotsc,\G_L^*).
\end{align}
To show this in a reasonably general setting (which the DGP posterior is a special case of), we consider an unnormalized probability density of the form $h(g)e^{N\L(g)}$,  and compute the moment as,
\begin{align}
  \E\sqb{f(g)} &= \frac{\int_\mathcal{G} dg \; f(g) h(g) e^{N\L(g)}}{\int_\mathcal{G} dg \; h(g) e^{N\L(g)}}
\end{align}
where $g = (\G_1,\dotsc,\G_L)$ is all $L$ positive semi-definite matrices, $\G_\ell$.
Thus, $g \in \mathcal{G}$, where $\mathcal{G}$ is a convex set.

We consider the superlevel set $A(\Delta) = \{g| \L(g) \geq \L(g^*)-\Delta\}$, where $g^*$ is the unique global optimum.
%We take the global optimum, $g^*$ as being in a connected region, $A$, of non-zero measure such that all $g \in A(\Delta)$ are higher than all $g \in \mathcal{G}\setminus A(\Delta)$.
We select out a small region, $A(\Delta)$, surrounding the global maximum, and compute the integral as,
\begin{align}
  \E\sqb{f(g)} &= \frac{\int_{A(\Delta)} dg \; f(g) h(g) e^{N\L(g)} + \int_{\mathcal{G}\setminus A(\Delta)} dg \; f(g) h(g) e^{N\L(g)}}{\int_{A(\Delta)} dg \; h(g) e^{N\L(g)} + \int_{\mathcal{G}\setminus A(\Delta)} dg \; h(g) e^{N\L(g)}}\\
  \intertext{And divide the numerator and denominator by $\int_{A(\Delta)} dg \; h(g) e^{N\L(g)}$,}
  \label{eq:weak_ratios}
  \E\sqb{f(g)} &= \frac{\frac{\int_{A(\Delta)} dg \; f(g) h(g) e^{N\L(g)}}{\int_{A(\Delta)} dg \; h(g) e^{N\L(g)}} + \frac{\int_{\mathcal{G}\setminus A(\Delta)} dg \; f(g) h(g) e^{N\L(g)}}{\int_{A(\Delta)} dg \; h(g) e^{N\L(g)}}}{1 + \frac{\int_{\mathcal{G}\setminus A(\Delta)} dg \; h(g) e^{N\L(g)}}{\int_{A(\Delta)} dg \; h(g) e^{N\L(g)}}}
\end{align}
Now, we deal with each term separately.  The ratio in the denominator can be lower-bounded by zero, and upper bounded by considering a smaller superlevel set, $A(\Delta/2)$, in the denominator,
\begin{align}
  \nonumber
  0 \leq  \frac{\int_{\mathcal{G}\setminus A(\Delta)} dg \; h(g) e^{N\L(g)}}{\int_{A(\Delta)} dg \; h(g) e^{N\L(g)}} 
  &\leq  \frac{\int_{\mathcal{G}\setminus A(\Delta)} dg \; h(g) e^{N\L(g)}}{\int_{A(\Delta/2)} dg \; h(g) e^{N\L(g)}}\\
  \nonumber
  &\leq \frac{e^{N(\L(g^*)-\Delta)}\int_{\mathcal{G}\setminus A(\Delta)} dg \; h(g) }{e^{N(\L(g^*)-\Delta/2)}\int_{A(\Delta/2)} dg \; h(g) }\\
  \label{eq:weak_denom}
  &=\frac{\int_{\mathcal{G}\setminus A(\Delta)} dg \; h(g)}{\int_{A(\Delta/2)} dg \; h(g)} e^{- N \Delta/2}
\end{align}
The upper bound converges to zero (as $h(g)$ is independent of $N$), and therefore by the sandwich theorem the ratio of interest also tends to zero.

The second ratio in the numerator can be rewritten as,
\begin{align}
  \label{eq:weak_nom1}
  \frac{\int_{\mathcal{G}\setminus A(\Delta)} dg \; f(g) h(g) e^{N\L(g)}}{\int_{A(\Delta)} dg \; h(g) e^{N\L(g)}} &=  
  \frac{\int_{\mathcal{G}\setminus A(\Delta)} dg \; f(g) h(g) e^{N\L(g)}}{\int_{\mathcal{G}\setminus A(\Delta)} dg \; h(g) e^{N\L(g)}}
  \frac{\int_{\mathcal{G}\setminus A(\Delta)} dg \; h(g) e^{N\L(g)}}{\int_{A(\Delta)} dg \; h(g) e^{N\L(g)}}
\end{align}
The first term here is an expectation of a bounded function, $f(g)$, so is bounded, while second term converges to zero in the limit (by the previous result).

Finally, we consider the first ratio in the numerator,
\begin{align}
  \label{eq:first_ratio}
  \frac{\int_{A(\Delta)} dg \; f(g) h(g) e^{N\L(g)}}{\int_{A(\Delta)} dg \; h(g) e^{N\L(g)}}
\end{align}
which can be understood as an expectation over $f(g)$ in the region $A(\Delta)$.
As $f$ is continuous, for any $\epsilon > 0$, we can find a $\delta > 0$ such that for all $g$ with $|g^* - g| < \delta$, we have
\begin{align}
  \label{eq:cont}
  f(g^*) - \epsilon &< f(g) < f(g^*) + \epsilon. 
\end{align}
Further, because the continuous function, $\L(g)$, has a unique global optimum, $g^*$, for every $\delta > 0$ we are always able to find a $\Delta > 0$ such that all points $g \in A(\Delta)$ are within $\delta$ of $g^*$ i.e. $\abs{g^* - g} < \delta$.
Thus combining the previous two facts, given an $\epsilon$, we are always able to find a $\delta$ such that Eq.~\ref{eq:cont} holds for all $g$ with $\abs{g^* - g} < \delta$, and given a $\delta$ we are always able to find a $\Delta$ such that all $g \in A(\Delta)$ have $\abs{g^* - g} < \delta$. Hence for every $\epsilon >0$ we can find a $\Delta > 0$ such that Eq.~\ref{eq:cont} holds for all $g \in A(\Delta)$.
Choosing the appropriate $\epsilon$-dependent $\Delta$ and substituting Eq.~\ref{eq:cont} into Eq.~\ref{eq:first_ratio}, $\epsilon$ also bounds the error in the expectation,
\begin{align}
  \label{eq:weak_nom2}
  f(g^*) - \epsilon < \frac{\int_{A(\Delta)} dg \; f(g) h(g) e^{N\L(g)}}{\int_{A(\Delta)} dg \; h(g) e^{N\L(g)}} < f(g^*) + \epsilon.
\end{align}
Now, we use the results in Eq.~\eqref{eq:weak_denom}, Eq.~\eqref{eq:weak_nom1} and Eq.~\eqref{eq:weak_nom2} to take the limit of Eq.~\eqref{eq:weak_ratios} (we can compose these limits by the algebraic limit theorem as all the individual limits exist and are finite),
%for any $\epsilon$ error, we can find a distance $\delta$, and for any distance $\delta$, we can find a superlevel set $A(\Delta)$.
%Combining these results with an infitesimal choice of $\epsilon$, we get,
\begin{align}
  f(g^*) - \epsilon < \lim_{N\rightarrow \infty} \E\sqb{f(g)} < f(g^*) + \epsilon.
\end{align}
And as this holds for any $\epsilon$, we have,
\begin{align}
  f(g^*) = \lim_{N\rightarrow \infty} \E\sqb{f(g)}.
\end{align}
This result is applicable to the DGP posterior over Gram matrices, as that posterior can be written as,
\begin{align}
  \P[N]{\G_1,\dotsc,\G_L| \X, \Yr} &\propto h(g) e^{N \L(g)},
\end{align}
where $\L(g)$ is the usual DKM objective,
\begin{align}
  \L(g) &= \L(\G_1,\dotsc,\G_L)
\end{align}
and $h(g)$ is the remaining terms in the log-posterior which do not depend on $N$,
\begin{align}
  h(g) &= \exp\b{-\tfrac{P + 1}{2} \sum_\ell \log \abs{\G_\ell}}
\end{align}
(this requires $P \leq N$ so that $\G_\ell$ is full-rank).

\section{General models in the \limit{}}
\label{sec:true_post:bnn}

Overall, our goal is to compute the integral in Eq.~\eqref{eq:dkp:prior} in the limit as $N \rightarrow \infty$.
While the integral is intractable for general models such as BNNs, we can use variational inference to reason about its properties.
In particular, we can bound the integral using the ELBO,
\begin{align}
  \label{eq:elbo:bnn}
  \log \P{\G_\ell| \G_{\ell-1}} &\geq \text{ELBO}_\ell = \E_{\Q{\F_\ell}}\sqb{\log \P{\G_\ell| \F_\ell} + \log \P{\F_\ell| \G_{\ell-1}} - \log \Q{\F_\ell}}.
\end{align}
Note that $\Q{\F_\ell}$ here is different from $\Q[\theta_\ell]{\f_\lambda^\ell}$ in the main text, both because the approximate posterior here, $\Q{\F_\ell}$ is over all features jointly, $\F_\ell$, whereas the approximate posterior in the main text is only over a single feature, $\f_\lambda^\ell$, and because in the main text, we chose a specific family of distribution with parameters $\theta_\ell$, while here we leave the approximate posterior, $\Q{\F_\ell}$ completely unconstrained, so that it has the flexibility to capture the true posterior.
Indeed, if the optimal approximate posterior is equal to the true posterior, $\Qs{\F_\ell}=\P{\F_\ell| \G_{\ell-1}, \G_\ell}$, then the bound is tight, so we get $\log \P{\G_\ell| \G_{\ell-1}} = \text{ELBO}^*_\ell$.
Our overall strategy is thus to use variational inference to characterise the optimal approximate which is equal to the true posterior $\Qs{\F_\ell}=\P{\F_\ell| \G_{\ell-1}, \G_\ell}$ and use the corresponding ELBO to obtain $\log \P{\G_\ell| \G_{\ell-1}}$.

%We begin by characterising the true posteriors.
%While these are useful, as they tell us properties of those true posteriors, the resulting quantities cannot be computed in practice.
%We therefore give a separate variational approximating procedure to give tractable approximations to BNN posteriors and the BNN DKM objective.

\subsection{Characterising exact BNN posteriors}
\label{sec:exact_post}

Remember that if the approximate posterior family, $\Q{\F_\ell}$ is flexible enough to capture the true posterior $\P{\F_\ell| \G_{\ell-1}, \G_\ell}$, then the $\Qs{\F_\ell}$ that optimizes the ELBO is indeed the true posterior, the bound is tight, so the ELBO is equal to $\log \P{\G_\ell| \G_{\ell-1}}$ \citep{jordan1999introduction,blei2017variational}.
Thus, we are careful to ensure that our approximate posterior family captures the true posterior, by ensuring that we only impose constraints on $\Q{\F_\ell}$ that must hold for the true posterior, $\P{\F_\ell| \G_{\ell-1}, \G_\ell}$.
In particular, note that $\P{\G_\ell| \F_\ell}$ in Eq.~\eqref{eq:K:like} constrains the true posterior to give non-zero mass only to $\F_\ell$ that satisfy  $\G_\ell = \tfrac{1}{N_\ell} \phi(\F_\ell) \phi^T(\F_\ell)$.
However, this constraint is difficult to handle.
We therefore consider an alternative, weaker constraint on expectations, which holds for the true posterior (the first equality below) because Eq.~\eqref{eq:K:like} constrains $\G_\ell = \tfrac{1}{N_\ell} \phi(\F_\ell) \phi^T(\F_\ell)$, and impose the same constraint on the approximate posterior,
\begin{align}
  \label{eq:exp:constraint}
  \G_\ell &= \E_{\P{\F_\ell| \G_\ell, \G_{\ell-1}}}\sqb{\tfrac{1}{N_\ell} \phi(\F_\ell) \phi^T(\F_\ell)} = \E_{\Q{\F_\ell}}\sqb{\tfrac{1}{N_\ell} \phi(\F_\ell) \phi^T(\F_\ell)}.
\end{align}
Now, we can solve for the optimal $\Q{\F_\ell}$ with this constraint on the expectation.
In particular, the Lagrangian is obtained by taking the ELBO (Eq.~\ref{eq:elbo:bnn}), dropping the $\log \P{\G_\ell| \F_\ell}$ term representing the equality constraint (that $\G_\ell = \tfrac{1}{N_\ell} \phi(\F_\ell) \phi^T(\F_\ell)$) and including Lagrange multipliers for the expectation constraint, $\mathbf{\Lambda}$, (Eq.~\ref{eq:exp:constraint}) and the constraint that the distribution must normalize to $1$, $\Lambda$,
\begin{multline}
  L = \int d \F_\ell \Q{\F_\ell} \b{\log \P{\F_\ell| \G_{\ell-1}} - \log \Q{\F_\ell}} \\+ \tfrac{1}{2} \Tr\b{\mathbf{\Lambda} \b{\G_\ell - \int d\F_\ell \Q{\F_\ell} \phi(\F_\ell) \phi^T(\F_\ell)}} + \Lambda \b{1 - \int d\F_\ell \Q{\F_\ell}}
\end{multline}
Differentiating wrt $\Q{\F_\ell}$, and solving for the optimal approximate posterior, $\Qs{\F_\ell}$,
\begin{align}
  0 &= \left. \dd[L]{\Q{\F_\ell}} \right|_{\Qs{\F_\ell}}\\
  0 &= \b{\log \P{\F_\ell| \G_{\ell-1}} - \log \Qs{\F_\ell}} - 1 - \tfrac{1}{2} \Tr\b{\mathbf{\Lambda} \phi(\F_\ell) \phi^T(\F_\ell)} - \Lambda
  \intertext{Solving for $\log \Qs{\F_\ell}$,}
  \log \Qs{\F_\ell} &= \log \P{\F_\ell| \G_{\ell-1}} - \tfrac{1}{2} \Tr\b{\mathbf{\Lambda} \phi(\F_\ell)\phi^T(\F_\ell)  } + \const.\\
  \intertext{Using the cyclic property of the trace,}
  \log \Qs{\F_\ell} &= \log \P{\F_\ell| \G_{\ell-1}} - \tfrac{1}{2} \Tr\b{\phi^T(\F_\ell) \mathbf{\Lambda} \phi(\F_\ell) } + \const.\\
  \intertext{Thus, $\log \Q{\F_\ell}$ can be written as a sum over features,}
  \log \Qs{\F_\ell} &= \sum_{\lambda=1}^{N_\ell} \sqb{\log \P{\f_\lambda^\ell| \G_{\ell-1}} - \tfrac{1}{2} \phi^T(\f_\lambda^\ell) \mathbf{\Lambda} \phi(\f_\lambda^\ell)} + \const = \tsum_{\lambda=1}^{N_L} \log \Q{\f_\lambda^\ell}
\end{align}
so, the optimal approximate posterior is IID over features,
\begin{align}
  \Qs{\F_\ell} &= \tprod_{\lambda=1}^{N_\ell} \Qs{\f_\lambda^\ell}.
\end{align}
Remember that this approximate posterior was only constrained in expectation, and that this constraint held for the true posterior (Eq.~\ref{eq:exp:constraint}).
Thus, we might think that this optimal approximate posterior would be equal to the true posterior.
However, remember that the true posterior had a tighter equality constraint, that $\G_\ell = \tfrac{1}{N_\ell} \phi(\F_\ell) \phi^T(\F_\ell)$, while so far we have only imposed a weaker constraint in expectation (Eq.~\ref{eq:exp:constraint}).
We thus need to check that our optimal approximate posterior does indeed satisfy the equality constraint in the limit as $N \rightarrow \infty$.
This can be shown using the law of large numbers, as $\f_\lambda^\ell$ are IID under the optimal approximate posterior, and by using Eq.~\eqref{eq:exp:constraint} for the final equality,
\begin{align}
  \lim_{N \rightarrow \infty} \tfrac{1}{N_\ell} \phi(\F_\ell) \phi^T(\F_\ell) &= \lim_{N \rightarrow \infty} \tfrac{1}{N_\ell} \sum_{\lambda=1}^{N_\ell} \phi(\f_\lambda^\ell) \phi^T(\f_\lambda^\ell) = \E_{\Q{\f_\lambda^\ell}}\sqb{ \phi(\f_\lambda^\ell) \phi^T(\f_\lambda^\ell)} = \G_\ell.
\end{align}
Thus, the optimal approximate posterior does meet the constraint in the limit as $N_\ell \rightarrow \infty$, so in that limit, the true posterior, like the optimal approximate posterior is IID across features,
\begin{align}
  \label{eq:bnn:cond_joint}
  \P{\F_\ell| \G_{\ell-1}, \G_\ell} &= \Qs{\F_\ell} = \tprod_{\lambda=1}^{N_\ell} \Qs{\f_\lambda^\ell} = \tprod_{\ell=1}^{N_\ell} \P{\f_\lambda^\ell| \G_{\ell-1}, \G_\ell}.
\end{align}

\subsection{Exactly multivariate Gaussian DGP posteriors}
\label{sec:mvg_dgp_post}

For DGPs, we have $\phi(\f_\lambda^\ell) = \f_\lambda^\ell$, so the optimal approximate posterior is Gaussian,
\begin{align}
  \log \Qdgp{\f_\lambda^\ell} &=\log \Pdgp{\f_\lambda^\ell| \G_{\ell-1}} - \tfrac{1}{2} (\f_\lambda^\ell)^T \mathbf{\Lambda} \f_\lambda^\ell + \const\\
   &= - \tfrac{1}{2} (\f_\lambda^\ell)^T \b{\mathbf{\Lambda} + \K^{-1}(\G_{\ell-1})}  \f_\lambda^\ell + \const \\
   &= \log \N{\f_\lambda^\ell; \0, \b{\mathbf{\Lambda} + \K^{-1}(\G_{\ell-1})}^{-1}}.
\end{align}
As the approximate posterior and true posterior are IID, the constraint in Eq.~\eqref{eq:exp:constraint} becomes,
\begin{align}
  \G_\ell &= \E_{\Pdgp{\f^\ell_\lambda| \G_\ell, \G_{\ell-1}}}\sqb{\f_\lambda^\ell (\f_\lambda^\ell)^T} = \E_{\Qdgp{\f^\ell_\lambda}}\sqb{\f_\lambda^\ell (\f_\lambda^\ell)^T} = \b{\mathbf{\Lambda} + \K^{-1}(\G_{\ell-1})}^{-1}.
\end{align}
As the Lagrange multipliers are unconstrained, we can always set them such that this constraint holds.
In that case both the optimal approximate posterior and the true posterior become,
\begin{align}
  \label{eq:dgp:mvg_post_app}
  \Pdgp{\f_\lambda^\ell| \G_{\ell-1}, \G_\ell} &= \Qdgp{\f_\lambda^\ell} = \N{\f_\lambda^\ell; \0, \G_\ell},
\end{align}
as required.

%\subsection{General form for the conditional distribution over Gram matrices}
%%$\log \P{\G_\ell| \G_{\ell-1}}$}
%\label{app:P(GG)}
% 
%Now that we have shown that the true posterior, $\P{\F_\ell| \G_{\ell-1}, \G_\ell}$ factorises, we can obtain a simple form for $\log \P{\G_\ell| \G_{\ell-1}}$.
%\begin{align}
%  \P{\G_\ell| \G_{\ell-1}} &= \int d\F_\ell \; \P{\G_\ell| \F_\ell} \P{\F_\ell| \G_{\ell-1}}
%  \intertext{Now, multiply and divide by the posterior,}
%  \P{\G_\ell| \G_{\ell-1}} &= \int d\F_\ell \; \P{\F_\ell| \G_\ell, \G_{\ell-1}} \frac{\P{\G_\ell| \F_\ell} \P{\F_\ell| \G_{\ell-1}}}{\P{\F_\ell| \G_\ell, \G_{\ell-1}}}
%  \intertext{And write this as an expectation,}
%  \P{\G_\ell| \G_{\ell-1}} &= \E_{\P{\F_\ell| \G_{\ell-1}, \G_\ell}}\sqb{\frac{\P{\G_\ell| \F_\ell} \P{\F_\ell| \G_{\ell-1}}}{\P{\F_\ell| \G_\ell, \G_{\ell-1}}}}
%\intertext{As $\P{\G_\ell| \F_\ell}$ for $\F_\ell$ sampled from $\P{\F_\ell| \G_{\ell-1}, \G_\ell}$, is constant}
%  \P{\G_\ell| \G_{\ell-1}} &= \E_{\P{\F_\ell| \G_{\ell-1}, \G_\ell}}\sqb{\P{\G_\ell| \F_\ell}}
%  \E_{\P{\F_\ell| \G_{\ell-1}, \G_\ell}}\sqb{\frac{\P{\F_\ell| \G_{\ell-1}}}{\P{\F_\ell| \G_\ell, \G_{\ell-1}}}}
%\end{align}
%In the first term, we change variables from $\F_\ell$ to $\Gb=\tfrac{1}{N_\ell} \F_\ell \F_\ell^T$,
%\begin{align}
%  \E_{\P{\F_\ell| \G_{\ell-1}, \G_\ell}}\sqb{\P{\G_\ell| \F_\ell}} &= \int d\F_\ell \P{\F_\ell| \G_{\ell-1}, \G_\ell}\P{\G_\ell| \F_\ell}\\
%  &= \int d\F_\ell \P{\F_\ell| \G_{\ell-1}, \G_\ell}\delta(\G_\ell - \tfrac{1}{N_\ell} \F_\ell \F_\ell^T)\\
%\end{align}

\subsection{General form for the conditional distribution over Gram matrices}
%$\log \P{\G_\ell| \G_{\ell-1}}$}
\label{app:P(GG)}
 
Now that we have shown that the true posterior, $\P{\F_\ell| \G_{\ell-1}, \G_\ell}$ factorises, we can obtain a simple form for $\log \P{\G_\ell| \G_{\ell-1}}$.
In particular, $\log \P{\G_\ell| \G_{\ell-1}}$ is equal to the ELBO if we use the true posterior in place of the approximate posterior,
\begin{align}
  \lim_{N \rightarrow \infty} \tfrac{1}{N} \log \P{\G_\ell| \G_{\ell-1}} &= \lim_{N \rightarrow \infty} \tfrac{1}{N} \E_{\P{\F_\ell| \G_{\ell-1}, \G_\ell}}\sqb{\log \P{\G_\ell| \F_\ell} + \log\frac{\P{\F_\ell| \G_{\ell-1}}}{\P{\F_\ell| \G_{\ell-1}, \G_\ell}}}.
  \intertext{Under the posterior, the constraint represented by $\log \P{\G_\ell| \F_\ell}$ is satisfied, so in the limit we can include that term in a constant,}
  \lim_{N \rightarrow \infty} \tfrac{1}{N} \log \P{\G_\ell| \G_{\ell-1}} &= \lim_{N \rightarrow \infty} \tfrac{1}{N} \E_{\P{\F_\ell| \G_{\ell-1}, \G_\ell}}\sqb{\log\frac{\P{\F_\ell| \G_{\ell-1}}}{\P{\F_\ell| \G_{\ell-1}, \G_\ell}}} + \const.
  \intertext{Now, we use the fact that the prior, $\P{\F_\ell| \G_{\ell-1}}$ and posterior, $\P{\F_\ell| \G_{\ell-1}, \G_\ell}$, are IID across features,}
  \lim_{N \rightarrow \infty} \tfrac{1}{N} \log \P{\G_\ell| \G_{\ell-1}} &= \nu_\ell \E_{\P{\f_\lambda^\ell| \G_{\ell-1}, \G_\ell}}\sqb{\log \frac{\P{\f_\lambda^\ell| \G_{\ell-1}}}{\P{\f_\lambda^\ell| \G_{\ell-1}, \G_\ell}}} + \const
  \intertext{and this expectation is a KL-divergence,}
  \label{eq:GG}
  \lim_{N \rightarrow \infty} \tfrac{1}{N} \log \P{\G_\ell| \G_{\ell-1}} &= -\nu_\ell \KL{\P{\f_\lambda^\ell| \G_{\ell-1}, \G_\ell}}{\P{\f_\lambda^\ell| \G_{\ell-1}}} + \const,
\end{align}
which gives Eq.~\eqref{eq:MAP} when we combine with Eq.~\eqref{eq:dgp:joint}.

\subsection{Parametric approximate posteriors}
\label{sec:true_post:bnn:approx}
Eq.~\eqref{eq:bnn:cond_joint} represents a considerable simplification, as we now need to consider only a single feature, $\f_\lambda^\ell$, rather than the joint distribution over all features, $\F_\ell$.
However, in the general case, it is still not possible to compute Eq.~\eqref{eq:bnn:cond_joint} because the true posterior over a single feature is still not tractable.
Following the true posteriors derived in the previous section, we could chose a parametric approximate posterior that factorises across features,
\begin{align}
  \Q[\theta]{\F_1,\dotsc,\F_L} &= \tprod_{\ell=1}^L \tprod_{\lambda=1}^{N_\ell} \Q[\theta_\ell]{\f_\lambda^\ell}.
\end{align}
Remember that we optimize the approximate posterior parameters, $\theta$, directly, and set the Gram matrices as a function of $\theta$ (Eq.~\ref{eq:Kt}).
%\begin{align}
%  \label{eq:K(theta)}
%  \G_\ell(\theta_\ell) &= \E_{\Q[\theta_\ell]{\f_\lambda^\ell}}\sqb{\phi(\f_\lambda^\ell) \phi^T(\f_\lambda^\ell)},
%\end{align}
As before, we can bound, $\log \P{\G_\ell{=}\Gt{\ell}| \G_{\ell-1}}$ using the ELBO, and the bound is tight when the approximate posterior equals the true posterior,
\begin{align}
  \log & \P{\G_\ell=\Gt{\ell}| \G_{\ell-1}} \\
  &= \E_{\P{\F_\ell| \G_{\ell-1}, \G_\ell{=}\Gt{\theta}}}\sqb{\log \P{\G_\ell{=}\Gt{\ell}| \F_\ell} + \log\frac{\P{\F_\lambda^\ell| \G_{\ell-1}}}{\P{\F_\ell|\G_{\ell-1}, \G_\ell{=}\Gt{\ell}}}}\\
  &\geq \E_{\Q[\theta]{\F_\ell}}\sqb{\log \P{\G_\ell{=}\Gt{\ell}| \F_\ell} + \log\frac{\P{\F_\lambda^\ell| \G_{\ell-1}}}{\Q[\theta_\ell]{\F_\ell}}}.
\end{align}
Now, we can cancel the $\log \P{\G_\ell=\Gt{\ell}| \F_\ell}$ terms, as they represent a constraint that holds both under the true posterior, and under the approximate posterior,
\begin{align}
  \E_{\P{\F_\ell| \G_{\ell-1}, \G_\ell{=}\Gt{\ell})}}\sqb{\log\frac{\P{\F_\ell| \G_{\ell-1}}}{\P{\F_\ell|\G_{\ell-1}, \G_\ell{=}\Gt{\ell}}}}
  &\geq \E_{\Q[\theta_\ell]{\F_\ell}}\sqb{\log\frac{\P{\F_\ell| \G_{\ell-1}}}{\Q[\theta_\ell]{\F_\ell}}}.
\end{align}
Using the fact that the prior, posterior and approximate posterior are all IID over features, we can write this inequality in terms of distributions over a single feature, $\f_\lambda^\ell$ and divide by $N_\ell$,
\begin{align}
  \E_{\P{\f_\lambda^\ell| \G_{\ell-1}, \G_\ell{=}\Gt{\ell}}}\sqb{\log\frac{\P{\f_\lambda^\ell| \G_{\ell-1}}}{\P{\f_\lambda^\ell|\G_{\ell-1}, \G_\ell{=}\Gt{\ell}}}}
  &\geq \E_{\Q[\theta_\ell]{\f_\lambda^\ell}}\sqb{\log\frac{\P{\f_\lambda^\ell| \G_{\ell-1}(\theta)}}{\Q[\theta_\ell]{\f_\lambda^\ell}}}.
\end{align}
Noting that both sides of this inequality are negative KL-divergences, we obtain,
\begin{align}
%&\lim_{N \rightarrow \infty} \tfrac{1}{N} \log \P{\G_\ell{=}\Gt{\ell}| \G_{\ell-1}} &\geq - \nu_\ell \KL{\Q[\theta_\ell]{\f_\lambda^\ell}}{\P{\f_\lambda^\ell| \G_{\ell-1}}} + \const.\\
\label{eq:approx:kl}
-\KL{\P{\f_\lambda^\ell| \G_{\ell-1}, \G_\ell{=}\Gt{\ell}}}{\P{\f_\lambda^\ell| \G_{\ell-1}}} \geq -\KL{\Q[\theta_\ell]{\f_\lambda^\ell}}{\P{\f_\lambda^\ell| \G_{\ell-1}}},
\end{align}
which gives Eq.~\eqref{eq:bound} in the main text.

\section{Theoretical similarities in representation learning in finite and infinite networks}
%Here, we consider two arguments.
%First, we show that MAP inference over features gives the exact same result for any $N$ (finite or infinite).
%Second, we show that the expected Langevin updates to the Gram matrix are the same, irrespective of $N$ and are equivalent to preconditioned optimization of the DKM objective.
%
%\subsection{Equivalence of MAP for DGP features}
\label{sec:MAP}
In the main text, we considered probability densities of the Gram matrices, $\G_1,\dotsc,\G_L$.
However, we can also consider probability densities of the features, $\F_1,\dotsc,\F_L$, for a DGP,
\begin{align}
  \label{eq:PFF_}
  \log \P{\F_{\ell}| \F_{\ell-1}} 
  &= -\tfrac{N_{\ell}}{2} \log \abs{\K\b{\Gdgp\b{\F_{\ell-1}}}} - \tfrac{1}{2} \tr\b{\F^T_{\ell} \K^{-1}\b{\Gdgp\b{\F_{\ell-1}}} \F_{\ell}}+ \const. \\
  \intertext{We can rewrite the density such that it is still the density of features, $\F_\ell$, but it is expressed in terms of the DGP Gram matrix,}
  \label{eq:PFF}
  \log \P{\F_{\ell}| \F_{\ell-1}} &= -\tfrac{N_{\ell}}{2} \log \abs{\K(\G_{\ell-1})} - \tfrac{N_\ell}{2} \tr\b{\K^{-1}(\G_{\ell-1}) \G_{\ell}} + \const.
\end{align}
Here, we have used the cyclic property of the trace to combine the $\F_\ell$ and $\F_\ell^T$ to form $\G_\ell$, and we have used the fact that our kernels can be written as a function of the Gram matrix.
Overall, we can therefore write the posterior over features, $\P{\F_1,\dotsc,\F_L| \X, \Yr}$, in terms of only Gram matrices,
\begin{align}
  \label{eq:J}
  \mathcal{J}(\G_1,\dotsc,\G_L) &= \tfrac{1}{N} \log \P{\F_1,\dotsc,\F_L| \X, \Yr} = \log \P{\Y| \G_L} + \tfrac{1}{N} \sum_{\ell=1}^L \log \P{\F_{\ell}| \F_{\ell-1}},
\end{align}
substituting Eq.~\eqref{eq:PFF},
\begin{align}
  \label{eq:FMAP}
  \mathcal{J}(\G_1,\dotsc,\G_L)
  = \log \P{\Y| \G_L} - \tfrac{1}{2} \tsum_{\ell=1}^L \nu_{\ell} \b{\log \abs{\K(\G_{\ell-1})}+ \tr\b{\K^{-1}(\G_{\ell-1}) \G_{\ell}}} + \const.
\end{align}
Thus, $\mathcal{J}(\G_1,\dotsc,\G_L)$ does not depend on $N$, and thus the Gram matrices that maximize $\mathcal{J}(\G_1,\dotsc,\G_L)$ are the same for any choice of $N$.
The only restriction is that we need $N_\ell \geq P$, to ensure that the Gram matrices are full-rank.

\begin{figure*}
    \centering
    \includegraphics[width=\textwidth]{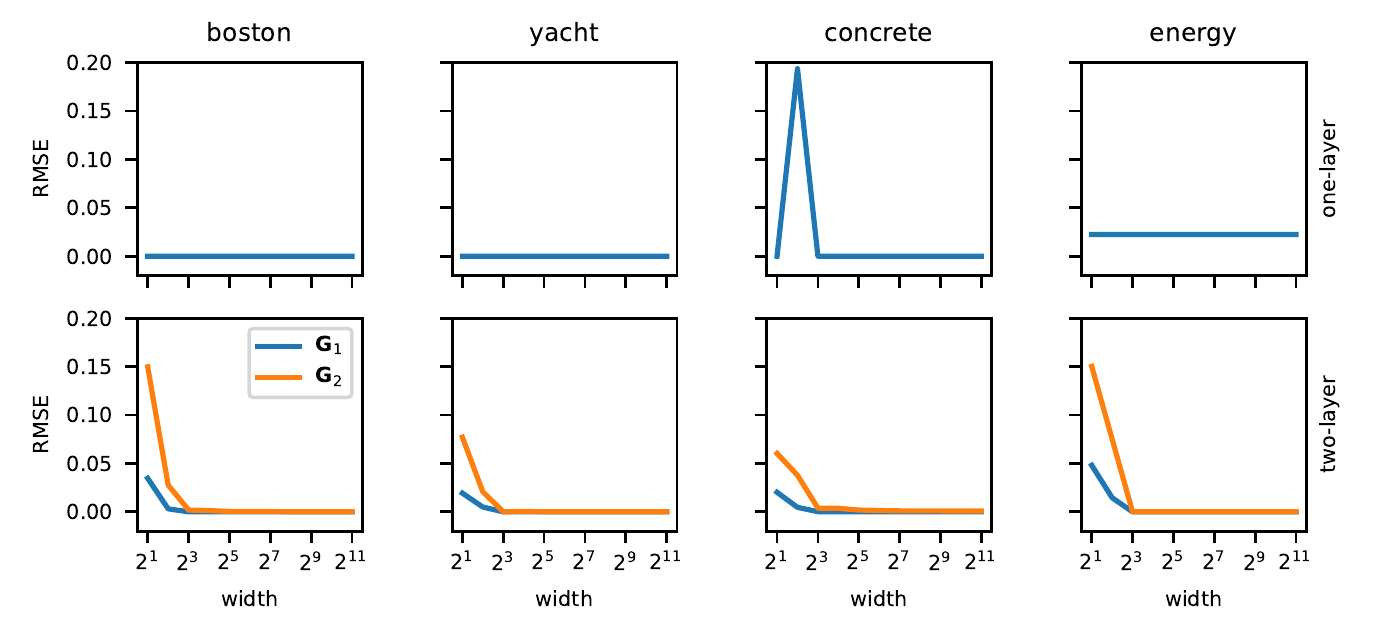}
    \caption{RMSE of trained Gram matrices between one-hidden-layer (first row) and two-hidden-layer (second row) DGPs of various width trained by gradient descent and the corresponding MAP limit. Columns correspond to different datasets (trained on a subset of 50 datapoints). }
    \label{fig:SGD_dist}
\end{figure*}

To confirm these results, we used Adam with a learning rate of $10^{-3}$ to optimize full-rank Gram matrices with Eq.~\eqref{eq:FMAP} and to directly do MAP inference over features using Eq.~\eqref{eq:PFF_}.
As expected, as the number of features increased, the Gram matrix from MAP inference over features converged rapidly to that expected using Eq.~\eqref{eq:FMAP} (Fig.~\ref{fig:SGD_dist}).

\section{Additional experimental details}
\label{sec:methods}

To optimize the analytic DKM objective for DGPs and the variational DKM objective for DGPs (Figs.~\ref{fig:SGLD_marg_1d}--%
%fig:twolayer_yacht,fig:SGLD_deep_hist
\ref{fig:SGLD_deep_gram}), we parameterised the Gram matrices (or covariances for the variational approximate posterior) as the product of a square matrix, $\R_\ell\in\mathbb{R}^{P\times P}$, with itself transposed, $\G_\ell = \tfrac{1}{P} \R_\ell \R_\ell^T$, and we used Adam with a learning rate of $10^{-3}$ to learn $\R_\ell$.
To do Bayesian inference in finite BNNs and DGPs, we used Langevin sampling with 10 parallel chains, and a step size of $10^{-3}$.
Note that in certain senarios, Langevin sampling can be very slow, as the features have a Gaussian prior with covariance $\K(\G_{\ell-1})$ which has some very small and some larger eigenvalues, which makes sampling difficult.
Instead, we reparameterised the model in terms of the standard Gaussian random variables, $\V_\ell \in \mathbb{R}^{P\times N_\ell}$. 
We then wrote $\F_\ell$ in terms of $\V_\ell$, 
\begin{align}
  \F_\ell = \mathbf{L}_{\ell-1} \V_\ell.
\end{align}
Here, $\mathbf{L}_{\ell-1}$ is the Cholesky of $\K(\G_{\ell-1})$, so $\K(\G_{\ell-1}) = \mathbf{L}_{\ell-1} \mathbf{L}_{\ell-1}^T$.
This gives an equivalent distribution $\P{\F_\ell| \F_{\ell-1}}$. 
Importantly, as the prior on $\V_\ell$ is IID standard Gaussian, sampling $\V_\ell$ is much faster.
To ensure that the computational cost of these expensive simulations remained reasonable, we used a subset of 50 datapoints from each dataset.

For the DKM objective for BNNs, we used Monte-Carlo to approximate the Gram matrices,
\begin{align}
  \Gt{\ell} &\approx \tfrac{1}{K}\sum_{k=1}^K \phi(\f_k^\ell)\phi^T(\f_k^\ell).
\end{align}
with $\f_k^\ell$ drawn from the appropriate approximate posterior, and $K=2^{16}$.
We can use the reparameterisation trick \citep{kingma2013auto,rezende2014stochastic} to differentiate through these Monte-Carlo estimates.

\section{Additional comparisons with finite-width DGPs}
\label{sec:app:finite}
\begin{figure*}
    \centering
    \includegraphics[width=\textwidth]{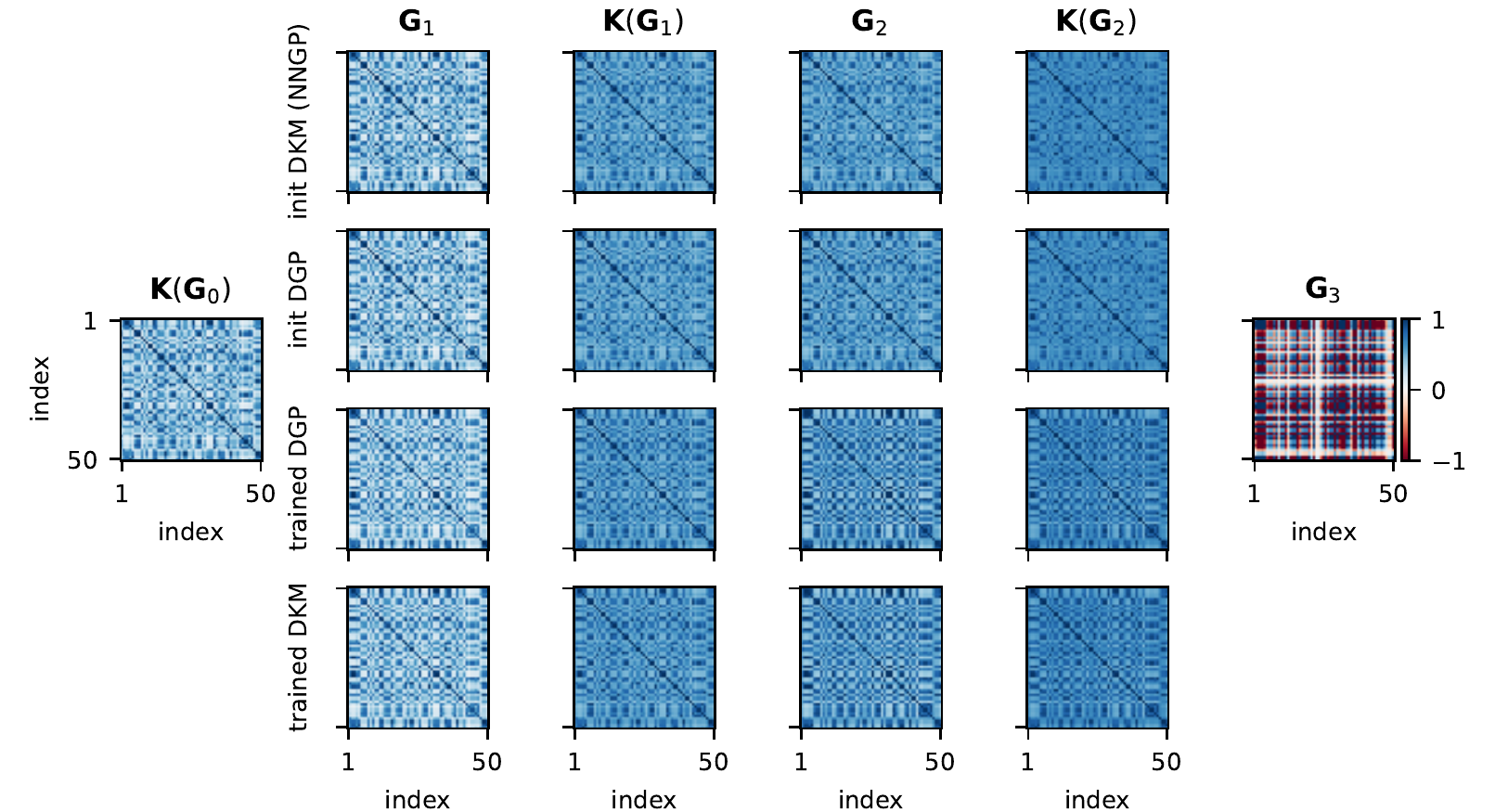}
    \caption{One hidden layer DGP and DKM with squared exponential kernel trained on a subset of energy. First and second row: initializations of DGP and DKM. Third and fourth row: trained DGP (by Langevin sampling) and DKM Gram matrices and kernels.}
    \label{fig:KG_twolayer_energy_sqexp}
\end{figure*}

\begin{figure*}
    \centering
    \includegraphics[width=\textwidth]{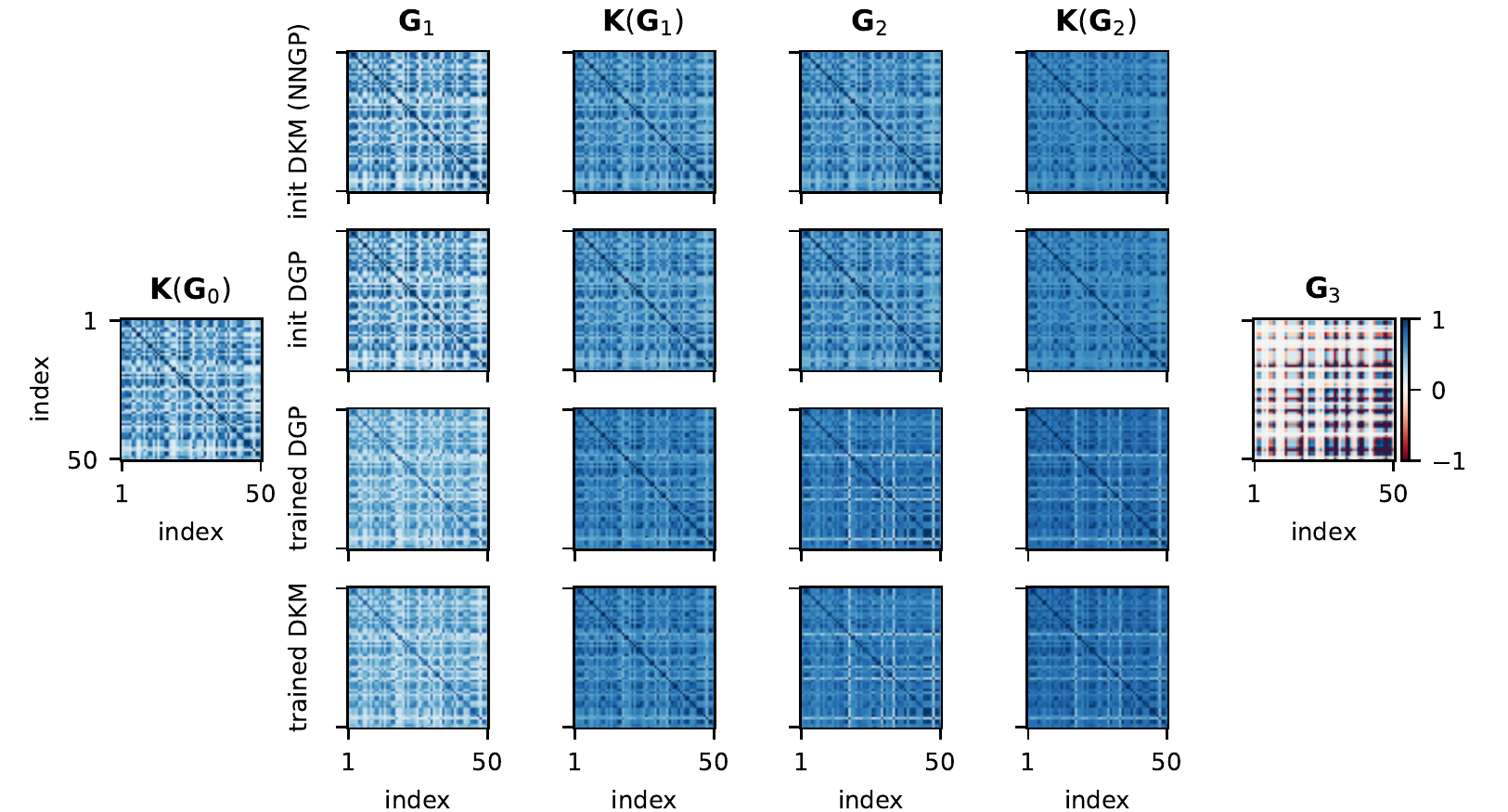}
    \caption{One hidden layer DGP and DKM with squared exponential kernel trained on a subset of boston. First and second row: initializations of DGP and DKM. Third and fourth row: trained DGP (by Langevin sampling) and DKM Gram matrices and kernels.}
    \label{fig:KG_twolayer_boston_sqexp}
\end{figure*}

\begin{figure*}
    \centering
    \includegraphics[width=\textwidth]{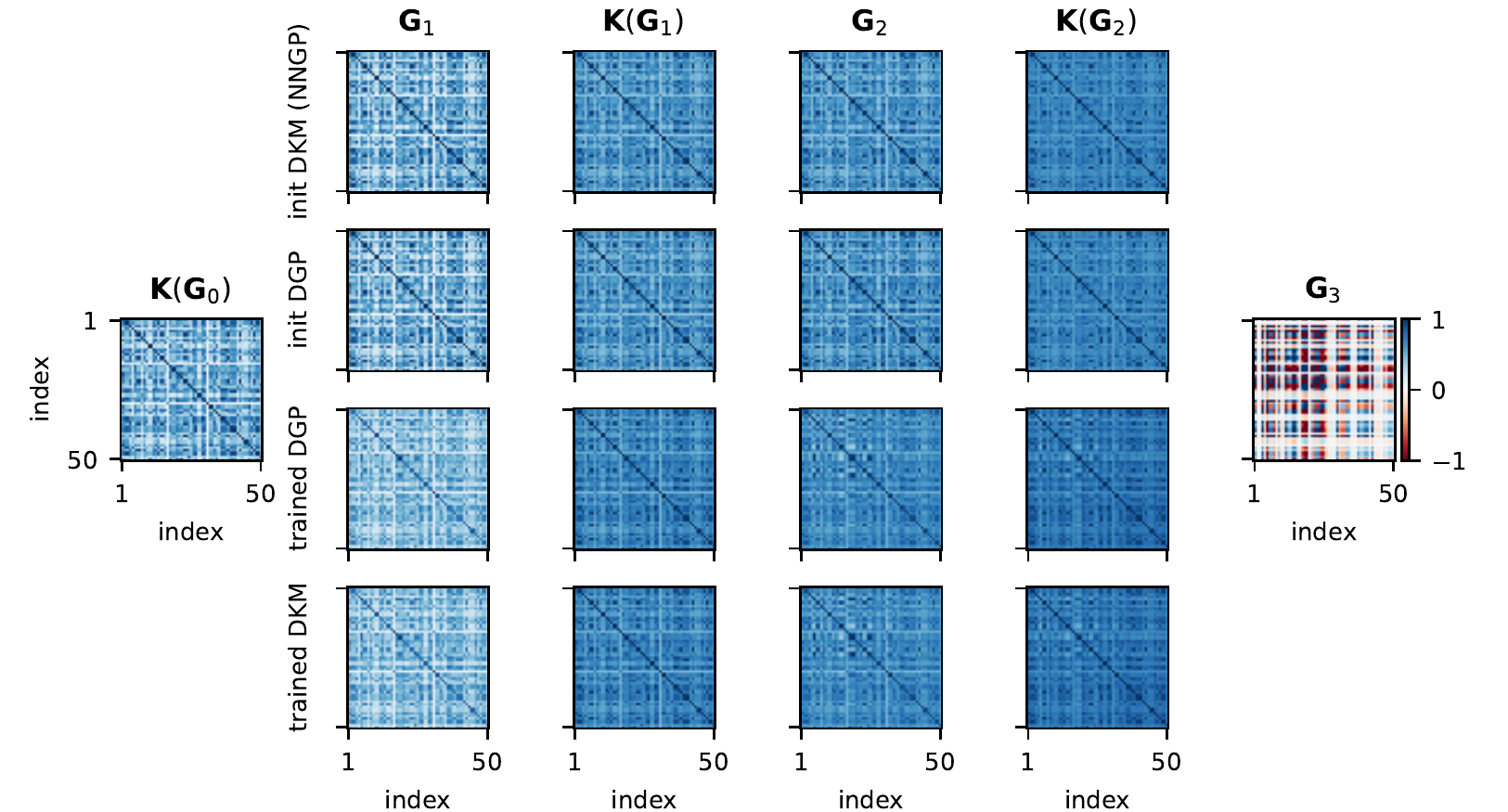}
    \caption{One hidden layer DGP and DKM with squared exponential kernel trained on a subset of concrete. First and second row: initializations of DGP and DKM. Third and fourth row: trained DGP (by Langevin sampling) and DKM Gram matrices and kernels.}
    \label{fig:KG_twolayer_concrete_sqexp}
\end{figure*}

Here, we give additional results supporting those in Sec.~\ref{sec:finite}, Fig.~\ref{fig:SGLD_marg_1d}--Fig.~\ref{fig:SGLD_deep_gram}.
In particular, we give the DGP representations learned by two-layer networks on all UCI datasets (boston, concrete, energy, yacht), except those already given in the main text Fig.~\ref{fig:KG_twolayer_energy_sqexp}--\ref{fig:KG_twolayer_concrete_sqexp}.

\FloatBarrier
\clearpage
\section{The flow posterior in a 2-layer BNN}
\begin{figure}[h]
    \centering
    \includegraphics[width=.5\textwidth]{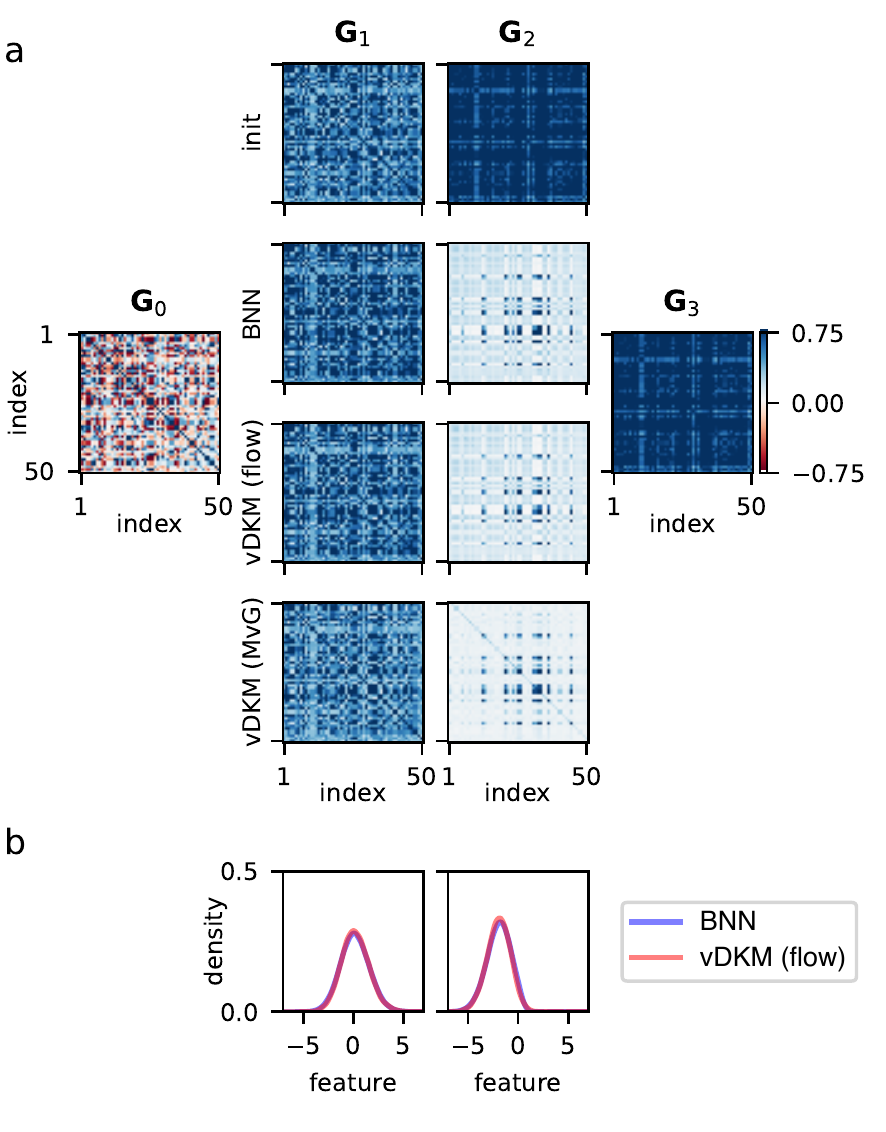}
  \caption{Two-layer ReLU BNN and variational DKM with flow. \textbf{a} Initialized (first row) and learned Gram matrices of a width 1024 BNN (second row), vDKM with flow (third row) and vDKM with multivariate Gaussian (fourth row) using $2^{14}$ Monte-Carlo samples. The Gram matrices between BNN and vDKM (flow) match closely after training.
  (MvG). \textbf{b} Marginal PDF over features at each layer for one input datapoint using kernel density estimation. The marginal PDFs of BNN are non-Gaussian (blue curves), vDKM with flow is able to capture the non-Gaussianity and match closely with BNNs marginals (red curves).
  }
  \label{fig:flow}
\end{figure}

Here, we give the 2-layer version (Fig.~\ref{fig:flow}) of Fig.~\ref{fig:flow_4} in the main text, which again shows a close match between the variational DKM with a flow posterior, and the BNN true posterior.

\FloatBarrier
\clearpage
\section{Multivariate Gaussian approximate posteriors in deeper networks}
There is a body of theoretical work (e.g.\ \citep{seroussi2021separation}), on BNNs that approximates BNN posteriors over features as Gaussian.
While we have shown that this is a bad idea in general (Fig.~\ref{fig:flow_4} and \ref{fig:flow}), we can nonetheless ask whether there are circumstances where the idea might work well.
In particular, we hypothesised that depth is an important factor.
In particular, in shallow networks, in order to get $\G_L$ close to the required representation, we may need the posterior over $\F_\ell$ to be quite different from the prior.
In contrast, in deeper networks, we might expect the posterior over $\F_\ell$ to be closer to its (Gaussian) prior, and therefore we might expect Gaussian approximate posteriors to work better.

However, we cannot just make the network deeper, because as we do so, we apply the nonlinearity more times and dramatically alter the network's inductive biases.
To resolve this issue, we derive a leaky relu nonlinearity that allows (approximately) independent control over the inductive biases (or effective depth) and the actual depth (Appendix~\ref{sec:leaky_relu}).
Using these nonlinearities, we indeed find that very deep networks are reasonably well approximated by multivariate Gaussian approximate posteriors (Appendix~\ref{sec:Qmvg}).

\subsection{Leaky relu nonlinearities}
\label{sec:leaky_relu}
Our goal is to find a pointwise nonlinearity, $\phi$, such that (under the prior),
\begin{align}
  \label{eq:full_desired}
  \E_{\Pdgp{\f_\lambda^\ell| \G_{\ell-1}}}\sqb{\phi(\f_\lambda^\ell) \phi^T(\f_\lambda^\ell)} &= p \E_{\P{\f_\lambda^\ell| \G_{\ell-1}}}\sqb{\text{relu}(\f_\lambda^\ell) \text{relu}^T(\f_\lambda^\ell)} + (1-p) \G_{\ell-1}.
\end{align}
We will set $p=\alpha/L$, where $\alpha$ is the ``effective'' depth of the network and $L$ is the real depth. 
These networks are designed such that their inductive biases in the infinite width limit are similar to a standard relu network with depth $\alpha$.
Indeed, we would take this approach if we wanted a well-defined infinite-depth DKM limit.

Without loss of generality, we consider a 2D case, where $x$ and $y$ are zero-mean bivariate Gaussian,
\begin{align}
  \pi(x, y) &= \N{\begin{pmatrix} x\\y\end{pmatrix}; \0, \begin{pmatrix} 
    \Sigma_\text{xx} & \Sigma_\text{xy} \\ 
    \Sigma_\text{xy} & \Sigma_\text{yy}
  \end{pmatrix}}
\end{align}
where $\pi(x, y)$ is the probability density for the joint distribution.
Note that we use a scaled relu,
\begin{align}
  \label{eq:scaled_relu}
  \text{relu}(x) &= \begin{cases}
    \sqrt{2} \; x & \text{for } 0 < x\\
    0 & \text{otherwise}
  \end{cases}
\end{align}
such that $\E\sqb{\text{relu}^2(x)} = \Sigma_\text{xx}$.
Mirroring Eq.~\ref{eq:full_desired}, we want the nonlinearity, $\phi$, to satisfy,
\begin{subequations}
\label{eq:desired}
\begin{align}
  \label{eq:desired_x}
  \E\sqb{\phi(x^2)} &= p \E\sqb{\text{relu}^2(x)} + (1-p) \Sigma_\text{xx} = \Sigma_\text{xx} \\
  \label{eq:desired_y}
  \E\sqb{\phi(y^2)} &= p \E\sqb{\text{relu}^2(y)} + (1-p) \Sigma_\text{yy} = \Sigma_\text{yy} \\
  \label{eq:desired_cross}
  \E\sqb{\phi(x) \phi(y)} &= p \E\sqb{\text{relu}(x) \text{relu}(y)} + (1-p) \Sigma_\text{xy}
\end{align}
\end{subequations}

We hypothesise that this nonlinearity has the form,
\begin{align}
  \label{eq:nonlin_gen}
  \phi(x) &= a \; \text{relu}(x) + b x.
  \intertext{We will write the relu as a sum of $x$ and $\abs{x}$,}
  \text{relu}(x) &= \tfrac{1}{\sqrt{2}} (x + \abs{x}),
\end{align}
because $\E\sqb{f(x, y)} = 0$ for $f(x, y) = x \abs{y}$ or $f(x, y) = \abs{x} y$.
It turns out that we get zero expectation for all functions where $f(-x, -y) = - f(x, y)$, which holds for the two choices above.
To show such functions have a zero expectation, we write out the integral explicitly,
\begin{align}
  \E\sqb{f(x, y)} &= \int_{-\infty}^\infty dx \int_{-\infty}^\infty dy\; \pi(x, y) f(x, y).\\
  \intertext{We split the domain of integration for $y$ at zero,}
  \E\sqb{f(x, y)} &= \int_{-\infty}^\infty dx \int_{-\infty}^0 dy\; \pi(x, y) f(x, y) + \int_{-\infty}^\infty dx \int_{0}^\infty dy\; \pi(x, y) f(x, y).
  \intertext{We substitute $y' = -y$ and $x'=-x$ in the first integral,}
  \E\sqb{f(x, y)} &= \int_{-\infty}^\infty dx' \int^{\infty}_0 dy'\; \pi(-x', -y') f(-x', -y') + \int_{-\infty}^\infty dx \int_{0}^\infty dy\; \pi(x, y) f(x, y).
  \intertext{As the variables we integrate over are arbitrary we can relabel $y'$ as $y$ and $x'$ as $x$, and we can then merge the integrals as their limits are the same,}
  \E\sqb{f(x, y)} &= \int_{-\infty}^\infty dx \int^{\infty}_0 dy\; \sqb{\pi(-x, -y) f(-x, -y) + \pi(x, y) f(x, y)}.
  \intertext{Under a zero-mean Gaussian, $\pi(-x, -y) = \pi(x, y)$,}
  \E\sqb{f(x, y)} &= \int_{-\infty}^\infty dx \int^{\infty}_0 dy\; \pi(x, y)\b{f(-x, -y) + f(x, y)}.
\end{align}
Thus, if $f(-x, -y) = -f(x, y)$, then the expectation of that function under a bivariate zero-mean Gaussian distribution is zero.

Remember that our overall goal was to design a nonlinearity, $\phi$, (Eq.~\ref{eq:nonlin_gen}) which satisfied Eq.~\eqref{eq:desired}.
We therefore compute the expectation,
\begin{align}
  \E\sqb{\phi(x) \phi(y)} &= \E\sqb{\b{a \; \text{relu}(x) + b x} \b{a \; \text{relu}(y) + b y}}\\
  &= \E\sqb{\b{\tfrac{a}{\sqrt{2}} \b{x + \abs{x}} + b x} \b{\tfrac{a}{\sqrt{2}} \b{y + \abs{y}} + b y}}\\
  \intertext{Using the fact that $\E\sqb{\; x \abs{y}\;} = \E\sqb{\;\abs{x} y\;} = 0$ under a multivariate Gaussian,}
  &= \E\sqb{a^2 \tfrac{1}{\sqrt{2}} \b{x + \abs{x}}\tfrac{1}{\sqrt{2}} \b{y + \abs{y}} + \b{\sqrt{2} a b + b^2} xy}\\
  &= a^2 \E\sqb{\text{relu}(x) \text{relu}(y)} + \b{\sqrt{2} a b + b^2} \E\sqb{xy}.
\end{align}
Thus, we can find the value of $a$ by comparing with Eq.~\eqref{eq:desired_cross},
\begin{align}
  \label{eq:p}
  p &= a^2 & a &= \sqrt{p}.\\
  \intertext{For $b$, things are a bit more involved,}
  \label{eq:1mp}
  1-p &= \sqrt{2} a b + b^2 = \sqrt{2 p} \; b + b^2
  \intertext{where we substitute for the value of $a$.  This can be rearranged to form a quadratic equation in $b$,}
  0 &= b^2 + \sqrt{2 p} \; b + (p-1),
\end{align}
which can be solved,
\begin{align}
  b &= \tfrac{1}{2} \b{- \sqrt{2 p} \pm \sqrt{2 p - 4 (p-1)}}\\
  b &= \tfrac{1}{2} \b{- \sqrt{2 p} \pm \sqrt{4 - 2p}}\\
  b &= -\sqrt{\tfrac{p}{2}} \pm \sqrt{1 - \tfrac{p}{2}}\\
  \intertext{Only the positive root is of interest,}
  b &= \sqrt{1 - \tfrac{p}{2}} -\sqrt{\tfrac{p}{2}} 
\end{align}
Thus, the nonlinearity is,
\begin{align}
  \phi(x) &= \sqrt{p} \; \text{relu}(x) + \b{\sqrt{1 - \tfrac{p}{2}} -\sqrt{\tfrac{p}{2}}} x
\end{align}
where we set $p=\alpha/L$, and remember we used the scaled relu in Eq.~\eqref{eq:scaled_relu}.
Finally, we established these choices by considering only the cross term, $\E\sqb{\phi(x) \phi(y)}$.
We also need to check that the $\E\sqb{\phi^2(x)}$ and $\E\sqb{\phi^2(y)}$ terms are as required (Eq.~\ref{eq:desired_x} and Eq.~\ref{eq:desired_y}).
In particular,
\begin{align}
  \E\sqb{\phi^2(x)} &= \E\sqb{\b{a \; \text{relu}(x) + b x}^2} = \E\sqb{\b{\tfrac{a}{\sqrt{2}} \b{x + \abs{x}} + b x}^2}\\
  \intertext{using $\E\sqb{x \abs{x}} = 0$ as $x \abs{x}$ is an odd function of $x$, and the zero-mean Gaussian is an even distribution,}
  \E\sqb{\phi^2(x)} &= a^2 \E\sqb{\text{relu}^2(x)} + \b{\sqrt{2} a b + b^2} \Sigma_\text{xx}\\
  \intertext{using Eq.~\eqref{eq:p} to identify $a^2$ and Eq.~\eqref{eq:1mp} to identify $\sqrt{2} a b + b^2$,}
  \E\sqb{\phi^2(x)} &= p \E\sqb{\text{relu}^2(x)} + (1-p) \Sigma_\text{xx},
\end{align}
as required.

% \begin{figure}
%   \includegraphics[width=\textwidth]{figures/deep_flow_relu_feat_1_reparam_reparam_0_65536_4.pdf}
%   \caption{
%     Four-layer ReLU BNN and variational DKM with flow. \textbf{a} Initialized (first row) and learned Gram matrices of a width 1024 BNN (second row), vDKM with flow (third row) and vDKM with multivariate Gaussian (fourth row) using $2^{15}$ Monte-Carlo samples. The Gram matrices match closely after training. \textbf{b} Marginal PDF over features at each layer for one input datapoint using kernel density estimation. The marginal PDFs of BNN are non-Gaussian (blue curves), vDKM with flow is able to capture the non-Gaussianity and match closely with BNNs marginals (red curves).
%     \label{fig:flow_4}
%   }
% \end{figure}

\subsection{Multivariate Gaussian in deeper networks}
\label{sec:Qmvg}
%Using the nonlinearity derived in Appendix~\ref{sec:leaky_relu}, we can alter the network depth without dramatically changing the inductive biases of the network.
%
%Perhaps the most obvious choice of approximate posterior is a Gaussian with non-zero mean,
%However, it turns out that this fails to closely match the marginals (Fig.~\ref{fig:SGLD_deep_hist} top; compare the blue histograms with the blue line, which is the closest fitting Gaussian), and this disrupts the learned representations (Fig.~\ref{fig:SGLD_deep_gram} top).

\begin{figure*}
  \centering
  \includegraphics[width=\textwidth]{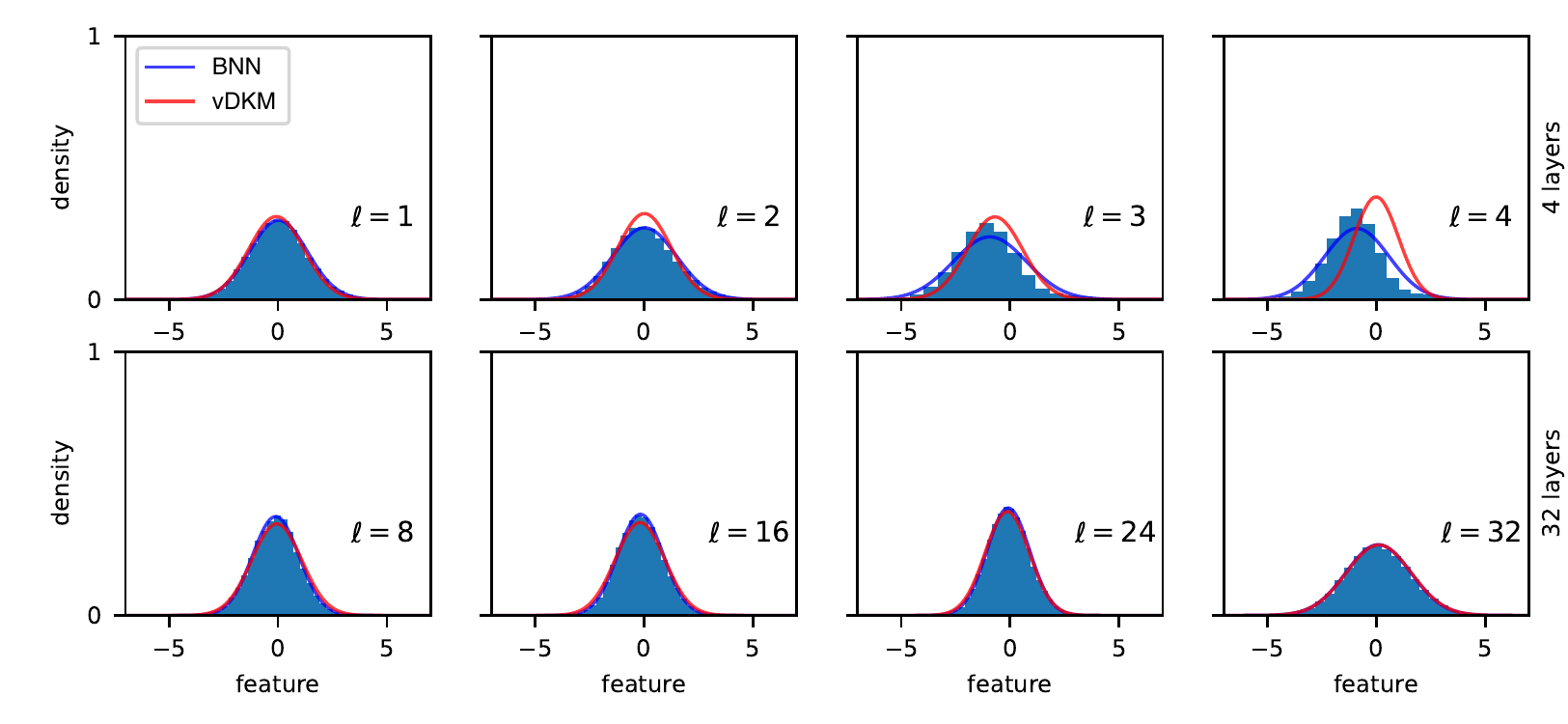}
  \caption{
    Comparison of posterior feature marginal distributions between a BNN of width 1024 (trained by Langevin sampling over features) and a variational DKM with $2^{16}$ Monte-Carlo samples, in a 4-layer (row 1) and a 32-layer (row 2) network. 
    We give the BNN posterior features from Langevin sampling (blue histogarm) and the best fitting Gaussian (blue line), and compare against the variational DKM approximate posterior Gaussian distribution (red line). 
  }
  \label{fig:SGLD_deep_hist}
\end{figure*}

\begin{figure*}
    \centering
    \includegraphics[width=\textwidth]{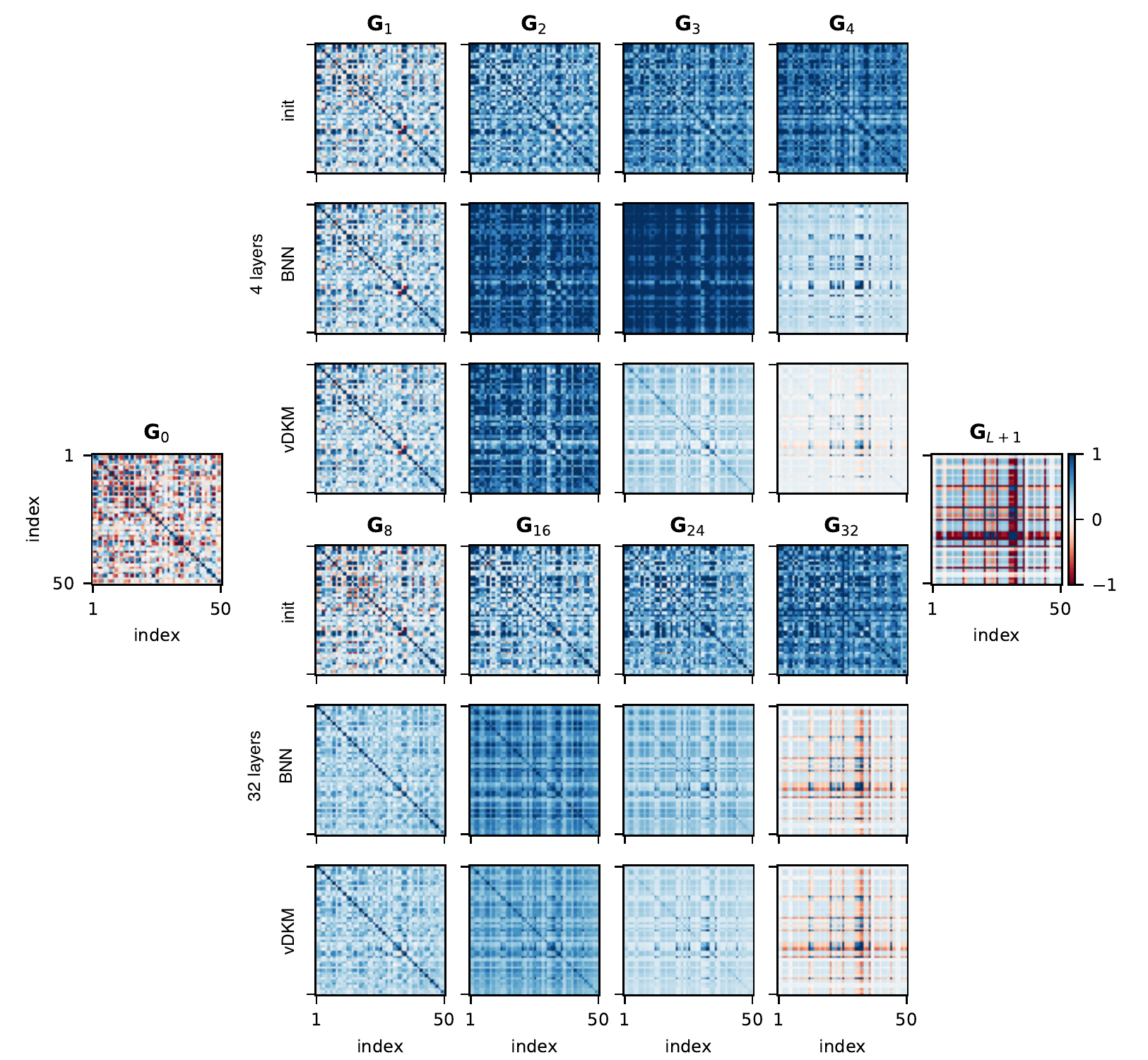}
    \caption{Comparison of Gram matrices between BNN of width 1024 (trained by Langevin sampling over features) and variational DKM, in 4-layer (row 1-3) and 32-layer networks (row 4-6). 
    Initializations are shown in row 1 and 4, trained BNN Gram matrices are shown in row 2 and 5, and trained variational DKM Gram matrices are shown in row 3 and 6. 
    As in Figure~\ref{fig:SGLD_deep_hist}, the variational DKM is a poor match to Langevin sampling in a BNN for a 4-layer network, but is very similar in a 32 layer network.}
    \label{fig:SGLD_deep_gram}
\end{figure*}

In the main text, we show that a more complex approximate posterior can match the distributions in these networks.
Here, we consider an alternative approach.
In particular, we hypothesise that these distributions are strongly non-Gaussian because the networks are shallow, meaning that the posterior needs to be far from the prior in order to get a top-layer kernel close to $\G_{L+1}$.
We could therefore make the posteriors closer to Gaussian by using leaky-relu nonlinearities (Appendix~\ref{sec:leaky_relu}) with fixed effective depth ($\alpha=2$), but increasing real depth, $L$.
In particular, we use multivariate Gaussian approximate posteriors with learned means,
\begin{align}
  \Q[\theta_\ell]{\f_\lambda^\ell} &= \N{\f_\lambda^\ell; \m_\ell, \S_\ell},
\end{align}
with $\theta_\ell = (\m_\ell, \S_\ell)$.
As expected, for a depth 32 network, we get similar marginals (Fig.~\ref{fig:SGLD_deep_hist} top) and learned representations (Fig.~\ref{fig:SGLD_deep_gram} top).

\FloatBarrier
\clearpage
\section{Unimodality in linear deep kernel machines}
\label{app:unimodality}

\subsection{Theory: unimodality with a linear kernel and same widths}
\label{sec:unimodality_linear}
Here, we show that the deep kernel machine objective is unimodal for a linear kernel.
A linear kernel simply returns the input Gram matrix,
\begin{align}
  \K\b{\G} = \G.
\end{align}
It is called a linear kernel, because it arises in the neural network setting (Eq.~\ref{eq:nn}) by choosing the nonlinearity, $\phi$ to be the identity, in which case, $\F_{\ell} = \F_{\ell-1} \W_{\ell-1}$.
For a linear kernel the objective becomes,
\begin{align}
  \label{eq:mean_obj}
  \mathcal{L}(\G_1,...,\G_L) &=
  %&= \tsum_{\ell=1}^{L+1} \nu_\ell \KL{\N{\0,\G_{\ell-1}}}{\N{\0,\G_{\ell}}}.\\
   \tsum_{\ell=1}^{L+1} \tfrac{\nu_\ell}{2} \b{\log\abs{\G_{\ell-1}^{-1} \G_\ell} - \Tr\b{\G_{\ell-1}^{-1}\G_\ell}}\\
  \intertext{where we have assumed there is no output noise, $\sigma^2 = 0$. Taking all $\nu_\ell$ to be equal, $\nu = \nu_\ell$ (see Appendix~\ref{app:unimodal} for the general case),}
  \mathcal{L}(\G_1,...,\G_L)&= \log\abs{\G_0^{-1} \G_{L+1}} - \tfrac{\nu}{2} \tsum_{\ell=1}^{L+1} \Tr\b{\G_{\ell-1}^{-1}\G_\ell}.
\end{align}
Note that $\G_0$ and $\G_{L+1}$ are fixed by the inputs and outputs respectively.  Thus, to find the mode, we set the gradient wrt $\G_1,\dotsc,\G_L$ to zero,
\begin{align}
  \0 &= \dd[\mathcal{L}]{\G_\ell} = \tfrac{\nu}{2} \b{\G_{\ell-1}^{-1} - \G_\ell^{-1} \G_{\ell+1} \G_\ell^{-1}}
\end{align}
Thus, at the mode, the recursive relationship must hold,
\begin{align}
  \T &= \G_{\ell-1}^{-1} \G_\ell = \G_{\ell}^{-1} \G_{\ell+1}.
\end{align}
Thus, optimal Gram matrices are given by,
\begin{align}
  \G_\ell &= \G_0 \T^\ell,
\end{align}
and we can solve for $\T$ by noting,
\begin{align}
  \label{eq:TLp1}
  \G_0^{-1} \G_{L+1} &= \T^{L+1}.
\end{align}
Importantly, $\T$ is the product of two positive definite matrices, $\T = \G_{\ell-1}^{-1} \G_\ell$, so $\T$ must have positive, real eigenvalues (but $\T$ does not have to be symmetric \citep{horn2012matrix}). 
There is only one solution to Eq.~\eqref{eq:TLp1} with positive real eigenvalues \citep{horn1994topics}.
Intuitively, this can be seen using the eigendecomposition, $\G_0^{-1} \G_{L+1} = \V^{-1} \D \V$, where $\D$ is diagonal,
\begin{align}
  \T &= \b{\V^{-1} \D \V}^{1/(L+1)} = \V^{-1} \D^{1/(L+1)} \V.
\end{align}
Thus, finding $\T$ reduces to finding the $(L+1)$th root of each positive real number on the diagonal of $\D$.
While there are $(L+1)$ complex roots, there is only one positive real root, and so $\T$ and hence $\G_1,\dotsc,\G_L$ are uniquely specified.
This contrasts with a deep linear neural network, which has infinitely many optimal settings for the weights.

Note that for the objective to be well-defined, we need $\K(\G)$ to be full-rank.
With standard kernels (such as the squared exponential) this is always the case, even if the input Gram matrix is singular.
However, a linear kernel will have a singular output if given a singular input, and with enough data points, $\G_0 = \tfrac{1}{\Nx} \X \X^T$ is always singular.
To fix this, we could e.g.\ define $\G_0 = \K(\tfrac{1}{\Nx} \X \X^T)$ to be given by applying a positive definite kernel (such as a squared exponential) to $\tfrac{1}{\Nx} \X \X^T$. This results in positive definite $\G_0$, as long as the input points are distinct.

\subsection{Theory: unimodality with a linear kernel and arbitrary widths}
\label{app:unimodal}

In the main text we showed that the deep kernel machine is unimodal when all $\nu_\ell$ are equal. 
Here, we show that unimodality in linear DKMs also holds for all choices of $\nu_\ell$.
Recall the linear DKM objective in Eq.~(\ref{eq:mean_obj}),
\begin{align}
    \mathcal{L}(\G_1,...,\G_L) &= \tsum_{\ell=1}^{L+1} \tfrac{\nu_\ell}{2} \b{\log\abs{\G_{\ell-1}^{-1} \G_\ell} - \Tr\b{\G_{\ell-1}^{-1}\G_\ell}} \\
    &= \tsum_{\ell=1}^{L+1} \tfrac{\nu_\ell}{2} \b{ \log|\G_\ell| - \log|\G_{\ell-1}| - \Tr(\G_{\ell-1}^{-1} \G_\ell) }.
\end{align}
To find the mode, again we set the gradient wrt $\G_\ell$ to zero,
\begin{align}
    \0 &= \pdv{\mathcal{L}}{\G_\ell} = -\tfrac{\nu_{\ell+1}-\nu_\ell}{2} \G_\ell^{-1} - \tfrac{\nu_\ell}{2} \G_{\ell-1}^{-1} + \tfrac{\nu_{\ell+1}}{2} \G_\ell^{-1} \G_{\ell+1} \G_\ell^{-1},
\end{align}
for $\ell=1,...,L$. 
Right multiplying by $2\G_\ell$ and rearranging,
\begin{align}
    \nu_{\ell+1} \G_\ell^{-1} \G_{\ell+1} = \nu_\ell \G_{\ell-1}^{-1} \G_\ell + \b{\nu_{\ell+1}-\nu_\ell}\I , \qquad \text{for } \ell=1,...,L.
\end{align}
Evaluating this expression for $\ell=1$ and $\ell=2$ gives,
\begin{align}
    \nu_{2} \G_1^{-1} \G_2 &= \nu_1 \G_0^{-1} \G_1 + \b{\nu_2-\nu_1} \I,\\
    \nu_{3} \G_2^{-1} \G_3 &= \nu_2 \G_1^{-1} \G_2 + \b{\nu_3-\nu_2} \I 
    = \nu_1 \G_0^{-1} \G_1 + \b{\nu_3-\nu_1}\I.
\end{align}
Recursing, we get,
\begin{align} 
  \label{eq:nuGG}
  \nu_{\ell}\G_{\ell-1}^{-1} \G_{\ell} = \nu_1 \G_0^{-1}\G_1 + \b{\nu_{\ell} - \nu_1}\I.
\end{align}
Critically, this form highlights constraints on $\G_1$.
In particular, the right hand side, $\G_{\ell-1}^{-1} \G_{\ell}$, is the product of two positive definite matrices, so has positive eigenvalues (but may be non-symmetric \citep{horn2012matrix}).
Thus, all eigenvalues of $\nu_1 \G_0^{-1} \G_1$ must be larger than $\nu_1 - \nu_\ell$, and this holds true at all layers.
This will become important later, as it rules out inadmissible solutions.

Given $\G_0$ and $\G_1$, we can compute any $\G_{\ell}$ using,
\begin{align}
  \G_0^{-1} \G_{\ell} &= \prod_{{\ell'}=1}^\ell \b{\G_{{\ell'}-1}^{-1} \G_{{\ell'}}} = \tfrac{1}{\prod_{{\ell'}=1}^\ell \nu_{{\ell'}}} \prod_{{\ell'}=1}^\ell \b{\nu_{\ell'} \G_{{\ell'}-1}^{-1} \G_{{\ell'}}}\\
  \b{\prod_{{\ell'}=1}^\ell \nu_{{\ell'}}} \G_0^{-1} \G_{\ell} &= \prod_{{\ell'}=1}^\ell \b{\nu_1 \G_0^{-1}\G_1 + \b{\nu_{\ell'} - \nu_1}\I}
\end{align}
where the matrix products are ordered as $\prod_{\ell=1}^L \A_\ell = \A_1\dotsm \A_L$.
Now, we seek to solve for $\G_1$ using our knowledge of $\G_{L+1}$.  Computing $\G_0^{-1} \G_{L+1}$,
\begin{align}
  \b{\prod_{\ell=1}^{L+1} \nu_{\ell}} \G_0^{-1} \G_{L+1} &= \prod_{\ell=1}^{L+1} \b{\nu_1 \G_0^{-1}\G_1 + \b{\nu_\ell - \nu_1}\I}.
\end{align}
%Products of positive definite matrices have positive real eigenvalues  
%, so $\G_0^{-1}\G_{\ell+1}$ and $\G_0^{-1}\G_1$ have positive real eigenvalues. 
We write the eigendecomposition of $\nu_1 \G_0^{-1} \G_1$ as,
\begin{align}
  \nu_1\G_0^{-1}\G_1=\V\D\V^{-1}.
\end{align}
Thus,
\begin{align}
  \b{\prod_{\ell=1}^{L+1} \nu_\ell} \G_0^{-1} \G_{L+1} &= \prod_{\ell=1}^{L+1} \b{\V \D \V^{-1}  + \b{\nu_\ell - \nu_1}\I} = \V \La \V^{-1} \\
  \intertext{where $\La$ is a diagonal matrix,}
  \La &= \prod_{\ell=1}^{L+1} \b{\D + \b{\nu_\ell - \nu_1}\I}.
\end{align}
Thus, we can identify $\V$ and $\La$ by performing an eigendecomposition of the known matrix, $\b{ \prod_{\ell=1}^{L+1} \nu_\ell } \G_0^{-1} \G_{L+1}$.
Then, we can solve for $\D$ (and hence $\G_1$) in terms of $\La$ and $\V$.
The diagonal elements of $\D$ satisfy,
\begin{align}
  \label{eq:general_poly}
  0 &= - \Lambda_{ii} + \prod_{k=1}^{L+1} \b{D_{ii} + \b{\nu_\ell - \nu_1}}.  %= - \Lambda_{ii} + \prod_{k=1}^{L+1} \b{D_{ii} - \Omega_k}.
  %\intertext{where,}
  %\Omega_k &= \nu_1 - \nu_k.
\end{align}
This is a polynomial, and remembering the constraints from Eq.~\eqref{eq:nuGG}, we are interested in solutions which satisfy,
\begin{align}
  \nu_1 - \nu_\text{min}  \leq D_{ii}.
  \intertext{where,}
  \label{eq:def:nu_min}
  \nu_\text{min} = \min\b{\nu_1,\dotsc,\nu_{L+1}}.
\end{align}
To reason about the number of such solutions, we use Descartes' rule of signs, which states that the number of positive real roots is equal to or a multiple of two less than the number of sign changes in the coefficients of the polynomial.
Thus, if there is one sign change, there must be one positive real root.
For instance, in the following polynomial,
\begin{align}
  0 &=  x^3 + x^2 - 1
\end{align}
the signs go as $(+), (+), (-)$, so there is only one sign change, and there is one real root.
To use Descartes' rule of signs, we work in terms of $D'_{ii}$, which is constrained to be positive,
\begin{align}
  0 \leq D'_{ii} &= D_{ii} - \b{\nu_1 - \nu_\text{min}} & D_{ii} &= D'_{ii} + \b{\nu_1 - \nu_\text{min}}.
\end{align}
Thus, the polynomial of interest (Eq.~\ref{eq:general_poly}) becomes,
\begin{align}
  0 &= - \Lambda_{ii} + \prod_{\ell=1}^{L+1} \b{D'_{ii} + \b{\nu_1 - \nu_\text{min}} - \b{\nu_1 - \nu_\ell}} = - \Lambda_{ii} + \prod_{\ell=1}^{L+1} \b{D'_{ii} + \b{\nu_\ell - \nu_\text{min}}}
\end{align}
where $0 < \nu_\ell - \nu_\text{min}$ as $\nu_\text{min}$ is defined to be the smallest $\nu_\ell$ (Eq.~\ref{eq:def:nu_min}). 
Thus, the constant term, $-\Lambda_{ii}$ is negative, while all other terms, $D'_{ii},\dotsc,(D'_{ii})^{L+1}$ in the polynomial have positive coefficients. Thus, there is only one sign change, which proves the existence of only one valid real root, as required.

\section{Unimodality experiments with nonlinear kernels} \label{sec:unimodality}

For the posterior over Gram matrices to converge to a point distribution, we need the DKM objective $\L(\G_1,\dotsc,\G_L)$ to have one unique global optimum.
As noted above, this is guaranteed when the prior dominates (Eq.~\ref{eq:nngp_lim}), and for linear models (Appendix~\ref{app:unimodality}).
While we believe that it might be possible to construct counter examples, in practice we expect a single global optimum in most practical settings.
To confirm this expectation, we did a number of experiments, starting with many different random initializations of a deep kernel machine and optimizing using gradient descent (Appendix~\ref{sec:unimodality}).
In all cases tested, the optimizers converged to the same maximum.

We parameterise Gram matrices $\G_\ell=\tfrac{1}{P}\V_\ell \V_\ell^T$ with $\V_\ell\in\mathbb{R}^{P\times P}$ being trainable parameters.
To make initializations with different seeds sufficiently separated while ensuring stability we initialize $\G_\ell$ from a broad distribution that depends on $\K(\G_{\ell-1})$. 
Specifically, we first take the Cholesky decomposition $\K(\G_{\ell-1})=\mathbf{L}_{\ell-1}\mathbf{L}_{\ell-1}^T$, then set $\V_\ell = \mathbf{L}_{\ell-1}\mathbf{\Xi}_\ell\D_\ell^{1/2}$ where each entry of $\mathbf{\Xi}_\ell\in\mathbb{R}^{P\times P}$ is independently sampled from a standard Gaussian, and $\D_\ell$ is a diagonal scaling matrix with each entry sampled i.i.d.\ from an inverse-Gamma distribution. 
The variance of the inverse-Gamma distribution is fixed to 100, and the mean is drawn from a uniform distribution $U[0.5,3]$ for each seed. 
Since for any random variable $x\sim\text{Inv-Gamma}(\alpha,\beta)$, $\E(x) = \tfrac{\beta}{\alpha-1}$ and $\Var(x)=\tfrac{\beta}{(\alpha-1)(\alpha-2)}$, once we fix the mean and variance we can compute $\alpha$ and $\beta$ as
\begin{align}
    \alpha &= \frac{\E(x)^2}{\Var(x)} + 2, \\
    \beta &= \E(x) (\alpha-1).
\end{align}

\begin{figure*}[t]
  \centering
  \includegraphics[width=\textwidth]{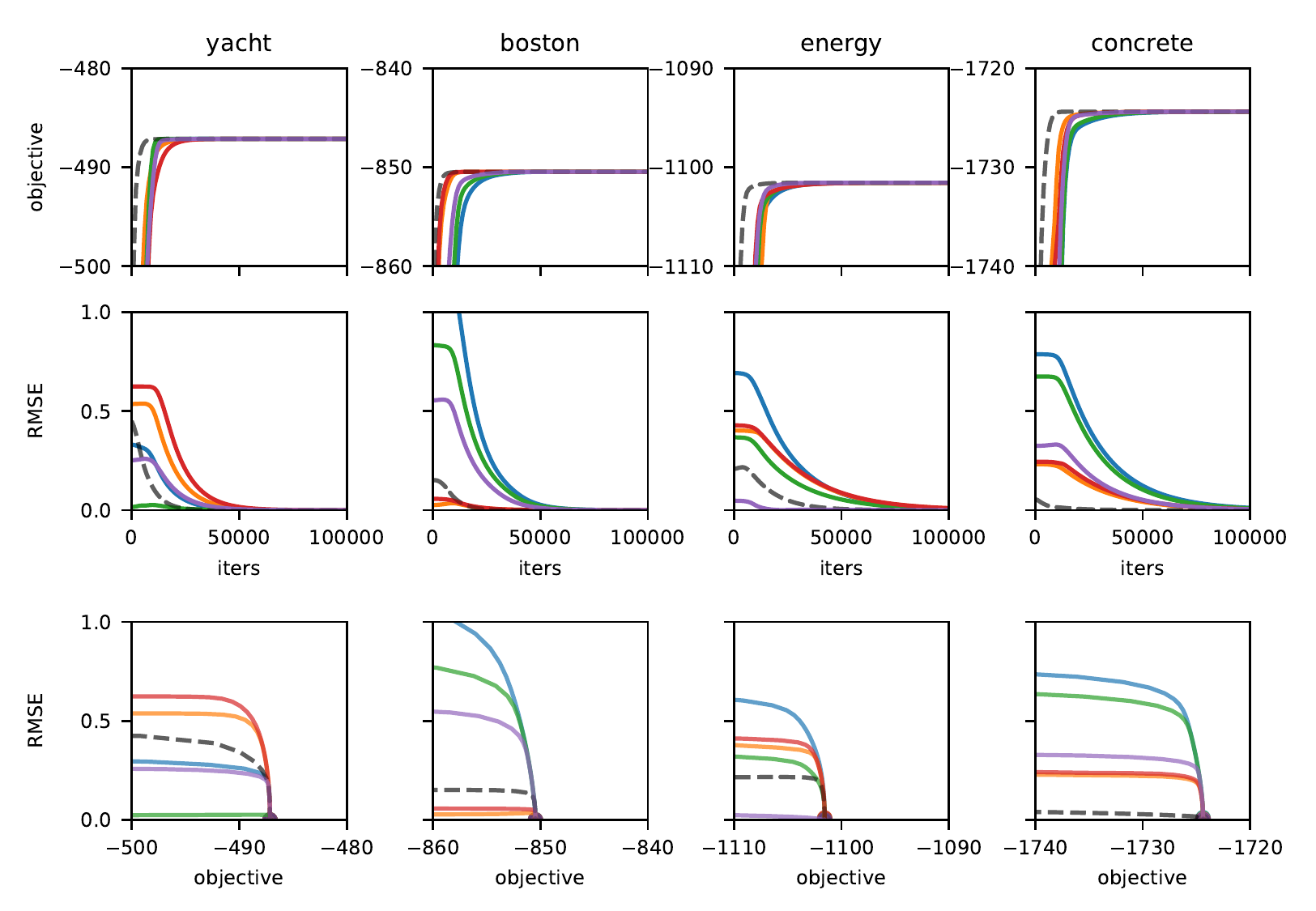}
  \caption{
    One-layer DKMs with squared exponential kernel trained on full UCI datasets (through columns) converges to the same solution, despite very different initializations by applying stochastic diagonal scalings described in Appendix~\ref{sec:methods} to the standard initialization with different seeds. Standard initialization is shown in dashed line, while scaled initializations are the color lines each denoting a different seed. 
    The first row shows the objective during training for all seeds that all converge to the same value. 
    The second row shows the element-wise RMSE between the Gram matrix of each seed and the optimized Gram matrix obtained from the standard initialization. 
    RMSE converges to 0 as all initializations converge on the same maximum.
    The last row plots RMSE versus objective value, again showing a single optimal objective value where all Gram matrices are the same.
    \label{fig:unimodal_onelayer}
  }
\end{figure*}

\begin{figure}
  \includegraphics[width=\textwidth]{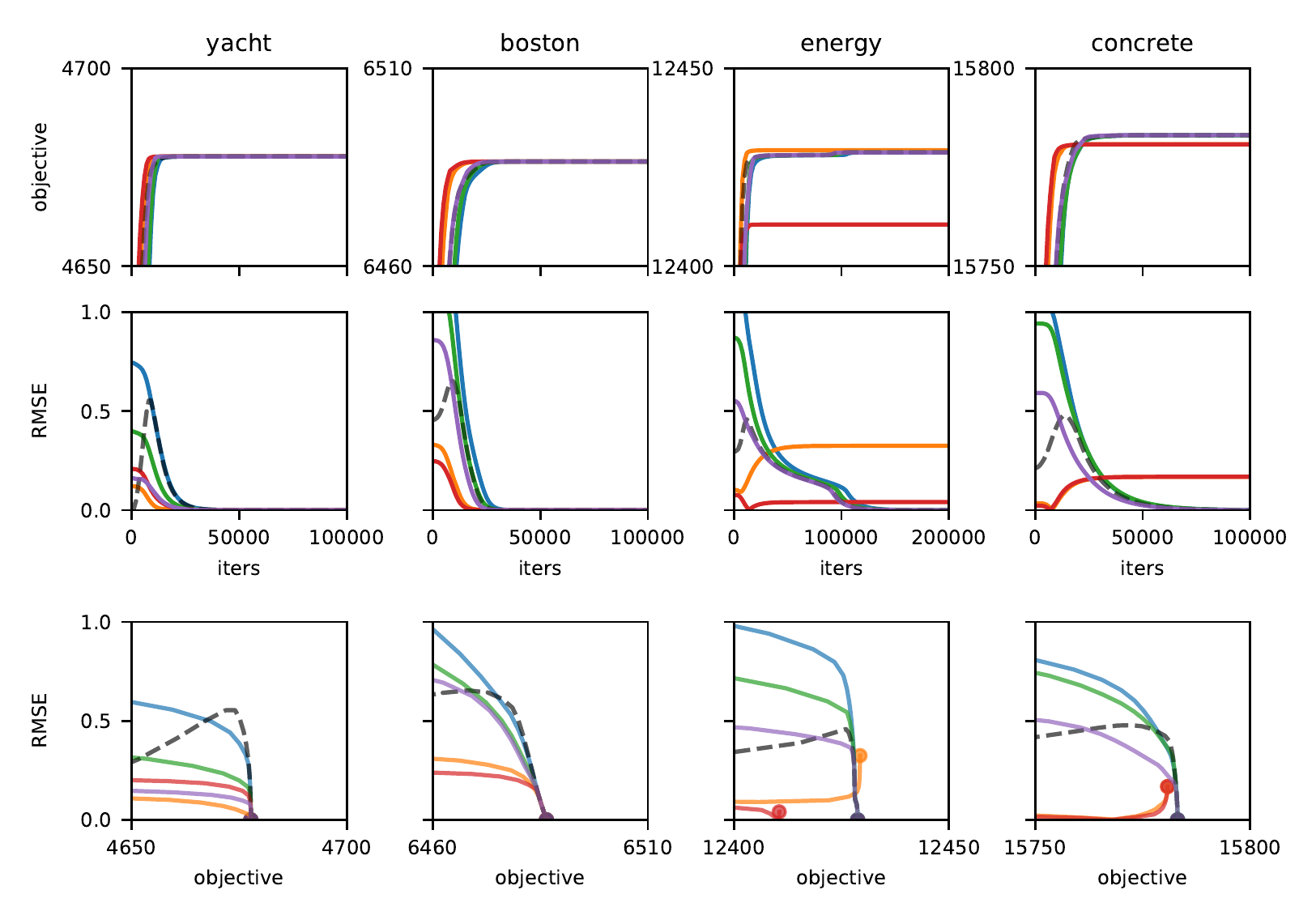}
  \caption{
    One-layer DGP with MAP inference over features as described in Appendix~\ref{sec:MAP} Eq.~\eqref{eq:FMAP}. Rows and columns are the same as in Figure~\ref{fig:unimodal_onelayer}. Using the same randomly scaled initializations described above, we are able to find multiple modes in energy and concrete showing our initializations are diverse enough, albeit there is still only a single global optimum.
    \label{fig:unimodal_map}
  }
\end{figure}

We set $\nu_\ell=5$, and use the Adam optimizer \citep{kingma2014adam} with learning rate $0.001$ to optimize parameters $\V_\ell$ described above. 
We fixed all model hyperparameters to ensure that any multimodality could emerge only from the underlying deep kernel machine.
As we did not use inducing points, we were forced to consider only the smaller UCI datasets (yacht, boston, energy and concrete).
For the deep kernel machine objective, all Gram matrices converge rapidly to the same solution, as measured by RMSE (Fig.~\ref{fig:unimodal_onelayer}).
Critically, we did find multiple modes for the MAP objective (Fig~\ref{fig:unimodal_map}), indicating that experiments are indeed powerful enough to find multiple modes (though of course they cannot be guaranteed to find them).
Finally, note that the Gram matrices took a surprisingly long time to converge: this was largely due to the high degree of diversity in the initializations; convergence was much faster if we initialised deterministically from the prior.

This might contradict our usual intuitions about huge multimodality in the weights/features of BNNs and DGPs.
This can be reconciled by noting that each mode, written in terms of Gram matrices, corresponds to (perhaps infinitely) many modal features.
In particular, in Sec.~\ref{sec:MAP}, we show that the log-probability for features, $\P{\F_\ell| \F_{\ell-1}}$ (Eq.~\ref{eq:PFF}) depends only on the Gram matrices, and note that there are many settings of features which give the same Gram matrix.
In particular, the Gram matrix is the same for any unitary transformation of the features, $\F_\ell' = \F_\ell \U$, satisfying $\U \U^T = \I$, as 
  $\tfrac{1}{N_\ell} \F^\prime_\ell \F_\ell^{\prime T} = \tfrac{1}{N_\ell} \F_\ell \U_\ell \U_\ell^T \F_\ell^T 
  =\tfrac{1}{N_\ell} \F_\ell \F_\ell^T = \G_\ell.$
For DGPs we can use any unitary matrix, so there are infinitely many sets of features consistent with a particular Gram matrix, while for BNNs we can only use permutation matrices, which are a subset of unitary matrices.
%\end{align}
% The true DGP posterior probability, is invariant under any such unitary transformations,
% \begin{multline}
%   \P{\F'_1,\dotsc,\F'_L, \F_{L+1}| \X, \Y} \\= \P{\F_1,\dotsc,\F_L, \F_{L+1}| \X, \Y}.
% \end{multline}
% This is intuitive, if we consider the posterior over $\F_\ell$ conditioned on the next features, $\F_{\ell+1}$, and the previous features, $\F_{\ell-1}$.
% \begin{align}
%   \P{\F_\ell| \F_{\ell+1}, \F_{\ell-1}}
%   &\propto \P{\F_{\ell+1}, \F_\ell| \F_{\ell-1}} \\
%   &= \P{\F_{\ell+1} | \F_\ell} \P{\F_\ell| \F_{\ell-1}}
% \end{align}
% The likelihood, $\P{\F_{\ell+1} | \F_\ell}$, is invariant to transformations of $\F_\ell$ because the downstream kernel depends only on the Gram matrix, and the Gram matrix is invariant to rotations of $\F_\ell$ (Eq.~\eqref{eq:gram-rotation}).
% At the same time, the prior, $\P{\F_\ell| \F_{\ell-1}}$ is invariant to unitary transformations of $\F_\ell$ because the distribution over $\F$ is zero-mean and independent across channel
% (For further details, see Appendix~D in \citet{aitchison2020deep}).
%Interestingly, a very similar phenomenon occurs in (Bayesian) neural networks, which are invariant to a specific family of unitary transformations --- permutations (again, see Appendix~D in \citet{aitchison2020deep}, or \citep{mackay1992practical,chen1993geometry}).
Thus, the objective landscape must be far more complex in the feature domain than with Gram matrices, as a single optimal Gram matrix corresponds to a large family of optimal features.

\FloatBarrier
\clearpage
\section{Inducing point DKMs}
\label{sec:app:pred}
To do large-scale experiments on UCI datasets, we introduce inducing point DKMs by extending Gaussian process inducing point methods \citep{damianou2013deep,salimbeni2017doubly} to the DKM setting.
This approach uses the variational interpretation of the deep kernel machine objective described in Appendix~\ref{sec:true_post:bnn}.

To do inducing-point variational inference, we need to explicitly introduce top-layer features mirroring $\F_{L+1} \in \mathbb{R}^{P \times \nu_{L+1}}$ in Appendix~\ref{app:like}, but replicated $N$ times, $\Fr_{L+1} \in \mathbb{R}^{P\times N_{L+1}}$.
Formally, each feature, $\fr_1^{L+1},\dotsc,\fr_{N_{L+1}}^{L+1}$ is IID, conditioned on $\F_L$,
\begin{subequations}
\begin{align}
  \P{\Fr_{L+1}| \F_{L}} &= \prodln \N{\fr_\lambda^{L+1}; \0, \K(\G(\F_L))},\\
  \P{\Yr| \Fr_{L+1}} &= \prodln \N{\yr_\lambda; \fr_\lambda^{L+1}, \sigma^2 \I},
\end{align}
\end{subequations}
where we give the likelihood for regression, but other likelihoods (e.g.\ for classification) are possible (Appendix~\ref{app:like}).

Further, we take the total number of points, $P$, to be made up of $P_\text{i}$ inducing points and $P_\text{t}$ test/train points, so that $P = P_\text{i} + P_\text{t}$.
Thus, we can separate all features, $\F_\ell \in \mathbb{R}^{P\times N_\ell}$, into the inducing features, $\F_\text{i}^\ell \in \mathbb{R}^{P_\text{i}\times N_\ell}$, and the test/train features, $\F_\text{t}^\ell \in \mathbb{R}^{P_\text{t}\times N_\ell}$.  
Likewise, we separate the inputs, $\X$, and outputs, $\Y$, into (potentially trained) inducing inputs, $\X_\text{i}$, and trained inducing outputs, $\Y_\text{i}$, and the real test/training inputs, $\X_\text{t}$, and outputs, $\Y_\text{t}$,
\begin{align}
  \F_\ell &= \begin{pmatrix} \F_\text{i}^\ell \\ \F_\text{t}^\ell \end{pmatrix} & 
  \Fr_{L+1} &= \begin{pmatrix} \Fr_\text{i}^{L+1} \\ \Fr_\text{t}^{L+1} \end{pmatrix} & 
  \X &= \begin{pmatrix} \X_\text{i} \\ \X_\text{t} \end{pmatrix} & 
  \Y &= \begin{pmatrix} \Y_\text{i} \\ \Y_\text{t} \end{pmatrix} &
  \Yr &= \begin{pmatrix} \Yr_\text{i} \\ \Yr_\text{t} \end{pmatrix}
\end{align}
We follow the usual doubly stochastic inducing point approach for DGPs.
In particular, we treat all the features at intermediate layers, $\F_1, \dotsc, \F_L$, and the top-layer train/test features, $\F_\text{t}^{L+1}$ as latent variables.
However, we deviate from the usual setup in treating the top-layer inducing outputs, $\F_\text{i}^{L+1}$, as learned parameters and maximize over them to ensure that the ultimate method does not require sampling, and at the same time allows minibatched training.
The prior and approximate posterior over $\F_1,\dotsc,\F_L$ are given by,
\begin{subequations}
\label{eq:prior_ap:layers}
\begin{align}
  \Q{\F_1,\dotsc\F_L| \X} &= \tprod_{\ell=1}^{L} \Q{\F_\ell| \F_{\ell-1}},\\
  \P{\F_1,\dots,\F_L| \X} &= \tprod_{\ell=1}^{L} \P{\F_\ell| \F_{\ell-1}},
\end{align}
\end{subequations}
and remember $\F_0 = \X$, so $\G_0 = \tfrac{1}{N_0} \X \X^T$.
The prior and approximate posterior at each layer factorises into a distribution over the inducing points and a distribution over the test/train points,
\begin{subequations}
\label{eq:prior_ap:factors}
\begin{align}
  \Q{\F_\ell| \F_{\ell-1}} &= \P{\F_\text{t}^\ell| \F_\text{i}^\ell, \F_{\ell-1}} \Q{\F_\text{i}^\ell},\\
  \P{\F_\ell| \F_{\ell-1}} &= \P{\F_\text{t}^\ell| \F_\text{i}^\ell, \F_{\ell-1}} \P{\F_\text{i}^\ell| \F_\text{i}^{\ell-1}}.
\end{align}
\end{subequations}
Critically, the approximate posterior samples for the test/train points is the conditional prior $\P{\F_\text{t}^\ell| \F_\text{i}^\ell, \F_{\ell-1}}$, which is going to lead to cancellation when we compute the ELBO.
Likewise, the approximate posterior over $\Fr^{L+1}_\text{t}$ is the conditional prior,
\begin{align}
  \label{eq:QFr}
  \Q{\Fr^{L+1}_\text{t}| \F^{L+1}_\text{i}, \F_L} &= \P{\Fr^{L+1}_\text{t}| \F^{L+1}_\text{i}, \F_L}.
\end{align}
Concretely, the prior approximate posterior over inducing points are given by,
\begin{subequations}
\begin{align}
  \label{eq:QFi}
  \Q{\F^\ell_\text{i}} &= \prodln \N{\f^{\ell}_{\text{i}; \lambda}; \0, \G_\text{ii}^\ell},\\
  \P{\F^\ell_\text{i}| \F_\text{i}^{\ell-1}} &= \prodln \N{\f^{\ell}_{\text{i}; \lambda}; \0, \K(\G(\F^{\ell-1}_\text{i}))}
\end{align}
\end{subequations}
The approximate posterior is directly analogous to Eq.~\eqref{eq:dgp:mvg_post_app} and the prior is directly analogous to Eq.~\eqref{eq:deepgp:top:F}, but where we have specified that this is only over inducing points.
Now we compute the ELBO
\begin{align}
  \text{ELBO}(\F_\text{i}^{L+1},\G_\text{ii}^1,\dotsc,\G_\text{ii}^L)
  = \E_{\operatorname{Q}}\sqb{\log \P{\Yr_\text{t}| \Fr^{L+1}_\text{t}} + \log \frac{\P{\Fr^{L+1}_\text{t}| \F^{L+1}_\text{i}, \F_L} \P{\F_1,\dotsc\F_L| \X}}{\Q{\Fr^{L+1}_\text{t}| \F^{L+1}_\text{i}, \F_L} \Q{\F_1,\dotsc\F_L| \X}}}
\end{align}
Note that the $\P{\F_\text{t}^\ell| \F_\text{i}^\ell, \F_{\ell-1}}$ terms are going to cancel in the ELBO (we consider them below when we come to describing sampling).
Substituting Eq.~(\ref{eq:prior_ap:layers}--\ref{eq:QFr}) and cancelling $\P{\F_\text{t}^\ell| \F_\text{i}^\ell, \F_{\ell-1}}$ and $\P{\Fr^{L+1}_\text{t}| \F^{L+1}_\text{i}, \F_L}$,
\begin{align}
  \text{ELBO}(\F_\text{i}^{L+1},\G_\text{ii}^1,\dotsc,\G_\text{ii}^L) = \E_{\operatorname{Q}}\sqb{\log \P{\Yr_\text{t}| \Fr^{L+1}_\text{t}} + \sum_{\ell=1}^L \log \frac{\P{\F_\text{i}^\ell| \F_\text{i}^{\ell-1}} }{\Q{\F_\text{i}^\ell}}}.
\end{align}
So far, we have treated the Gram matrices, $\G_\text{ii}^\ell$ as parameters of the approximate posterior.
However, in the infinite limit $N\rightarrow \infty$, these are consistent with the features generated by the approximate posterior.
In particular the matrix product $\tfrac{1}{N_\ell} \F_\text{i}^\ell \b{\F_\text{i}^\ell}^T$ can be written as an average over infinitely many IID vectors, $\f_{\text{i}; \lambda}^\ell$ (first equality), and by the law of large numbers, this is equal to the expectation of one term (second equality), which is $\G_\text{ii}^\ell$ (by the approximate posterior Eq.~\eqref{eq:QFi}),
\begin{align}
  \label{eq:large_numbers}
  \lim_{N \rightarrow \infty} \tfrac{1}{N_\ell} \F_\text{i}^\ell \b{\F_\text{i}^\ell}^T 
  = \lim_{N \rightarrow \infty} \tfrac{1}{N_\ell} \tsum_{\lambda=1}^{N_\ell} \f_{\text{i};\lambda}^\ell \b{\f_{\text{i}; \lambda}^\ell}^T
  = \E_{\Q{\f_{\text{i};\lambda}^\ell}}\sqb{\f_{\text{i};\lambda}^\ell \b{\f_{\text{i}; \lambda}^\ell}^T}
  = \G_\text{ii}^\ell.
\end{align}
By this argument, the Gram matrix from the previous layer, $\G_\text{ii}^{\ell-1}$ is deterministic. 
Further, in a DGP, $\F_\text{i}^\ell$ only depends on $\F_\text{i}^{\ell-1}$ through $\G_\text{ii}^{\ell-1}$ (Eq.~\ref{eq:prior:bnn}), and the prior and approximate posterior factorise.
Thus, in the infinite limit, individual terms in the ELBO can be written,
\begin{align}
  \lim_{N \rightarrow \infty} \tfrac{1}{N} \E_{\operatorname{Q}}\sqb{\log \frac{\P{\F_\text{i}^\ell| \F_\text{i}^{\ell-1}}}{\Q{\F_\text{i}^\ell}}} &= 
  \nu_\ell \E_{\operatorname{Q}}\sqb{\log \frac{\P{\f_{\text{i}; \lambda}^\ell| \G_\text{ii}^{\ell-1}}}{\Q{\f_{\text{i}; \lambda}^\ell}}}\\
  &= -\nu_\ell \KL{\N{\0, \K(\G_\text{ii}^{\ell})}}{\N{\0, \G_\text{ii}^{\ell-1}}},
\end{align}
where the final equality arises when we notice that the expectation can be written as a KL-divergence.
The inducing DKM objective, $\mathcal{L}_\text{ind}$, is the ELBO, divided by N to ensure that it remains finite in the infinite limit,
\begin{align}
  \mathcal{L}_\text{ind} (\F^{L+1}_\text{i}, \G_\text{ii}^1,\dotsc,\G_\text{ii}^L) &{=} 
  \lim_{N \rightarrow \infty} \tfrac{1}{N} \text{ELBO}(\F_\text{i}^{L+1},\G_\text{ii}^1,\dotsc,\G_\text{ii}^L)\\
  \nonumber
  &{=} \E_{\operatorname{Q}}\sqb{\log \P{\Y_\text{t}| \F_\text{t}^{L+1}}} {-} \sum_{\ell=1}^L \nu_\ell \KL{\N{\0, \K(\G_\text{ii}^{\ell})}}{\N{\0, \G_\text{ii}^{\ell-1}}}.
\end{align}
Note that this has almost exactly the same form as the standard DKM objective for DGPs in the main text (Eq.~\ref{eq:post_dkm_obj}). 
In particular, the second term is a chain of KL-divergences, with the only difference that these KL-divergences apply only to the inducing points.
The first term is a ``performance'' term that here depends on the quality of the predictions given the inducing points.
As the copies are IID, we have,
\begin{align}
  \E_{\operatorname{Q}}\sqb{\log \P{\Yr_\text{t}| \Fr_\text{t}^{L+1}}}  &= 
  N \E_{\operatorname{Q}}\sqb{\log \P{\Y_\text{t}| \F_\text{t}^{L+1}}}.
\end{align}

Now that we have a simple form for the ELBO, we need to compute the expected likelihood, $\E_{\operatorname{Q}}\sqb{\log \P{\Y_\text{t}| \F_\text{t}^{L+1}}}$. 
This requires us to compute the full Gram matrices, including test/train points, conditioned on the optimized inducing Gram matrices.
We start by defining the full Gram matrix,
\begin{align}
  \G_\ell &= \begin{pmatrix}
    \G^\ell_\text{ii} & \G^\ell_\text{it}\\
    \G^\ell_\text{ti} & \G^\ell_\text{tt}
  \end{pmatrix}
\end{align}
for both inducing points (labelled ``i'') and test/training points (labelled ``t'') from just $\G^\ell_\text{ii}$.
For clarity, we have $\G_\ell \in \mathbb{R}^{P \times P}$, $\G^\ell_\text{ii} \in \mathbb{R}^{P_\text{i} \times P_\text{i}}$, $\G^\ell_\text{ti} \in \mathbb{R}^{P_\text{t} \times P_\text{i}}$,  $\G^\ell_\text{tt} \in \mathbb{R}^{P_\text{t} \times P_\text{t}}$, where $P_\text{i}$ is the number of inducing points, $P_\text{t}$ is the number of train/test points and $P = P_\text{i} + P_\text{t}$ is the total number of inducing and train/test points.

The conditional distribution over $\F^\ell_\text{t}$ given $\F^\ell_\text{i}$ is,
\begin{align}
  \label{eq:Ft|Fi}
  %\Pc{\f^\ell_{\text{t}; \lambda}}{\f^\ell_{\text{i}; \lambda}, \G_{\ell-1}} &= \N{\K_\text{ti} \K_\text{ii}^{-1} \f^\ell_{\text{i}; \lambda}, \K_{\text{tt}\cdot \text{i}}}
  \Pc{\F^\ell_{\text{t}}}{\F^\ell_{\text{i}}, \G_{\ell-1}} &= \prodln \N{\f^\ell_{\text{t}; \lambda}; \K_\text{ti} \K_\text{ii}^{-1} \f^\ell_{\text{i}; \lambda}, \K_{\text{tt}\cdot \text{i}}}
\end{align}
where $\f^\ell_{\text{t}; \lambda}$ is the activation of the $\lambda$th feature for all train/test inputs, $\f^\ell_{\text{i}; \lambda}$ is the activation of the $\lambda$th feature for all train/test inputs, and $\f^\ell_{\text{i}; \lambda}$, and 
\begin{align}
    \begin{pmatrix}
    \K_\text{ii} & \K_\text{ti}^T\\
    \K_\text{ti} & \K_\text{tt}
  \end{pmatrix}
  &= \K\b{\tfrac{1}{N_{\ell-1}} \F_{\ell-1} \F_{\ell-1}^T} = \K\b{\G_{\ell-1}}\\
  \K_{\text{tt}\cdot \text{i}} &= \K_\text{tt} - \K_\text{ti} \K_\text{ii}^{-1} \K_\text{ti}^T.
\end{align}
In the infinite limit, the Gram matrix becomes deterministic via the law of large numbers (as in Eq.~\ref{eq:large_numbers}), and as such $\G_\text{it}$ and $\G_\text{tt}$ become deterministic and equal to their expected values.
Using Eq.~\eqref{eq:Ft|Fi}, we can write,
\begin{align}
  \F^\ell_\text{t} &= \K_\text{ti} \K_\text{ii}^{-1} \F^\ell_\text{i} + \K_{\text{tt}\cdot \text{i}}^{1/2} \mathbf{\Xi}.
\end{align}
where $\mathbf{\Xi}$ is a matrix with IID standard Gaussian elements.
Thus,
\begin{align}
  \G^\ell_\text{ti} &= \tfrac{1}{\nu} \E\sqb{\F^\ell_\text{t} (\F^{\ell}_\text{i})^T} \\
  &= \tfrac{1}{\nu} \K_\text{ti} \K_\text{ii}^{-1} \E\sqb{\F^\ell_\text{i} (\F^{\ell}_\text{i})^T} \\
  &= \K_\text{ti} \K_\text{ii}^{-1} \G^\ell_\text{ii}
%\end{align}
%We can do the same for $\F_\text{t} \F_\text{t}^T$:
%\begin{align}
  \intertext{and,}
  \G_\text{tt}^\ell &= \tfrac{1}{\nu} \E\sqb{\F^\ell_\text{t} (\F^{\ell}_\text{t})^T} \\
  &= \tfrac{1}{\nu} \K_\text{ti} \K_\text{ii}^{-1} \E\sqb{\F^\ell_\text{i} (\F^{\ell}_\text{i})^T} \K_\text{ii}^{-1} \K^T_\text{ti} + \tfrac{1}{\nu} \K_{\text{tt}\cdot \text{i}}^{1/2} \E\sqb{\mathbf{\Xi} \mathbf{\Xi}^T} \K_{\text{tt}\cdot \text{i}}^{1/2}\\
  &= \K_\text{ti} \K_\text{ii}^{-1} \G_\text{ii} \K_\text{ii}^{-1} \K^T_\text{ti} + \K_{\text{tt}\cdot \text{i}}
\end{align}
For the full prediction algorithm, see Alg.~\ref{alg:pred}.

\begin{algorithm}[t]
\caption{DKM prediction
  \label{alg:pred}
}
\begin{algorithmic}
  \STATE {\bfseries Parameters:} $\slL{\nu_\ell}$
  \STATE {\bfseries Optimized Gram matrices} $\slL{\G^\ell_\text{ii}}$
  \STATE {\bfseries Inducing and train/test inputs:} $\X_\text{i}$, $\X_\text{t}$
  \STATE {\bfseries Inducing outputs:} $\F_\text{i}^{L+1}$
  \STATE \textcolor{gray}{Initialize full Gram matrix}
  \STATE $\begin{pmatrix}
    \G^0_\text{ii} & \G^{0; T}_\text{ti} \\
    \G^0_\text{ti} & \G^0_\text{tt}
  \end{pmatrix} = \frac{1}{\nu_0}
  \begin{pmatrix}
    \X_\text{i} \X_\text{i}^T & \X_\text{i} \X_\text{t}^T\\
    \X_\text{t} \X_\text{i}^T & \X_\text{t} \X_\text{t}^T
  \end{pmatrix}$
  \STATE \textcolor{gray}{Propagate full Gram matrix}
  \FOR{$\ell$ {\bfseries in} $(1,\dotsc,L)$}
    \STATE $\begin{pmatrix}
      \K_\text{ii} & \K_\text{ti}^T \\
      \K_\text{ti} & \K_\text{tt}
    \end{pmatrix} = \K\b{
    \begin{pmatrix}
      \G^{\ell-1}_\text{ii} & (\G^{\ell-1}_\text{ti})^T \\
      \G^{\ell-1}_\text{ti} & \G^{\ell-1}_\text{tt}
    \end{pmatrix}}$
    \STATE $\K_{\text{tt}\cdot \text{i}} = \K_\text{tt} - \K_\text{ti} \K_\text{ii}^{-1} \K_\text{ti}^T.$
    \STATE $\G_\text{ti}^\ell = \K_\text{ti} \K_\text{ii}^{-1} \G^\ell_\text{ii}$
    \STATE $\G_\text{tt}^\ell = \K_\text{ti} \K_\text{ii}^{-1} \G^\ell_\text{ii} \K_\text{ii}^{-1} \K_\text{ti}^T + \K_{\text{tt}\cdot \text{i}}$
  \ENDFOR
  \STATE \textcolor{gray}{Final prediction using standard Gaussian process expressions}
    \STATE $\begin{pmatrix}
      \K_\text{ii} & \K_\text{ti}^T \\
      \K_\text{ti} & \K_\text{tt}
    \end{pmatrix} = \K\b{
    \begin{pmatrix}
      \G^{L}_\text{ii} & (\G^{L}_\text{ti})^T \\
      \G^{L}_\text{ti} & \G^{L}_\text{tt}
    \end{pmatrix}}$ 
  \STATE $\Y_\text{t} \sim \N{\K_\text{ti} \K_\text{ii}^{-1} \F_\text{i}^{L+1}, \K_\text{tt} - \K_\text{ti} \K_\text{ii}^{-1} \K_\text{ti}^T + \sigma^2 \I}$
\end{algorithmic}
\end{algorithm}

\end{document}